\documentclass[10pt,twocolumn,letterpaper]{article}

\usepackage{cvpr}              

\usepackage[dvipsnames]{xcolor}

\usepackage{graphicx}
\usepackage{amsmath}
\usepackage{amssymb}
\usepackage{booktabs,caption}
\usepackage{multirow}   
\usepackage{bm}
\usepackage{dsfont}
\usepackage[linesnumbered,ruled,vlined]{algorithm2e}
\usepackage{makecell}
\usepackage[flushleft]{threeparttable}
\usepackage{enumitem}
\usepackage{siunitx}
\usepackage[accsupp]{axessibility} 

\usepackage[dvipsnames]{xcolor}

\definecolor{MyBlack}{HTML}{323A45}

\graphicspath{{figures/}}
\definecolor{cvprblue}{rgb}{0.21,0.49,0.74}
\usepackage[pagebackref,breaklinks,colorlinks,citecolor=cvprblue]{hyperref}
\usepackage[capitalize]{cleveref}

\def\eg{\emph{e.g.,\ }}
\def\ie{\emph{i.e.,\ }}

\def\paperID{103} 
\def\confName{CVPR}
\def\confYear{2024}

\begin{document}
\title{The Ninth NTIRE 2024 Efficient Super-Resolution Challenge Report}


\author{Bin Ren$^*$ \and
Yawei Li$^*$ \and
Nancy Mehta$^*$ \and
Radu Timofte$^*$ \and
Hongyuan Yu \and
Cheng Wan \and
Yuxin Hong \and
Bingnan Han \and
Zhuoyuan Wu \and
Yajun Zou \and
Yuqing Liu \and
Jizhe Li \and
Keji He \and
Chao Fan \and
Heng Zhang \and
Xiaolin Zhang \and
Xuanwu Yin \and
Kunlong Zuo \and
Bohao Liao \and
Peizhe Xia \and
Long Peng \and
Zhibo Du \and
Xin Di \and
Wangkai Li \and
Yang Wang \and
Wei Zhai \and
Renjing Pei \and
Jiaming Guo \and
Songcen Xu \and
Yang Cao \and
Zhengjun Zha \and
Yan Wang \and
Yi Liu \and
Qing Wang \and
Gang Zhang \and
Liou Zhang \and
Shijie Zhao \and
Long Sun \and
Jinshan Pan \and
Jiangxin Dong \and
Jinhui Tang \and
Xin Liu \and
Min Yan \and
Qian Wang \and
Menghan Zhou \and 
Yiqiang Yan \and
Yixuan Liu \and
Wensong Chan \and
Dehua Tang \and
Dong Zhou \and
Li Wang \and
Lu Tian \and
Barsoum Emad \and
Bohan Jia \and
Junbo Qiao \and
Yunshuai Zhou \and
Yun Zhang \and
Wei Li \and
Shaohui Lin \and
Shenglong Zhou \and
Binbin Chen \and
Jincheng Liao \and
Suiyi Zhao \and
Zhao Zhang \and
Bo Wang \and
Yan Luo \and
Yanyan Wei \and
Feng Li \and
Mingshen Wang \and
Yawei Li \and 
Jinhan Guan \and
Dehua Hu \and
Jiawei Yu \and 
Qisheng Xu \and
Tao Sun \and
Long Lan \and
Kele Xu \and
Xin Lin \and
Jingtong Yue \and
Lehan Yang \and
Shiyi Du \and
Lu Qi \and
Chao Ren \and
Zeyu Han \and 
Yuhan Wang \and 
Chaolin Chen \and 
Haobo Li \and
Mingjun Zheng \and
Zhongbao Yang \and
Lianhong Song \and
Xingzhuo Yan \and
Minghan Fu \and
Jingyi Zhang \and 
Baiang Li \and
Qi Zhu \and 
Xiaogang Xu \and
Dan Guo \and
Chunle Guo \and 
Jiadi Chen \and
Huanhuan Long \and 
Chunjiang Duanmu \and
Xiaoyan Lei \and
Jie Liu \and
Weilin Jia \and
Weifeng Cao \and
Wenlong Zhang \and
Yanyu Mao \and
Ruilong Guo \and
Nihao Zhang \and 
Qian Wang \and 
Manoj Pandey \and 
Maksym Chernozhukov \and 
Giang Le \and 
Shuli Cheng \and
Hongyuan Wang \and
Ziyan Wei \and
Qingting Tang \and
Liejun Wang \and
Yongming Li \and
Yanhui Guo \and 
Hao Xu \and
Akram Khatami-Rizi \and
Ahmad Mahmoudi-Aznaveh \and
Chih-Chung Hsu \and
Chia-Ming Lee \and
Yi-Shiuan Chou \and
Amogh Joshi \and 
Nikhil Akalwadi \and 
Sampada Malagi \and 
Palani Yashaswini \and 
Chaitra Desai \and
Ramesh Ashok Tabib \and 
Ujwala Patil \and 
Uma Mudenagudi
}
\maketitle
\let\thefootnote\relax\footnotetext{
$^*$ B Ren (bin.ren@unitn.it, University of Pisa \& University of Trento, Italy), Y. Li (yawei.li@vision.ee.ethz.ch, ETH Z\"urich, Switzerland), N. Mehta (nancy.mehta@uni-wuerzburg.de, University of W\"urzburg, Germany), and R. Timofte (Radu.Timofte@uni-wuerzburg.de, University of W\"urzburg, Germany) were the challenge organizers, while the other authors participated in the challenge.\\ 
Appendix~\ref{sec:teams} contains the authors' teams and affiliations.\\
NTIRE 2024 webpage: \url{https://cvslai.net/ntire/2024/}.\\ 
Code: \url{https://github.com/Amazingren/NTIRE2024_ESR/}.
}

\begin{abstract}
This paper provides a comprehensive review of the NTIRE 2024 challenge, focusing on efficient single-image super-resolution (ESR) solutions and their outcomes. 
The task of this challenge is to super-resolve an input image with a magnification factor of $\times4$ based on pairs of low and corresponding high-resolution images. 
The primary objective is to develop networks that optimize various aspects such as runtime, parameters, and FLOPs, while still maintaining a peak signal-to-noise ratio (PSNR) of approximately 26.90 dB on the DIV2K\_LSDIR\_valid dataset and 26.99 dB on the DIV2K\_LSDIR\_test dataset. 
In addition, this challenge has 4 tracks including the main track (overall performance), sub-track 1 (runtime), sub-track 2 (FLOPs), and sub-track 3 (parameters). 
In the main track, all three metrics (\ie runtime, FLOPs, and parameter count) were considered. The ranking of the main track is calculated based on a weighted sum-up of the scores of all other sub-tracks.
In sub-track 1, the practical runtime performance of the submissions was evaluated, and the corresponding score was used to determine the ranking.
In sub-track 2, the number of FLOPs was considered. The score calculated based on the corresponding FLOPs was used to determine the ranking.
In sub-track 3, the number of parameters was considered. The score calculated based on the corresponding parameters was used to determine the ranking.
RLFN is set as the baseline for efficiency measurement. The challenge had 262 registered participants, and 34 teams made valid submissions. They gauge the state-of-the-art in efficient single-image super-resolution. To facilitate the reproducibility of the challenge and enable other researchers to build upon these findings, the code and the pre-trained model of validated solutions are made publicly available at \url{https://github.com/Amazingren/NTIRE2024_ESR/}. 
\end{abstract}

\section{Introduction}
\label{sec:introduction}
Single image super-resolution (SR) aims at enhancing the resolution of low-resolution (LR) images to generate high-resolution (HR) counterparts. 
Typically, LR images are acquired through a degradation process that involves blurring and down-sampling. 
Among the models used to simulate this degradation in classical image SR, bicubic down-sampling stands out as widely adopted~\cite{li2023efficient,liang2021swinir,mehta2022adaptive,mehta2023gated}. Its prevalence as a benchmark enables the evaluation of different SR methods and facilitates direct comparisons between them, thereby validating the efficacy of novel SR methods.

Cutting-edge deep neural networks for SR face significant challenges, including parameter overparameterization, resource-intensive computation, and substantial latency. These obstacles hinder their integration into mobile devices for real-time SR applications. However, ongoing innovation continues to address these challenges.
Enter a diverse array of research efforts focused on improving the efficiency of various architectures, including Convolutional Neural Networks (CNNs), Multi-Layer Perceptrons (MLPs), Transformers, and Mamba~\cite{lecun1998gradient,zamir2021multi,bishop2006pattern,tu2022maxim,vision_transformer,ren2023masked,mamba}. From the meticulous approach of network pruning~\cite{liu2019metapruning,li2020dhp} to the straightforward technique of low-rank filter decomposition, and from the systematic process of network quantization to the advanced methods of neural architecture search~\cite{zoph2017nas,zoph2018nas,li2021heterogeneity}, a range of effective solutions have emerged. Among these, knowledge distillation stands out as particularly promising. These efforts in network compression mark a significant advancement for image SR, offering not only improved resolution but also enhanced efficiency and accessibility for all.

The efficiency of a deep neural network encompasses various dimensions, evaluated across a range of metrics such as runtime, parameter count, and computational complexity (measured in FLOPs). These metrics play a crucial role in determining the network's feasibility for deployment across diverse platforms. Among these, runtime emerges as particularly significant, providing a direct indication of a network's operational efficiency and often serving as the primary criterion for evaluation.
Of utmost concern is the relationship between computational complexity and energy consumption, a pivotal axis that directly impacts the viability of mobile devices. Higher computational complexity correlates with increased energy consumption, posing a significant threat to the delicate balance of battery life. Additionally, the number of parameters exerts a substantial influence on AI chip design, determining chip area and manufacturing costs. An increase in parameter counts can lead to larger chip sizes and elevated production expenses, thereby shaping the landscape of the AI device market.

In partnership with the 2024 New Trends in Image Restoration and Enhancement (NTIRE 2024) workshop, we are proud to announce the inception of the Efficient Super-Resolution Challenge. The challenge's primary objective is to achieve super-resolution of a low-resolution (LR) image with a magnification factor of $\times4$, employing a network that optimizes runtime, parameters, and FLOPs, building upon the baseline method laid by RLFN~\cite{kong2022residual}. 
Participants are tasked with maintaining a minimum peak signal-to-noise ratio (PSNR) of 26.90 dB on the DIV2K\_LSDIR\_valid dataset and 26.99 dB on the DIV2K\_LSDIR\_test dataset.
This challenge serves as a platform for exploring state-of-the-art solutions in efficient super-resolution. We aim to rigorously assess their effectiveness and identify key trends in the design of efficient SR networks. We welcome participants to contribute to this endeavor by pushing the boundaries of innovation and advancing streamlined and effective image restoration and enhancement techniques.

This challenge is one of the NTIRE 2024 Workshop~\footnote{\url{https://cvlai.net/ntire/2024/}} associated challenges on: dense and non-homogeneous dehazing~\cite{ntire2024dehazing}, night photography rendering~\cite{ntire2024night}, blind compressed image enhancement~\cite{ntire2024compressed}, shadow removal~\cite{ntire2024shadow}, efficient super resolution (this challenge), image super resolution ($\times$4)~\cite{ntire2024srx4}, light field image super-resolution~\cite{ntire2024lightfield}, stereo image super-resolution~\cite{ntire2024stereosr}, HR depth from images of specular and transparent surfaces~\cite{ntire2024depth}, bracketing image restoration and enhancement~\cite{ntire2024bracketing}, portrait quality assessment~\cite{ntire2024QA_portrait}, quality assessment for AI-generated content~\cite{ntire2024QA_AI}, restore any image model (RAIM) in the wild~\cite{ntire2024raim}, RAW image super-resolution~\cite{ntire2024rawsr}, short-form UGC video quality assessment~\cite{ntire2024QA_UGC}, low light enhancement~\cite{ntire2024lowlight}, and RAW burst alignment and ISP challenge.
\section{NTIRE 2024 Efficient Super-Resolution Challenge}
\label{sec:esr_challenge}
The objectives of this challenge are multifaceted: 
(1) To stimulate curiosity and exploration within the field of efficient super-resolution.
(2) To create a platform where diverse methodologies can be directly compared in terms of efficiency.
(3) And to act as a dynamic hub where academic and industrial leaders come together, exchanging ideas and laying the groundwork for potential collaborations.
This section will elucidate the intricate details of the challenge, providing participants with clear guidance through its complex structure and objectives.

\subsection{Dataset}
The DIV2K~\cite{agustsson2017ntire} dataset and the LSDIR~\cite{lilsdir} dataset are utilized for this challenge. 
Specifically, the DIV2K dataset consists of 1,000 diverse 2K resolution RGB images, which are split into a training set of 800 images, a validation set of 100 images, and a test set of 100 images. 
The LSDIR dataset contains 86,991 high-resolution high-quality images, which are split into a training set of 84,991 images, a validation set of 1,000 images, and a test set of 1,000 images. 
In this challenge, the corresponding LR DIV2K and LSDIR images are generated by bicubic downsampling with a down-scaling factor of $\times$4. 
The training images from DIV2K and LSDIR are provided to the participants of the challenge. 
During the validation phase, 100 images from the DIV2K validation set and 100 images from the LSDIR validation set, forming the DIV2K\_LSDIR\_valid set, which is made available to participants. 
During the test phase, 100 images from the DIV2K test set and another 100 images from the LSDIR test set are used, forming the DIV2K\_LSDIR\_test set.
Throughout the entire challenge, the testing HR images remain hidden from the participants.

\subsection{RLFN Baseline Model}
\label{sec:baseline_model}
The Residual Feature Distillation Network (RLFN)~\cite{kong2022residual} serves as the baseline model in this challenge. 
The aim is to improve its efficiency in terms of runtime, number of parameters, and FLOPs, while at least maintaining 26.90 dB on the DIV2K\_LSDIR\_valid dataset and 26.99 dB on the DIV2K\_LSDIR\_test dataset. 

The main idea within RLFN is the use of three convolutional layers for residual local feature learning to simplify feature aggregation, which achieves a good trade-off between model performance and inference time. 
Moreover, the popular contrastive loss was explored and RLFN proposed that the selection of intermediate features of its feature extractor has a great influence on the overall performance. 
Specifically, the initial feature extraction is carried out by a $3\times3$ convolution that generates coarse features from the input LR image. 
The second part of RLFB consists of four RLFBs, stacked in a chain-like manner, to progressively refine the extracted features. 
After gradual refinement by the RLFBs, all intermediate features are combined using a $1\times1$ convolution layer. 
An additional $3\times3$ convolution layer is then utilized to smooth the aggregated features. 
Finally, the super-resolved images are generated by pixel shuffle operation.

The baseline RLFN emerges as the winner of the NTIRE2022 Challenge on Efficient Super-Resolution~\cite{kong2022residual}. The quantitative performance and efficiency metrics of RLFN are given in Table~\ref{tbl:final_results}, and summarized as follows:
(1) The number of parameters is 0.317M. 
(2) The average PSNRs on validation (DIV2K 100 valid images and LSDIR 100 valid images) and testing (DIV2K 100 test images and LSDIR 100 test images) sets of this challenge are 29.96 dB and 27.07 dB, respectively. 
(3) The runtime averaged 11.77 ms on the validation and test set with PyTorch 1.13.1+cu117 on a single NVIDIA GeForce RTX 3090 GPU. 
(4) The number of FLOPs for an input of size $256\times256$ is 19.67G. 

\subsection{Tracks and Competition}
This challenge aims to devise a network that reduces one or several aspects such as runtime, parameters, and FLOPs, while at least maintaining the 26.90 dB on the DIV2K\_LSDIR\_valid dataset, and 26.99 dB on the DIV2K\_LSDIR\_test dataset with a common GPU (\ie NVIDIA GeForce RTX 3090 GPU). 

\medskip
\noindent{\textbf{Main Track: Overall Performance.}} The aim is to obtain a network design/solution with the best overall performance in terms of inference runtime, FLOPS, and parameters on a common GPU while being constrained to maintain or improve the threshold PSNR results.

\medskip
\noindent{\textbf{Sub-Track 1: Runtime Performance.}} The aim is to obtain a network design/solution with the lowest inference time (runtime) on a common GPU while being constrained to maintain or improve over the baseline method RLFN in terms of number of parameters, FLOPs, and the threshold PSNR result.

\medskip
\noindent{\textbf{Sub-Track 2: FLOPs Performance.}} The aim is to obtain a network design/solution with the lowest amount of FLOPs on a common GPU while being constrained to maintain or improve the inference runtime, the parameters, and the threshold PSNR results.

\medskip
\noindent{\textbf{Sub-Track 3: Parameters Performance.}} The aim is to obtain a network design/solution with the lowest amount of parameters on a common GPU while being constrained to maintain the FLOPs, the inference time (runtime), and the threshold PSNR results.

\medskip
\noindent{\textbf{Challenge phases: }}
\textit{(1) Development and validation phase}: Participants were given access to 800 LR/HR training image pairs and 200 LR/HR validation image pairs from the DIV2K and the LSDIR datasets. 
Additional 84,991 LR/HR training image pairs from the LSDIR dataset are also provided to the participants. 
The RLFN model, pre-trained parameters, and validation demo script are available on GitHub \url{https://github.com/Amazingren/NTIRE2024_ESR}, allowing participants to benchmark their models' runtime on their systems. 
Participants could upload their HR validation results to the evaluation server to calculate the PSNR of the super-resolved image produced by their models and receive immediate feedback. 
The corresponding number of parameters, FLOPs, and runtime will be computed by the participants.
\textit{(2) Testing phase}: In the final testing phase, participants were granted access to 100 LR testing images from DIV2K and 100 LR testing images from LSDIR, while the HR ground-truth images remained hidden. Participants submitted their super-resolved results to the Codalab evaluation server and emailed the code and factsheet to the organizers. 
The organizers verified and ran the provided code to obtain the final results, which were then shared with participants at the end of the challenge.

\medskip
\noindent{\textbf{Evaluation protocol: }}
Quantitative evaluation metrics included validation and testing PSNRs, runtime, FLOPs, and the number of parameters during inference. 
PSNR was measured by discarding a 4-pixel boundary around the images. 
The average runtime during inference was computed on the 200 LR validation images and the 200 LR testing images.
The average runtime on the validation and testing sets served as the final runtime indicator. 
FLOPs are evaluated on an input image of size 256$\times$256. 
Among these metrics, runtime was considered the most important. 
Participants were required to maintain a PSNR of at least 26.90 dB on the DIV2K\_LSDIR valid dataset, and 26.99 dB on the DIV2K\_LSDIR test dataset during the challenge.  
The constraint on the testing set helped prevent overfitting on the validation set.
It's important to highlight that methods with a PSNR below the specified threshold (\ie 26.90 dB on DIV2K\_LSDIR\_valid and, 26.99 dB on DIV2K\_LSDIR\_test) will not be considered for the subsequent ranking process. It is essential to meet the minimum PSNR requirement to be eligible for further evaluation and ranking.
A code example for calculating these metrics is available at \url{https://github.com/Amazingren/NTIRE2024_ESR}.

To better quantify the rankings, we have designed a scoring function for three evaluation metrics in this challenge: runtime, FLOPs, and parameters. This scoring aims to convert the performance of each metric into corresponding scores to make the rankings more significant. Especially, the score for each separate metric (\ie Runtime, FLOPs, and parameter) for each sub-track is calculated as:
\begin{equation}
    \begin{aligned}
    Score\_{Metric} = \dfrac{exp (2 \times Metric_{TeamX})}{Metric_{Baseline}},
    \end{aligned}
    \label{equ:score1}
\end{equation}
based on the score of each metric, the final score used for the main track is calculated as:
\begin{equation}
    \begin{aligned}
    Score\_{Final} & = w_{1} \times Score\_Runtime \\
    & + w_{2} \times Score\_FLOPs \\
    & + w_{3} \times Score\_Params,
    \end{aligned}
    \label{equ:score2}
\end{equation}
where $w_{1}$, $w_{2}$, and $w_{3}$ are set to 0.7, 0.15, and 0.15, respectively. This setting is intended to incentivize participants to design a method that prioritizes speed efficiency while maintaining a reasonable model complexity.
\section{Challenge Results}
\label{sec:esr_results}
\begin{table*}[!ht]
\caption{Results of Ninth NTIRE 2024 Efficient SR Challenge. The performance of the solutions is compared thoroughly from three perspectives including the runtime, FLOPs, and the number of parameters. The underscript numbers associated with each metric score denote the ranking of the solution in terms of that metric. For runtime, ``Val.'' is the runtime averaged on DIV2K\_LSDIR\_valid validation set. ``Test'' is the runtime averaged on a test set with 200 images from DIV2K\_LSDIR\_test set, respectively. ``Ave.'' is averaged on the validation and test datasets. 
``\#Params'' is the total number of parameters of a model. ``FLOPs'' denotes the floating point operations. 
Main Track combines all three evaluation metrics.
The ranking for the main track is based on the score calculated via Eq.~\ref{equ:score2}, and the ranking for other sub-tracks is based on the score of each metric score via Eq.~\ref{equ:score1}. Please note that \textbf{this is not a challenge for PSNR improvement. The ``validation/testing PSNR'' is not ranked. For all the scores, the lower, the better}.
}
\label{tbl:final_results}
\centering
\begin{threeparttable}
\resizebox{\linewidth}{!}
{
\setlength{\tabcolsep}{11pt}
\begin{tabular}{@{\extracolsep{\fill}} 
                                l|
                                |
                                S[table-format=2.2]
                                S[table-format=2.2]
                                |
                                S[table-format=2.2$_{(2)}$] 
                                S[table-format=2.2] 
                                S[table-format=2.2]
                                |
                                S[table-format=1.3$_{(2)}$] 
                                S[table-format=3.2$_{(2)}$] 
                                |
                                S[table-format=3.2$_{(2)}$] 
                                S[table-format=3.2] 
                                S[table-format=2.2]
                                |
                                S[table-format=3.2]
                                S[table-format=2]}
                                
\toprule
\multirow{2}{*}{Teams}      & \multicolumn{2}{c|}{PSNR [dB]} & \multicolumn{3}{c|}{Runtime [ms]}  & {\#Params}  &   {FLOPs}  & \multicolumn{3}{c|}{Sub-Track Scores}  &\multicolumn{2}{c}{Main-Track}   \\ \cline{2-6} \cline{9-13}
&{Val.} & {Test} & {Val.} & {Test} & {Ave.}  & {[M]} & {[G]} & {Runtime} & {\#Params} & {FLOPs} & {Score} & {Ranking}  \\ 
\midrule
XiaomiMM & 26.94 & 27.01 & 5.62 & 5.57 & 5.59 & 0.151 & 9.83 & 2.59$_{(1)}$ & 2.59$_{(7)}$ & 2.72$_{(8)}$ & 2.61 & 1 \\
Cao Group & 26.90 & 27.00 & 10.99 & 5.76 & 8.37 & 0.215 & 13.05 & 4.15$_{(2)}$ & 3.88$_{(13)}$ & 3.77$_{(13)}$ & 4.05 & 2 \\
BSR & 26.90 & 27.00 & 11.96 & 6.80 & 9.38 & 0.218 & 11.95 & 4.93$_{(3)}$ & 3.96$_{(14)}$ & 3.37$_{(10)}$ & 4.55 & 3 \\
VPEG\_O & 26.90 & 27.01 & 12.20 & 7.06 & 9.63 & 0.212 & 13.86 & 5.14$_{(4)}$ & 3.81$_{(12)}$ & 4.09$_{(14)}$ & 4.78 & 4 \\
CMVG & 26.90 & 27.01 & 12.58 & 7.46 & 10.02 & 0.202 & 12.17 & 5.49$_{(5)}$ & 3.58$_{(10)}$ & 3.45$_{(12)}$ & 4.90 & 5 \\
LeESR & 26.91 & 27.02 & 13.21 & 8.04 & 10.62 & 0.165 & 9.75 & 6.08$_{(7)}$ & 2.83$_{(9)}$ & 2.69$_{(6)}$ & 5.09 & 6 \\
AdvancedSR & 26.91 & 27.02 & 12.94 & 8.04 & 10.49 & 0.263 & 16.20 & 5.94$_{(6)}$ & 5.26$_{(17)}$ & 5.19$_{(17)}$ & 5.73 & 7 \\
ECNU\_MViC & 26.90 & 27.00 & 14.10 & 8.87 & 11.49 & 0.163 & 9.78 & 7.05$_{(10)}$ & 2.80$_{(8)}$ & 2.70$_{(7)}$ & 5.76 & 8 \\
HiSR & 26.91 & 27.03 & 13.69 & 8.83 & 11.26 & 0.208 & 11.99 & 6.78$_{(9)}$ & 3.71$_{(11)}$ & 3.38$_{(11)}$ & 5.81 & 9 \\
MViC\_SR & 26.90 & 27.00 & 14.53 & 9.23 & 11.88 & 0.138 & 8.16 & 7.53$_{(11)}$ & 2.39$_{(6)}$ & 2.29$_{(5)}$ & 5.97 & 10 \\
LVTeam & 26.91 & 27.02 & 13.79 & 8.71 & 11.25 & 0.266 & 16.34 & 6.77$_{(8)}$ & 5.36$_{(18)}$ & 5.27$_{(18)}$ & 6.33 & 11 \\
Fresh & 26.90 & 27.00 & 14.52 & 9.29 & 11.90 & 0.245 & 14.97 & 7.56$_{(12)}$ & 4.69$_{(16)}$ & 4.58$_{(16)}$ & 6.68 & 12 \\
Lanzhi & 26.93 & 27.02 & 14.86 & 9.53 & 12.19 & 0.318 & 19.70 & 7.94$_{(13)}$ & 7.44$_{(20)}$ & 7.41$_{(20)}$ & 7.79 & 13 \\
Supersr & 26.90 & 27.01 & 15.16 & 9.90 & 12.53 & 0.298 & 18.67 & 8.40$_{(15)}$ & 6.55$_{(19)}$ & 6.67$_{(19)}$ & 7.87 & 14 \\
MeowMeowMeow & 26.92 & 27.03 & 16.18 & 10.95 & 13.57 & 0.238 & 14.47 & 10.03$_{(16)}$ & 4.49$_{(15)}$ & 4.35$_{(15)}$ & 8.35 & 15 \\
Just Try & 26.90 & 27.00 & 15.23 & 9.75 & 12.49 & 0.380 & 24.81 & 8.35$_{(14)}$ & 11.00$_{(22)}$ & 12.46$_{(23)}$ & 9.36 & 16 \\
VPEG\_C & 26.90 & 27.03 & 18.76 & 13.31 & 16.03 & 0.084 & 4.97 & 15.24$_{(17)}$ & 1.70$_{(3)}$ & 1.66$_{(2)}$ & 11.17 & 17 \\
VPEG\_E & 26.90 & 27.01 & 21.21 & 15.74 & 18.48 & 0.093 & 5.89 & 23.09$_{(20)}$ & 1.80$_{(5)}$ & 1.82$_{(4)}$ & 16.71 & 18 \\
BU-ESR & 27.00 & 27.11 & 19.47 & 14.07 & 16.77 & 0.433 & 27.05 & 17.28$_{(18)}$ & 15.36$_{(24)}$ & 15.65$_{(24)}$ & 16.75 & 19 \\
Lasagna & 26.90 & 27.00 & 20.11 & 14.59 & 17.35 & 0.657 & 41.21 & 19.08$_{(19)}$ & 63.12$_{(25)}$ & 66.03$_{(26)}$ & 32.73 & 20 \\
ZHEstar & 26.93 & 27.04 & 31.07 & 24.66 & 27.87 & 0.090 & 5.81 & 113.87$_{(21)}$ & 1.76$_{(4)}$ & 1.81$_{(3)}$ & 80.25 & 21 \\
BlingBling & 26.93 & 27.04 & 40.75 & 34.63 & 37.69 & 0.424 & 20.17 & 604.72$_{(22)}$ & 14.51$_{(23)}$ & 7.77$_{(21)}$ & 426.64 & 22 \\
PiXupt & 26.91 & 27.00 & 52.91 & 44.21 & 48.56 & 0.060 & 9.84 & 383.46$e1_{(23)}$ & 1.46$_{(1)}$ & 2.72$_{(9)}$ & 268.49$e1$ & 23 \\
Minimalist & 26.91 & 27.02 & 53.37 & 47.02 & 50.20 & 0.346 & 20.65 & 506.53$e1_{(24)}$ & 8.87$_{(21)}$ & 8.16$_{(22)}$ & 354.82$e1$ & 24 \\
XJU\_100th Ann & 26.90 & 27.02 & 61.89 & 55.78 & 58.84 & 0.069 & 4.39 & 219.74$e2_{(25)}$ & 1.55$_{(2)}$ & 1.56$_{(1)}$ & 153.82$e2$ & 25 \\
MagicSR & 27.03 & 27.17 & 461.32 & 443.38 & 452.35 & 1.019 & 38.47 & 241.07$e31_{(26)}$ & 619.57$_{(26)}$ & 49.98$_{(25)}$ & 168.75$e31$ & 26 \\

\midrule

\multicolumn{13}{c}{The following methods are not ranked since their validation/testing PSNR (underlined) is not on par with the threshold.}\\ \midrule
FireWork & \underline{26.86} & \underline{26.96} & 13.26 & 7.99 & 10.62 & 0.441 & 28.85 & 6.08 & 16.16 & 18.79 & 9.50 & \\
DIRN & \underline{26.22} & \underline{26.33} & 21.45 & 15.96 & 18.71 & 0.097 & 6.48 & 24.01 & 1.84 & 1.93 & 17.38 & \\
VIP & \underline{26.89} & 27.02 & 25.60 & 19.87 & 22.73 & 0.088 & 5.56 & 47.61 & 1.74 & 1.76 & 33.85 &  \\
DASE-IDEALab & \underline{26.89} & 27.00 & 38.95 & 33.42 & 36.19 & 0.088 & 4.50 & 468.34 & 1.74 & 1.58 & 328.34 &  \\
ACVLAB & \underline{25.31} & \underline{25.45} & 206.29 & 196.24 & 201.26 & 0.753 & 55.89 & 712.28$e12$ & 115.68 & 293.76 & 499.60$e12$ &  \\
SPAN-T & 26.92 & \underline{26.98} & 5.91 & 6.04 & 5.98 & 0.131 & 8.54 & 2.76 & 2.29 & 2.38 & 2.63 & \\
hajnal & 27.00 & \underline{26.95} & 13.40 & 7.94 & 10.67 & 0.243 & 14.89 & 6.13 & 4.63 & 4.54 & 5.67 &  \\
\makecell[l]{KLETech \\ CEVI\_Lowlight\_Hypnotise} & \underline{24.53} & \underline{24.75} & 315.17 & 302.12 & 308.64 & 16.698 & 1174.94 & 597.94$e20$ & 566.23$e43$ & 763.95$e49$ & 114.59$e49$ &  \\

\midrule
RLFN (Baseline) &	26.96&	27.07&	14.35&	9.19&	11.77&	0.317& 	19.67 & 7.39 & 7.39 & 7.39 & 7.39 & \\
\bottomrule
\end{tabular}
}
\end{threeparttable}
\end{table*}
The final test results and rankings are presented in Table ~\ref{tbl:final_results}\footnote{Please note that we have corrected the order of "\#Params" and "FLOPs" in Table~\ref{tbl:final_results}. We apologize for the mistake.}. 
The table also includes the baseline method RLFN~\cite{kong2022residual} for comparison. 
In Sec.\ref{sec:methods_and_teams}, the methods evaluated in Table \ref{tbl:final_results} are briefly explained, while the team members are listed in \ref{sec:teams}. 
The performance of different methods is compared from four different perspectives including the runtime, FLOPs, the parameters, and the overall performance.
Furthermore, in order to promote a fair competition emphasizing efficiency, the criteria for image reconstruction quality in terms of test PSNR are set to 26.90 and 26.99 on the DIV2K\_LSDIR\_valid and DIV2K\_LSDIR\_test sets, resp.

\noindent\textbf{Runtime.} In this challenge, runtime stands as the paramount evaluation metric. XiaomiMM's solution emerges as the frontrunner with the shortest runtime among all entries in the efficient SR challenge, securing the top position. Following closely, the Cao Group and BSR claim the second and third spots, respectively. Remarkably, the average runtime of the top three solutions on both the validation and test sets remains below 10 ms. Impressively, the first 15 teams present solutions with an average runtime below 13 ms, showcasing a continuous enhancement in the efficiency of image SR networks. Despite the slight differences in runtime among the top three teams, the challenge retains its competitive edge. Furthermore, the XiaomiMM team achieves the highest PSNR on both the validation and test sets among the top three teams.

\noindent\textbf{FLOPs.} FLOPs, representing the number of floating-point operations, serve as a critical metric for assessing model complexity. 
In this sub-track, PiXupt secures the top position, followed by XJU\_100 th Ann and VPEG\_C in second and third places, respectively. 
Remarkably, the disparity among the top three methods is minimal, underscoring their competitiveness and proficiency in managing model complexity. However, it's noteworthy that both PiXupt and XJU\_100th Ann exhibit relatively high runtimes. Further exploration is warranted to address and mitigate such challenges.

\noindent\textbf{Parameters.} Parameters serve as another critical metric for assessing model complexity, which was also evaluated in this challenge. As shown in Table~\ref{tbl:final_results}, XJU\_100th Ann, VPEG\_C, and ZHEstar secured the first three places. However, akin to the FLOPs sub-track, the runtime of these methods lags significantly behind that of the other methods. Further exploration is warranted to address and mitigate such challenges.

\noindent\textbf{Overall evaluation.} In the final assessment, performance is meticulously evaluated based on an aggregate metric that intricately weaves together runtime, FLOPs, and the number of parameters. Notably, the XiaomiMM Group emerges triumphant, securing the coveted top spot under this comprehensive metric, with the Cao Group and BSR clinching the 2nd and 3rd places, respectively. This outcome underscores the meticulous craftsmanship and ingenuity embedded within their methodologies. With runtime bearing significant weight in the scoring system, it's noteworthy how closely the overall performance of each method mirrors their rankings in the runtime sub-track, reflecting the discernible impact of efficiency optimization on overall success. Indeed, our overarching objective this year has been to incentivize participants to embark on a quest for speed and efficiency in their designs—a mission that has undoubtedly borne fruit in the form of groundbreaking advancements and innovative solutions.

\noindent\textbf{PSNR.} While MagicSR, BU-ESR, ZHEstar, and BingBing showcase impressive PSNR performance—traditionally regarded as a cornerstone metric in method evaluation—MagicSR notably attains a remarkable 27.17 dB, closely followed by BU-ESR at 27.11 dB, and ZHEstar and BingBing both achieving 27.04 dB on the DIV2K\_LSDIR\_test set. 
However, amidst these accolades, it is paramount to underscore the overarching focus of this challenge: efficiency in super-resolution. 
Thus, in alignment with this objective, we opted to relax the PSNR threshold to a stringent lower bound of 26.90 and 26.99 for ranking on both the DIV2K\_LSDIR\_valid and DIV2K\_LSDIR\_test sets. 
This strategic adjustment aims to underscore the importance of balancing performance with efficiency. Notably, a total of 26 teams successfully met this revised requirement, showcasing their adeptness in navigating the delicate equilibrium between quality and computational efficiency. 
While several teams, such as SPAN-T, hajnal, FireWork, DIRN, and VIP, exhibit commendable efficiency performance, it is lamentable that they fell short of meeting the PSNR threshold, highlighting the multifaceted challenges inherent in the pursuit of efficient super-resolution excellence.

\subsection{Main Ideas}
Throughout this challenge, several techniques have been proposed to enhance the efficiency of deep neural networks for image super-resolution (SR) while striving to maintain optimal performance. The choice of techniques largely depends on the specific metrics that a team aims to optimize. Below, we outline some typical ideas that have emerged:

\begin{itemize}
    \item \textbf{Parameter-free attention mechanism is validated as a useful technique to enhance computational efficiency}~\cite{du2022parameter,yang2021simam}. Specifically, XiaomiMM proposed a swift parameter-free attention network based on parameter-free attention, which achieves the lowest runtime while maintaining a decent PSNR performance.
    \item \textbf{Re-parameterization}~\cite{ding2021repvgg} ~\cite{du2022parameter,yang2021simam} \textbf{is commonly used in this challenge}. Usually, a normal convolutional layer with multiple basic operations (3 × 3 convolution, 1 × 1 operation, first and second-order derivative operators, skip connections) is parameterized during network training. During inference, the multiple operations that reparameterize a convolution could be merged back into a single convolution. \eg All the top three teams (\ie XiaomiMM, Cao Group, and BSR) used this operation in their solutions.
    \item \textbf{Incorporating multi-scale information and hierarchical module design are proven strategies for effectively fusing critical information}. For instance, solutions such as PiXupt, XJU\_100th Ann, and ZHEstar have successfully utilized multi-scale residual connections and hierarchical module designs to enhance their performance.
    \item \textbf{Network pruning plays an important role}. It is observed that a couple of teams (\ie CMVG, AdvancedSR, and HiSR) used network pruning techniques to slightly compress a network. This leads to more lightweight architecture without heavy performance drop. 
    \item \textbf{Exploration with new network architectures is conducted.} Besides the common CNN or Transformers, the state space model (\ie vision mamba~\cite{mamba}) was tried by BlingBling for the first time in this challenge.
    \item \textbf{Various other techniques are also attempted}. Some teams also proposed solutions based on neural architecture search, vision transformers, and advanced training strategies.
\end{itemize}

\subsection{Fairness}
To uphold the integrity and fairness of the efficient SR challenge, a series of rules were meticulously crafted, primarily focusing on the dataset utilized for training the network. 
Firstly, participants were granted permission to train their models with additional external datasets, such as Flickr2K, thereby fostering a diverse and comprehensive training regimen. 
However, to ensure unbiased evaluation, the use of the additional DIV2K and LSDIR validation sets comprising either high-resolution (HR) or low-resolution (LR) images was strictly prohibited during training. 
This measure was implemented to preserve the integrity of the validation set, which served as a crucial yardstick for assessing the overall performance and generalizability of the proposed networks. Furthermore, training with DIV2K and LSDIR test LR images was unequivocally forbidden, safeguarding the sanctity of the test dataset and maintaining the sanctity of the evaluation process.
Lastly, employing advanced data augmentation strategies during training was deemed a fair and equitable approach, empowering participants to optimize their models while adhering to the established guidelines and regulations.

\subsection{Conclusions}
Several conclusions can be drawn from the analysis of different solutions as follows.
Firstly, The competition for the efficient image SR community is still fierce. This
year the challenge had 262 registered participants, and 34 teams made valid submissions. All the proposed methods improve the state-of-the-art for efficient SR. 
Secondly, re-parameterization and network compression play an important role in efficient SR. More exploration is still encouraged to further improve the model's efficiency with these techniques.
Thirdly, regarding the training, the adoption of large-scale dataset~\cite{lilsdir} for pre-training improves the accuracy of the network and for most of the methods, the training of the network proceeds in several phases with increased patch size and reduced learning rate. 
Fourthly, the state space model was explored for the first time in the challenge, which may draw a new model choice for the upcoming works.
Finally, by jointly considering runtime, FLOPs, and the number of parameters, it is possible to design a balanced model that optimizes more than one evaluation metric. 
Finally, as computational capabilities advance, the optimization of models for runtime, FLOPs, and parameter count will become increasingly crucial. With ongoing developments in hardware and algorithmic efficiency, there is a strong likelihood of even more sophisticated and resource-efficient models emerging in the field of super-resolution (SR). We anticipate that the pursuit of efficiency in SR will persist and be further explored, leading to continued progress and innovation in the field.
\section{Challenge Methods and Teams}
\label{sec:methods_and_teams}

\subsection{XiaomiMM}
\noindent\textbf{Method.}
The authors propose the Swift Parameter-free Attention Network (SPAN)~\cite{wan2023swift}, a highly efficient SISR model that balances parameter count, inference speed, and image quality. 
\begin{figure*}[ht]
\centering
\vspace{-1.5em} 
\includegraphics[width=\linewidth]{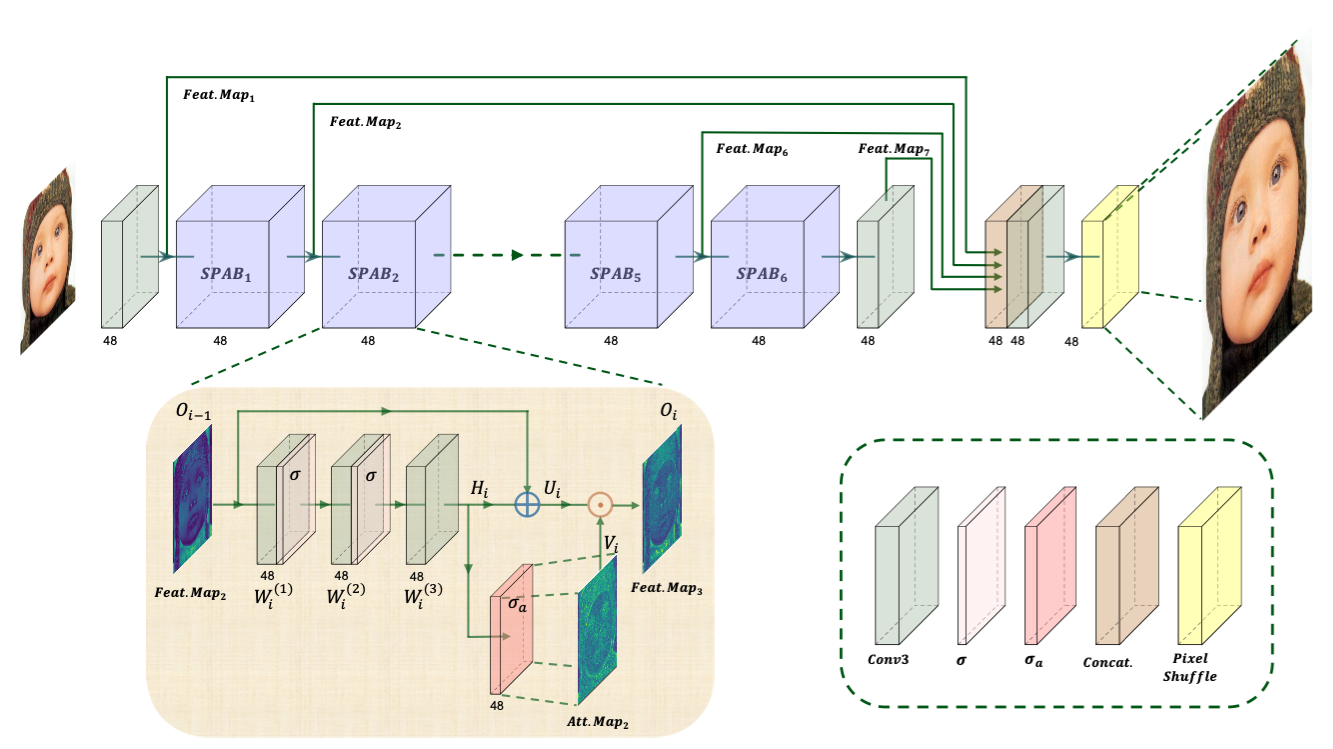}
\caption{\textit{Team XiaomiMM:} The proposed SPAN architecture~\cite{wan2023swift}. The brown area indicates the internal structure of each SPAB module. $Att. Map_2$ denotes the generated attention map.} 
\label{fig:pipeline_team38}
\end{figure*}
As shown in Figure \ref{fig:pipeline_team38}, SPAN consists of 6 consecutive SPABs and each SPAB block extracts progressively higher-level features sequentially through three convolutional layers with $C'$-channeled $H' \times W'$-sized kernels (In our model, they choose $H' = W' = 3$). The extracted features $H_i$ are then added with a residual connection from the input of SPAB, forming the pre-attention feature map $U_i$ for that block. The features extracted by the convolutional layers are passed through an activation function $\sigma_a(\cdot)$ that is symmetric about the origin to obtain the attention map $V_i$. The feature map and attention map are element-wise multiplied to produce the final output $O_i=U_i \odot V_i$ of the SPAB block, where $\odot$ denotes element-wise multiplication. They use $W_i^{(j)}\in R^{C' \times H'\times W'}$ to represent the kernel of the $j$-th convolutional layer of the $i$-th SPAB block and $\sigma$ to represent the activation function following the convolutional layer.  Then the SPAB block can be expressed as:
\begin{equation}
    \begin{aligned}
        O_i&=F_{W_i}^{(i)}(O_{i-1})=U_i \odot V_i,\\
        U_i&=O_{i-1}\oplus H_i, \quad V_i=\sigma_a(H_i),\\ 
        H_i&=F_{c,W_i}^{(i)}(O_{i-1}),\\
        &=W_i^{(3)}\otimes\sigma(W_i^{(2)}\otimes\sigma(W_i^{(1)}\otimes O_{i-1})),
    \end{aligned}
    \label{equ:SPAB}
\end{equation}
where $\oplus$ and $\otimes$ represent the element-wise sum between extracted features and residual connections,  and the convolution operation, respectively.  $F_{W_i}^{(i)}$ and $F_{c,W_i}^{(i)}$ are the function representing the $i$-th SPAB and the function representing the $3$ convolution layers of $i$-th SPAB with parameters   $W_i=(W_i^{(1)},W_i^{(2)},W_i^{(3)})$, respectively. $O_0=\sigma(W_0\otimes I_{\text{LR}})$ is a $C'$-channeled $H \times W$ feature map from the $C$-channeled $H \times W$-sized low-resolution input image $I_{LR}$ undergone a convolutional layer with $3 \times 3$ sized kernel $W_0$.  This convolutional layer ensures that each SPAB has the same number of channels as input.  The whole SPAN neural network can be described as
\begin{equation}
    \begin{aligned}
        I_{\text{HR}}&=F(I_{\text{LR}})=\text{PixelShuffle}[W_{f2}\otimes O)],\\
        O&=\text{Concat}(O_0,O_1,O_5,W_{f1}\otimes O_6),
    \end{aligned}
    \label{equ:SPAN}
\end{equation}
where $O$ is a $4C'$-channeled $H \times W$-sized feature map with multiple hierarchical features obtaining by concatenating $O_0$ with the outputs of the first, fifth, and the convolved output of the sixth SPAB blocks by $C'$-channeled $3 \times 3$-sized kernel $W_{f1}$.  $O$ is processed through a $3\times 3$ convolutional layer to create an $r^2C$ channel feature map of size $H\times W$. Then, this feature map goes through a pixel shuffle module to generate a high-resolution image of $C$ channels and dimensions $rH\times rW$, where $r$ represents the super-resolution factor. The idea of computing attention maps directly without parameters from features extracted by convolutional layers led to two design considerations for our neural network: the choice of activation function for computing the attention map and the use of residual connections, more details about activation function and SPAB module are in~\cite{wan2023swift}.

\noindent\textbf{Training Details.}
The dataset utilized for training comprises DIV2K and LSDIR. During each training batch, 64 HR RGB patches are cropped, measuring $256 \times 256$, and subjected to random flipping and rotation. The learning rate is initialized at $5\times 10^{-4}$ and undergoes a halving process every $2\times10^5$ iterations. The network undergoes training for a total of $10^6$ iterations, with the L1 loss function being minimized through the utilization of the Adam optimizer~\cite{kingma2014adam}. They repeated the aforementioned training settings four times after loading the trained weights. Subsequently, fine-tuning is executed using the L1 and L2 loss functions, with an initial learning rate of $1\times10^{-5}$ for $5\times10^5$ iterations, and an HR patch size of 512. They conducted the finetuning on four models utilizing both L1 and L2 losses, employed batch sizes of 64 and 128, and integrated these four models to obtain the ultimate model.
\subsection{Cao Group}
\begin{figure}
  \centering
  \includegraphics[width=0.45\textwidth]{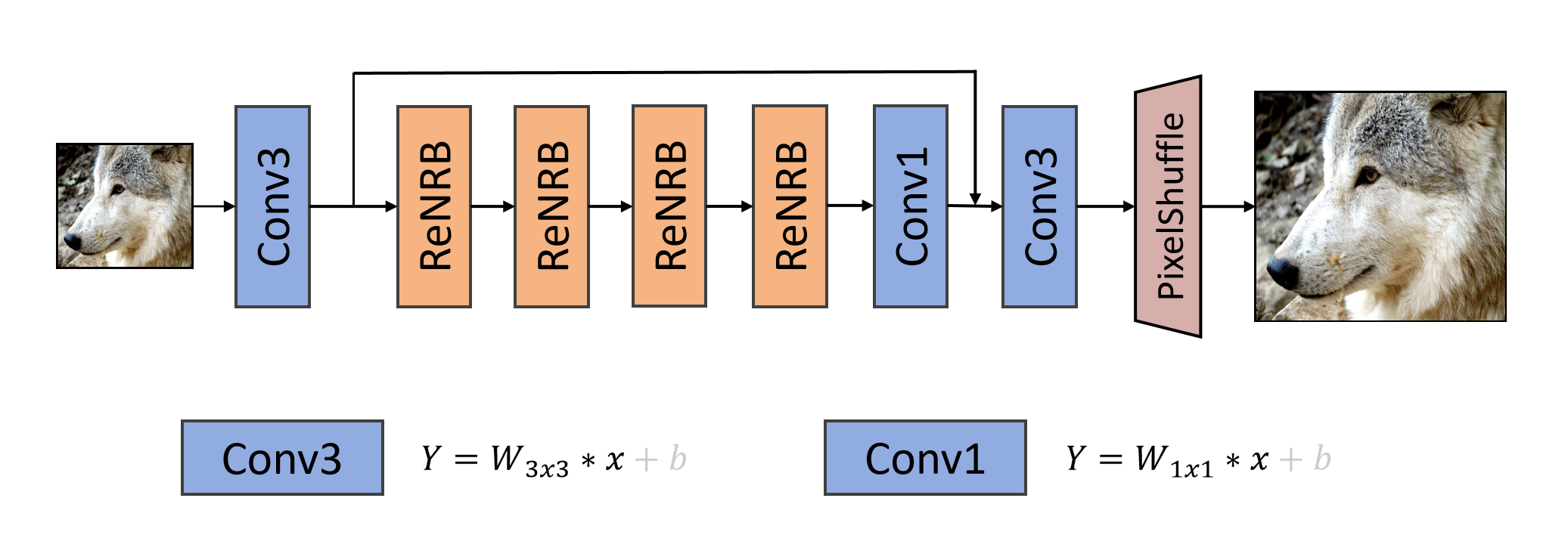} 
  \caption{\textit{Team Cao Group:} The structure of R2Net} 
  \label{R2Net} 
\end{figure}
\begin{figure}
  \centering
  \includegraphics[width=0.45\textwidth]{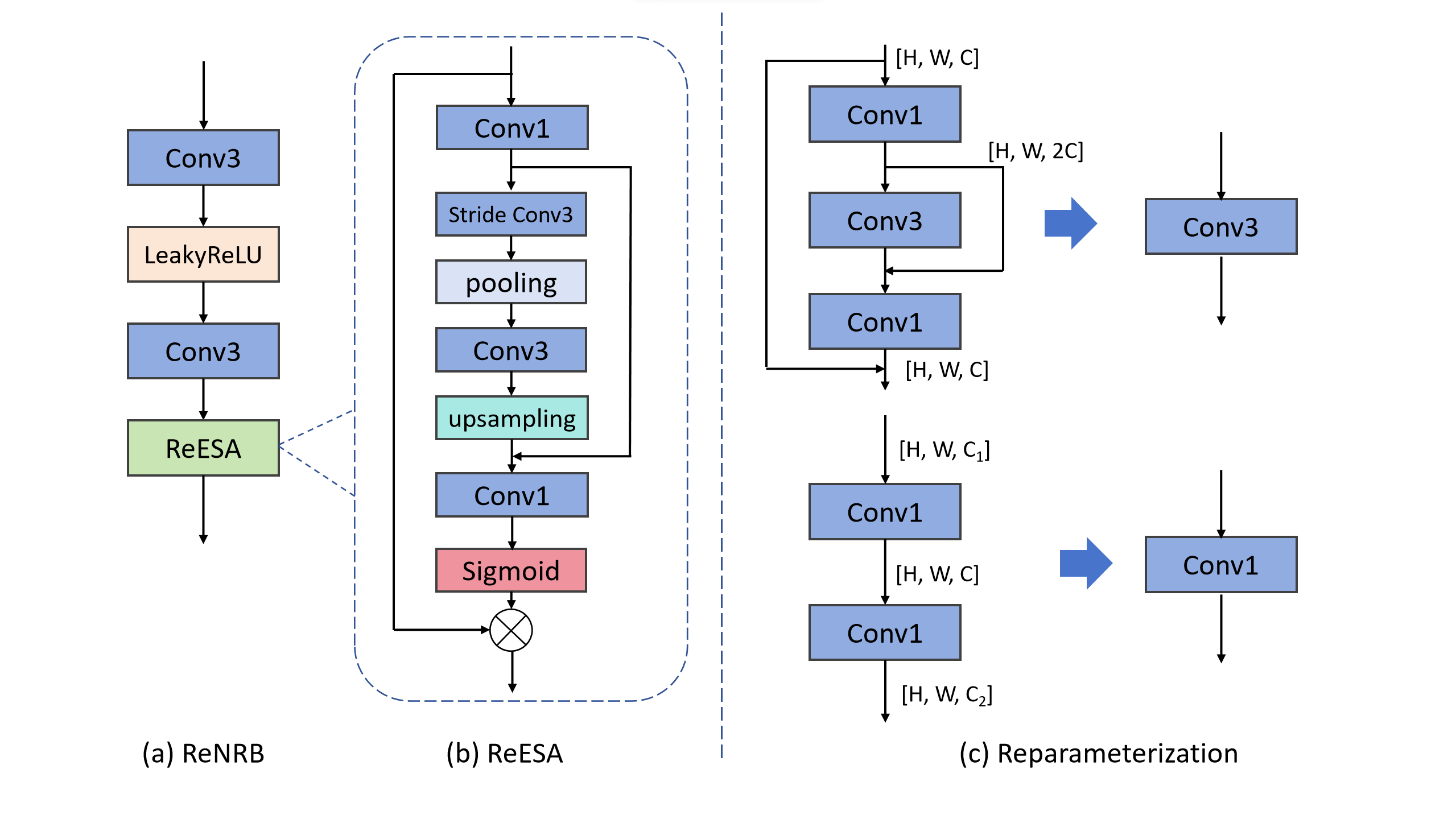} 
  \caption{\textit{Team Cao Group:} The structure of ReNRB and reparameterization in R2Net} 
  \label{R2Net block} 
\end{figure}

\textbf{Method.}
The overall architecture of their network is shown in Figure~\ref{R2Net},~which is inspired by previous leading methods, DIPNet~\cite{yu2023dipnet} and SRN~\cite{wang2023single}. They propose a Double Reparameterization Network (R2Net). Specifically, they build upon the SRN framework by combining Residual Blocks (RB) and Enhanced Spatial Attention (ESA) to form a new feature extraction module
with reparameterization, named reNRB, as shown in Figure ~\ref{R2Net block}. They remove the residual connections within RB and the 1x1 convolutions on the residual connections in ESA, retaining only the global residual connections. Furthermore, different from SRN~\cite{wang2023single}, they find that preserving the last convolution before the global residual connection significantly boosts performance. For the sake of lightweight design, they set its kernel size to 1x1. Notably, they eliminate biases in all convolutional layers, as this not only significantly accelerates inference speed and reduces parameters but also slightly enhances network performance. 

As for the reparameterization technique, they adopt RRRB~\cite{du2022fast} for all 3x3 convolutions. And inspired by the Feed Forward Network (FFN) module structure in transformer, they propose a reparameterization module to enhance the performance of 1x1 convolutions by expanding the intermediate channel, thereby harnessing the representation capability of complex structures during optimization, as shown in Figure~\ref{R2Net block}(c).

\noindent\textbf{Training Details. }
They train the network on RGB channels and augment the training data with random flipping and rotation. The number of ESA channels is set to 16, while the number of feature channels is set to 48. Following previous methods, the training process is divided into three stages:

\indent1. In the first stage, they randomly crop 256x256 HR image patches from ground truth images, with a batch size of 32. They use the Adam optimizer, setting $\beta$1 = 0.9 and $\beta$2 = 0.999, and minimize the Charbonnier loss function. The initial learning rate is set to 2e-4, with a cosine learning rate decay strategy. The number of iterations is set to 2e-6.

\indent2. In the second stage, they increase the size of the HR image patches to 512x512, with other settings remaining the same as in the first stage.

\indent3. In the third stage, the batch size is set to 64, and L2 loss is adopted to optimize over 2e-6 iterations. The initial learning rate is set to 2e-5.

Throughout the entire training process, they employ an Exponential Moving Average (EMA) strategy to enhance the robustness of training.
\subsection{BSR}


\begin{figure}
\centering
\includegraphics[width=0.47\textwidth]{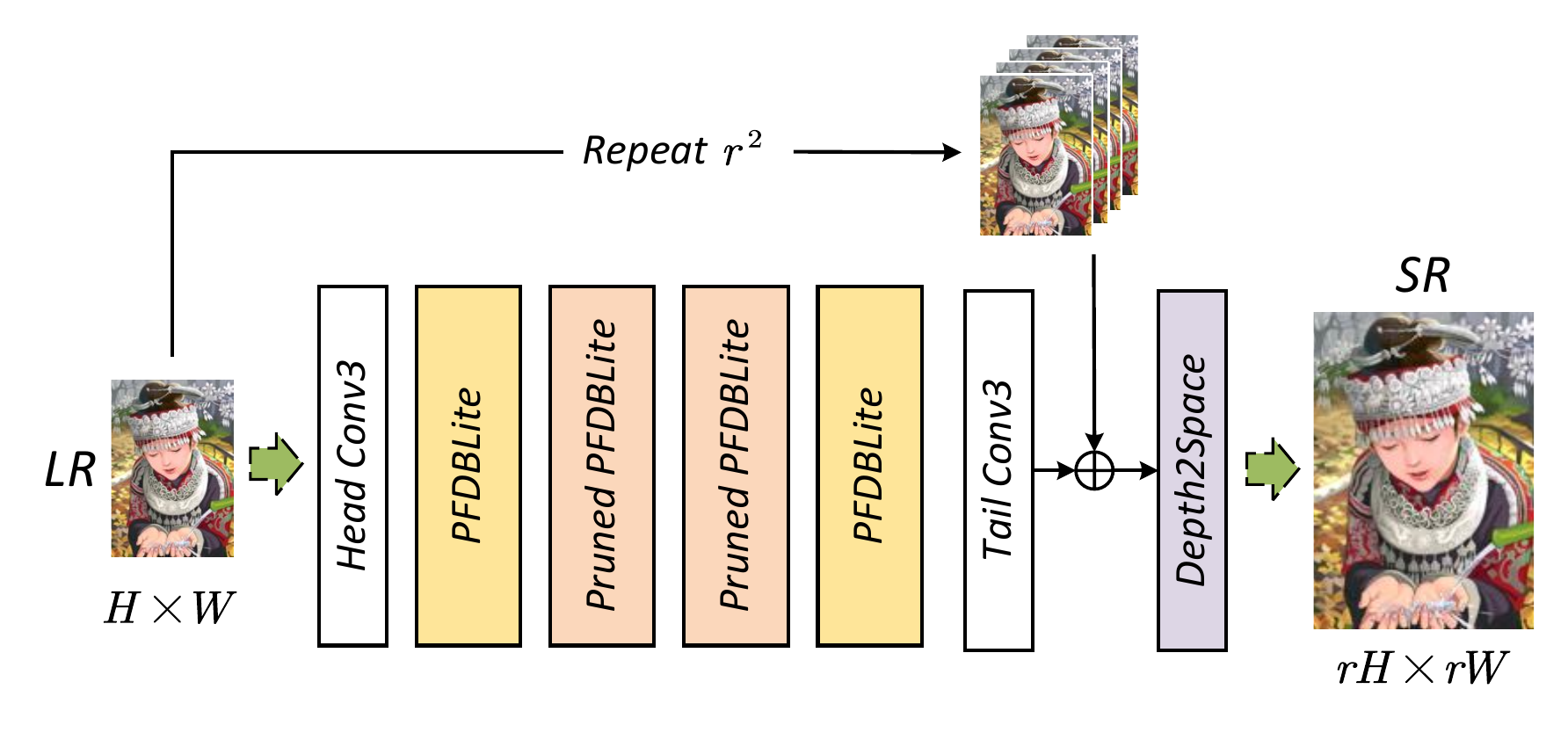}
\caption{\textit{Team BSR:} PFDNLite Architecture.} 
\label{pfdnlite}
\end{figure}

\begin{figure*}
\centering
\includegraphics[width=0.9\linewidth]{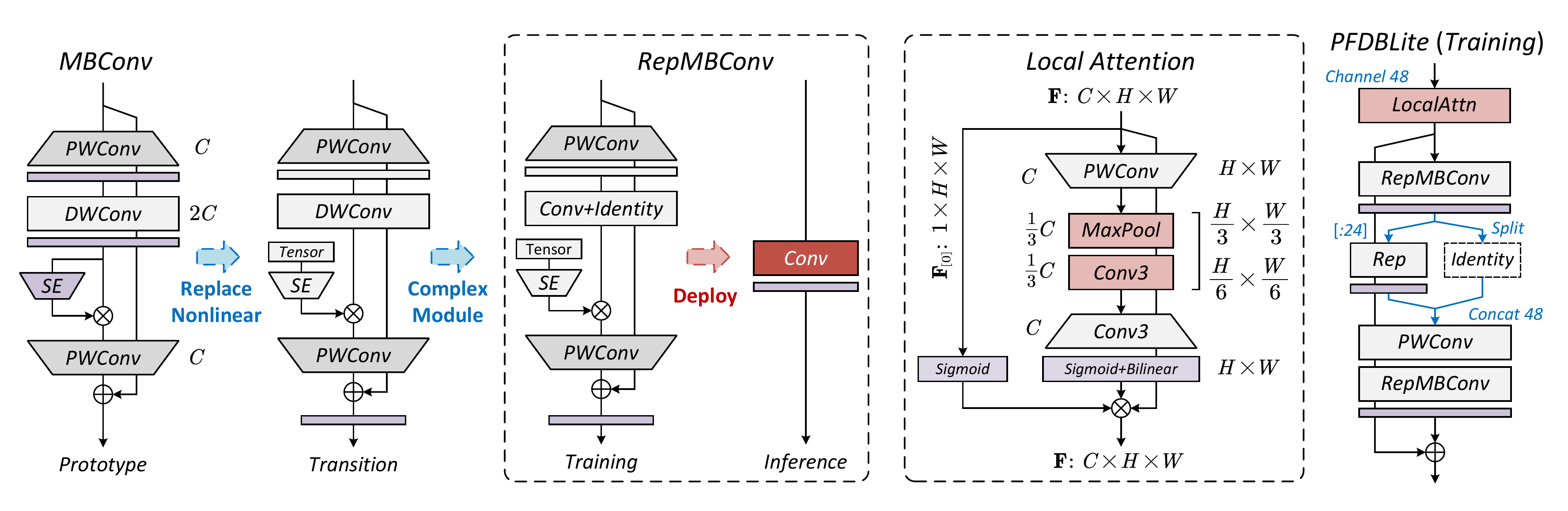}
\caption{\textit{Team BSR:} PFDBLite Block.} 
\label{pfdblite}
\end{figure*}

\textbf{Method.} Inspired by ABPN~\cite{ABPN} and  PFDN~\cite{li2023ntire_esr}, the PFDNLite, as shown in~\cref{pfdnlite}, consists of two PFDBLite blocks and two pruned-PFDBLite blocks. For $\times$4 SR, the input image is repeated $r^2$ times and then added to the final feature. Based on EFDN~\cite{EFDN} and PFDN~\cite{li2023ntire_esr}, they modify a more lightweight partial feature distillation block, dubbed PFDBLite to chase faster feature extraction. Generally, they execute two modifications focusing on the reparameterizable convolution and attention module. For the convolution block, they employ RepMBConv, which squeezes the MobileNetv3 block into vanilla convolution for better trade-offs between performance and memory access. Moreover, they add a reparameterizable point-wise convolution to cooperate with the middle RepMBConv as an approximation of partial convolution~\cite{chen2023run}. For the attention module, they propose a local attention (LocalAttn), which applies a local gate and MaxPool-based importance map to modulate input features. As illustrated in~\cref{pfdblite}, they provide the details of RepMBConv and LocalAttn.

\begin{figure}
\centering
\includegraphics[width=0.5\textwidth]{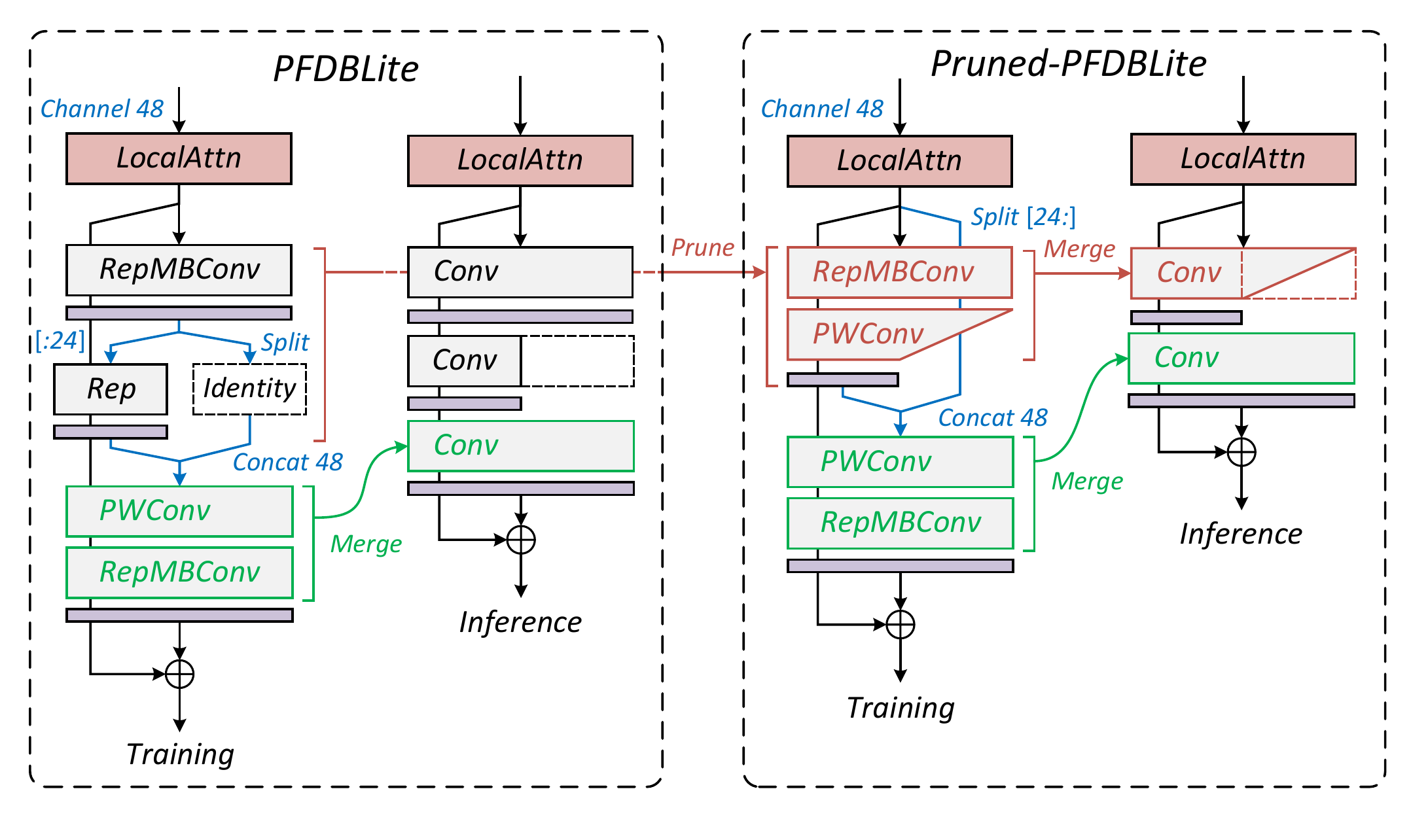}
\caption{\textit{Team BSR:} Pruned PFDBLite Block.} 
\label{prunedpfdblite}
\end{figure}

Additionally, as exhibited in \cref{prunedpfdblite}, the Pruned-PFDBLite is similar to PFDBLite but drops the second RepMBConv of PFDBLite and decreases the output channels of the first RepMBConv from 48 to 24. Besides, they add the point-wise convolution after the first RepMBConv.

\noindent\textbf{Training details.} The training process contains two stages with four steps. And the training dataset is the DIV2K\_LSDIR\_train~\cite{agustsson2017ntire}.

I. At the first stage, they only use PFDBLite blocks in the PFDNLite. 
\begin{itemize}
\item Step1. HR patches of size 256$\times$256 are randomly cropped from HR images, and the mini-batch size is set to 96. L1 loss with AdamW optimizer is used and the initial learning rate is set to 0.0005 and halved at every 100k iterations. The total iterations is 500$k$.

\item Step2. HR patches of size 256$\times$256 are randomly cropped from HR images, and the mini-batch size is set to 96. Charbonieer loss with AdamW optimizer is used and the initial learning rate is set to 0.0003 and halved at every 100k iterations. The total iterations is 500$k$.

\item Step3. HR patches of size 480$\times$480 are randomly cropped from HR images, and the mini-batch size is set to 64. MSE loss with AdamW optimizer is used and the initial learning rate is set to 0.0001 and halved at every 100k iterations. The total iterations is 500$k$.
\end{itemize}

II. At the second stage, they replace the second and the third PFDBLite block with Pruned-PFDBLite and use the weight of PFDBLite to initialize Pruned-PFDBLite.

\begin{itemize}

\item Step4. HR patches of size 480x480 are randomly cropped from HR images, and the mini-batch size is set to 64. MSE loss with AdamW optimizer is used and the initial learning rate is set to 0.0001 and halved at every 100k iterations. The total iterations is 500k.
\end{itemize}

\subsection{PiXupt}
\begin{figure}
    \centering
    \includegraphics[width=0.49\textwidth]{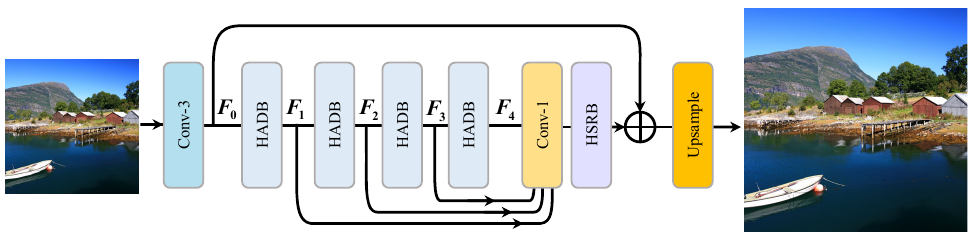}
    \caption{\textit{Team PiXupt:} The whole framework of Hierarchical Attention Residual Network (HARN)}
    \label{fig:harn}
\end{figure}

\begin{figure}
    \centering
    \includegraphics[width=0.49\textwidth]{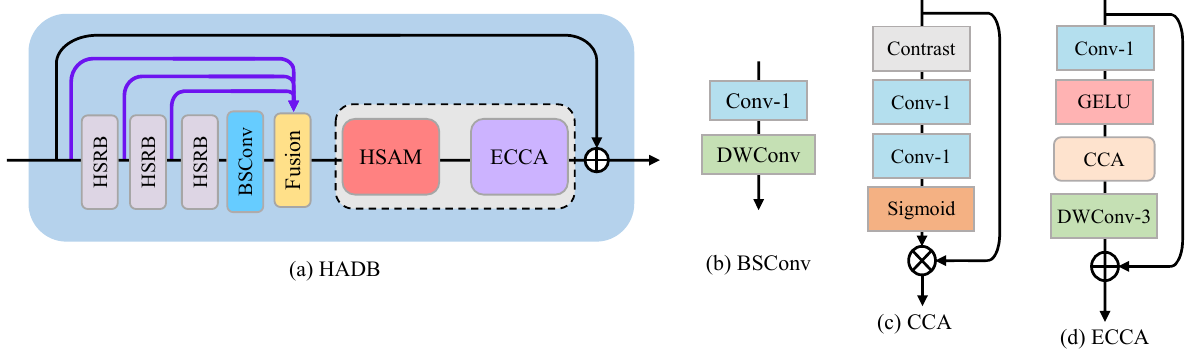}
    \caption{\textit{Team PiXupt:} Hierarchical Attention Distillation Block (HADB).}
    \label{fig:hadb}
\end{figure}

\begin{figure}
    \centering
\includegraphics[width=0.32\textwidth]{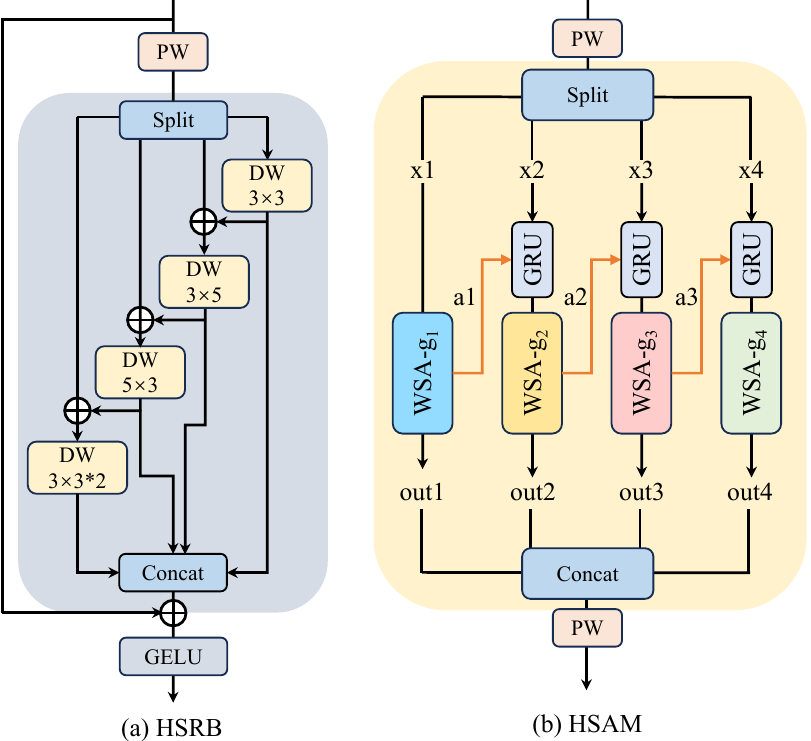}
    \caption{\textit{Team PiXupt:} (a) Hierarchical Separable Residual Block (HSRB). (b) Hierarchical Self-Attention Module (HSAM).}
    \label{fig:HSAM_HSRB}
\end{figure}

\noindent\textbf{Method.} The PiXupt team proposed the Hierarchical Attention Residual Network (HARN), as shown in \cref{fig:harn}. HARN adopts a similar basic framework to \cite{IMDN, liu2020residual, li2022blueprint, MDRN}. However, HARN uses the Hierarchical Self-Attention Module (HSAM) instead of the original spatial attention in \cite{MDRN}, and uses the Hierarchical Separable Residual Block (HSRB) instead of the original Blueprint Shallow Residual Block (BSRB) in \cite{MDRN, li2022blueprint}. As shown in the \cref{fig:HSAM_HSRB} (b), HSAM first divides the inputs into four groups from the channel dimension, where the first group performs the self-attention calculation within a large window, and then the output features are fused as hidden state with the inputs of the next group using a Gate Recurrent Unit\cite{GRU} (GRU). Such an approach allows feature information to be shared between different groups, which can help different groups to use the large window information without having to use the large window for each head. Secondly, as shown in \cref{fig:HSAM_HSRB} (a), the proposed HSRB improves on the depth-wise convolution used in the BSRB. HSRB first fuses the channel information of the input features using point-wise convolution, then groups the features and uses different sizes of depth-wise convolution kernels for different groups and connects them using a hierarchical structure to extract richer local features. HSAM and HSRB are contained in the basic module of HARN, Hierarchical Attention Distillation Block (HADB), as shown in \cref{fig:hadb}.


\noindent\textbf{Training Details.} The proposed HARN has 4 HADBs, in which the number of feature channels is set to 20. The details of training steps are as follows:
\begin{enumerate}
    \item Pretraining on DIV2K\cite{DIV2K}: HR patches of size 256 × 256 are randomly cropped from HR images, and the mini-batch size is set to 64. The model is trained by minimizing L1 loss function with Adam optimizer. The initial learning rate is set to $2 \times {10^{ - 3}}$ and halved at $\left\{ {100k, 500k, 800k,900k,950k} \right\}$-iteration. The total number of iterations is 1000$k$.
    \item Finetuning on 800 images of DIV2K and the first 10k images of LSDIR. HR patch size and  mini-batch size are set to 384 × 384 and 32, respectively. The model is fine-tuned by minimizing Charbonnier loss function. The initial learning rate is set to $5 \times {10^{ - 4}}$ and halved at $\left\{ {100k, 500k, 800k,900k,950k} \right\}$-iteration. The total number of iterations is 1000$k$.
    \item  Finetuning on 800 images of DIV2K and the first 10k images of LSDIR again. HR patch size and the mini-batch size are set to 384 × 384 and 32, respectively. The model is fine-tuned by minimizing the L2 loss function. The initial learning rate is set to $2 \times {10^{ - 4}}$ and halved at $\left\{ {100k, 300k, 600k} \right\}$-iteration. The total number of iterations is 650$k$.
\end{enumerate}
\subsection{XJU\_100th Ann}

\begin{figure*}[!ht]
\centering
\includegraphics[width=0.7\linewidth]{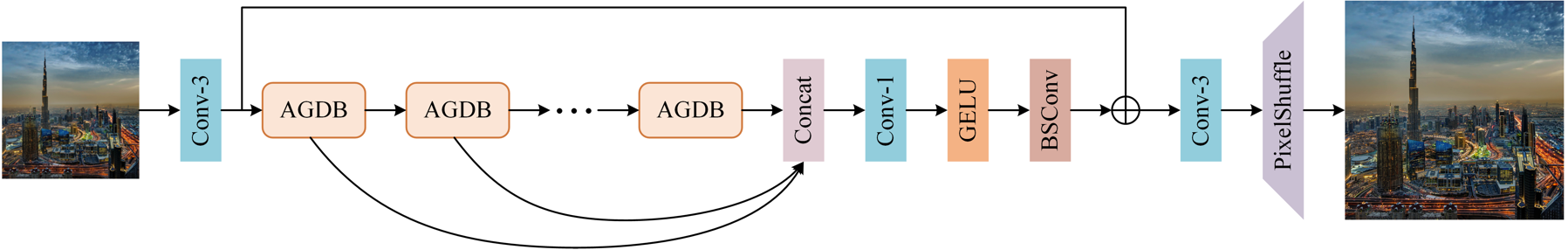}
\caption{\textit{Team XJU\_100th Ann: }The overall architecture of attention guidance distillation network (AGDN).}
\label{Figure AGDN}
\end{figure*}

\noindent\textbf{Method.}
They propose an attention guidance distillation network (AGDN) for efficient image super-resolution, which is influenced by existing studies such as IMDN\cite{IMDN}, RFDN\cite{liu2020residual}, BSRN\cite{li2022blueprint}, and MDRN\cite{MDRN}, and further improved based on these studies. Fig.\ref{Figure AGDN} illustrates the overall architecture of their network, which has been extensively validated in previous studies.

They have reconsidered the previous network structure, and the feature extraction phase remains a key limiting factor for network performance. The feature distillation block comprises distillation in the pre-phase and enhancement in the post-phase. Thus, improving both distillation and enhancement can significantly boost network performance.

\begin{figure*}[!ht]
\centering
\includegraphics[width=0.7\linewidth]{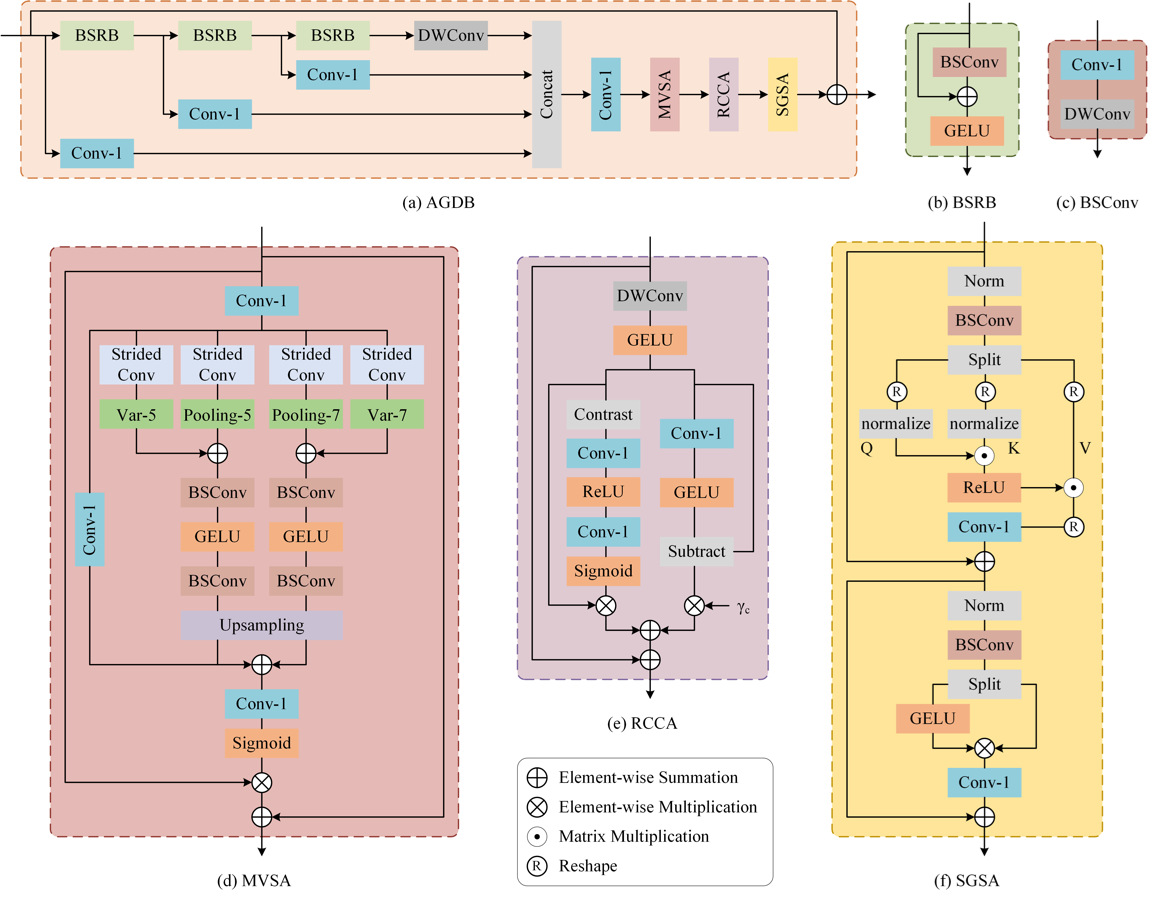}
\caption{\textit{Team XJU\_100th Ann: }The details of each component. (a) AGDB: Attention Guidance Distillation Block; (b) BSRB: Blueprint Shallow Residual Block; (c) BSConv: Blueprint Separable Convolution; (d) MVSA: Multi-level Variance-aware Spatial Attention; (e) RCCA: Reallocated Contrast-aware Channel Attention; (f) SGSA: Sparse Global Self-attention.}
\label{Figure AGDB}
\end{figure*}

Based on the above analysis, they propose the new attention guidance distillation block (AGDB) with more efficient spatial attention, channel attention and self-attention as the base block of AGDN. As shown in Fig.\ref{Figure AGDB}, they use the multi-level variance-aware spatial attention (MVSA) and reallocated contrast-aware channel attention (RCCA) as alternatives to the enhanced spatial attention (ESA)~\cite{RFANet} and contrast-aware channel attention (CCA)~\cite{IMDN}, and introduce sparse global self-attention (SGSA)~\cite{DLGSANet} to achieve further feature enhancement.

In MVSA, they consider the impact of multi-level branching and local variance on performance. Multi-level branches with small windows cannot cover a sufficient range of information while using local variance in a single branch can lead to large differences in weights between branches. Therefore, they improved MDSA~\cite{MDRN} to obtain D5 and D7 branches that contain both local variance to better capture structurally information-rich regions while balancing performance and model complexity In RCCA, they not only consider the reallocation of weights across channels by traditional channel attention but also enhance the treatment of common information across all channels. They added complementary branches with 1 $\times$ 1 convolution and GELU activation representations to reallocate complementary channel information, promoting the uniqueness of each channel. Finally, they introduce SGSA for selecting the most useful similarity values, aiming to better utilize essential global features for image reconstruction. In image reconstruction, there is usually a gap in global attention between the training and testing phases. Therefore, they adopt the default enhancement approach of SGSA, which is to apply the test-time localizer converter (TLC)~\cite{TLC} approach during the testing phase.

\noindent\textbf{Training Details.}
The proposed AGDN has 4 AGDBs, in which the number of feature channels is set to 24. The details of the training steps are as follows:

1. Pretraining on the DIV2K~\cite{agustsson2017ntire} and Flickr2K~\cite{lim2017enhanced} datasets. HR patches of size 256 $\times$ 256 are randomly cropped from HR images, and the mini-batch size is set to 64. The model is trained by minimizing the L1 loss function with the Adam optimizer. The initial learning rate is set to 2 $\times$ 10$^{-3}$ and halved at \{100k, 500k, 800k, 900k, 950k\}-iteration. The total number of iterations is 1000k.

2. Finetuning on 800 images of DIV2K and the first 10k images of LSDIR\cite{lilsdir}. HR patch size and mini-batch size are set to 384 $\times$ 384 and 32, respectively. The model is fine-tuned by minimizing the L2 loss function. The initial learning rate is set to 5 $\times$ 10$^{-4}$ and halved at 50k iteration. The total number of iterations is 100k.
\subsection{VPEG\_C}
\begin{figure*}
	\hspace{7mm}
	\includegraphics[width=0.98\textwidth] {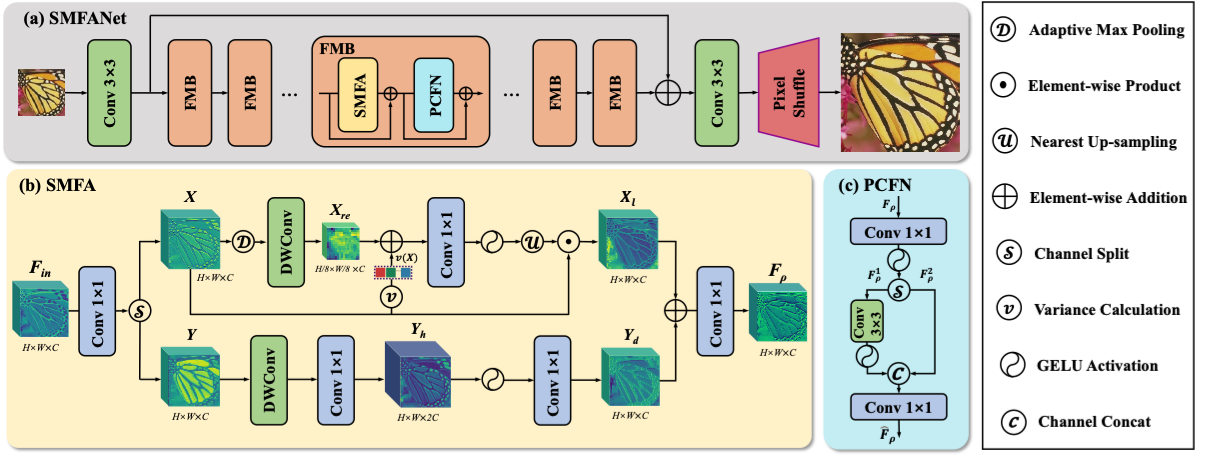}
	\vspace{-3mm}
	\caption{\textit{Team VPEG\_C} : An overview of the proposed SMFAN.}  
	\label{fig:smfan_arch}
\end{figure*}

\textbf{Method.} They introduce a self-modulation feature aggregation (SMFA) module as shown in Figure~\ref{fig:smfan_arch} to collaboratively exploit both local and non-local feature interactions for image super-resolution. 
Specifically, the SMFA module employs an efficient approximation of the self-attention (EASA) branch to model non-local information 
and uses a local detail estimation (LDE) branch to capture local details.
Additionally, they further introduce a partial convolution-based feed-forward network (PCFN) to refine the representative features derived from the SMFA.
%
%
%
%
%
Given the input feature $F_{in} \in \mathbb{R}^{H \times W \times C}$, where $H\times W$ denotes the spatial size and $C$ is the number of channels, 
they first apply a $1\times1$ convolution to the normalized $F_{in}$ to expand the channel, and then split the channel into two parts as inputs to the efficient approximation of self-attention (EASA) and local detail estimation (LDE) branches: 
\begin{equation}
    \{X, Y\} = \mathcal{S}(Conv_{1\times1}(||F_{in}||_2)), 
    \label{eq: smfa-split}
\end{equation}
where $||\cdot||_2$ is the $L_2$ normalization, $Conv_{1\times1}(\cdot)$ denotes a $1\times 1$ convolutional layer, $\mathcal{S}(\cdot)$ denotes a channel splitting operation, and $\{X, Y\} \in \mathbb{R}^{H \times W \times C}$.
They then process the features $X$ and $Y$ in parallel via the EASA and LED branches, producing the non-local feature $X_l$ and local feature $Y_d$, respectively.
Finally, they fuse $X_l$ and $Y_d$ together with element-wise addition and feed them into a $1\times1$ convolution to form a representative output of the SMFA module.
%
%
This process can be formulated as:
\begin{equation}
     F_{\rho} = F_{in} + Conv_{1\times1}(X_{l} + Y_{d}),
    \label{eq: smfa-fusion}
\end{equation}
where $F_{\rho} \in \mathbb{R}^{H \times W \times C}$ is the output feature.

\noindent
For an efficient approximation of self-attention, they obtain the low-frequency components through a downsampling operation and feed them into a $3\times3$ depth-wise convolution to generate non-local structure information $X_{s}\in \mathbb{R}^{H/8 \times W/8 \times C}$:
 \begin{equation}
      X_{s} = DWConv_{3\times3}(\mathcal{D}(X)), 
      \label{eq: smfa-downsampling}
\end{equation}
where $\mathcal{D}(\cdot)$ denotes the adaptive max pooling with a scaling factor of 8, $DWConv_{3\times3}(\cdot)$ is a $3\times3$ depth-wise convolutional layer. 
To embed global descriptions for modulating non-local representation $X_{s}$,
they introduce variance $\nu(X) \in \mathbb{R}^{1 \times 1 \times C}$ of the input $X$ and calculate it in the spatial dimension. $X_{s}$ and $\nu(X)$ are then added and fed into a $1\times1$ convolution to fuse the information thoroughly:
\begin{equation}
     X_{m} = Conv_{1\times1}(X_{s} + \nu(X)),
  \label{eq: smfa-variance}
\end{equation}
where $X_{m} \in \mathbb{R}^{H \times W \times C}$ represents the modulated feature. 
This variance modulation mechanism facilitates better exploring non-local information.

Finally, they use the modulated features to aggregate the input feature $X$ for extracting the representative structure information $X_{l}$:
\begin{equation}
  \begin{split}
     & X_{l} = X \odot \mathcal{U}(\phi(X_{m})), \\
  \end{split}
  \label{eq: smfa-aggregation}
\end{equation}
where $\phi(\cdot)$ refers to the GELU activation function~\cite{hendrycks2016gaussian}, $\mathcal{U} (\cdot)$ denotes a nearest upsampling operation, and $\odot$ represents the element-wise product operation.

\noindent\textbf{Training Details.} The proposed SMFAN consists of 8 FMBs and the number of channels is set to 24. They first train the proposed SMFAN on the DIV2K~\cite{DIV2K} and Flickr2K~\cite{lim2017enhanced} datasets.
The cropped LR image size is 96 $\times$ 96 and the mini-batch size is set to 64. 
The SMFAN is trained by minimizing L1 loss and the frequency loss~\cite{cho2021rethinking} with Adam optimizer for total of 800,000 iterations. 
They set the initial learning rate to $ 2 \times 10^{-3}$ and the minimum one to $1 \times 10^{-5}$, which is updated by the Cosine Annealing scheme~\cite{SGDR}.
After that, they use the first 10,000 images of LSDIR~\cite{lilsdir} dataset for fine-tuning. 
The cropped LR image size is 160 $\times$ 160 and the mini-batch size is set to 32. 
The fine-tuning stage uses MSE loss and the frequency loss~\cite{cho2021rethinking}  with 500,000 iterations.
The initial learning rate is set to $2 \times 10^{-5}$ and the minimum one to $1 \times 10^{-7}$.

\subsection{ZHEstar}
\begin{figure}
\centering
\includegraphics[width=1\columnwidth]{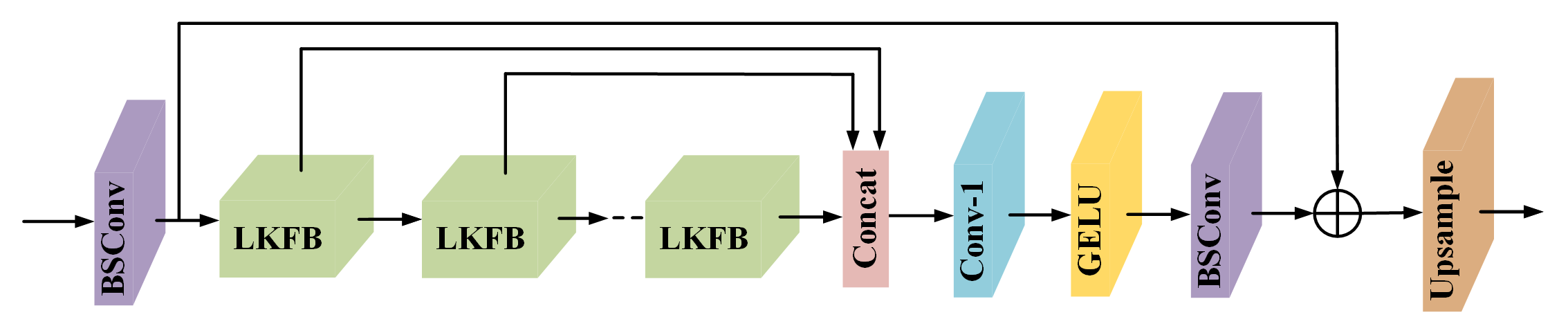}
\caption{\textit{Team ZHEstar:} The framework of Large Kernel Frequency-enhanced Network (LKFN)}
\label{fig:LKFN}
\end{figure}
\begin{figure}
\centering
\includegraphics[width=1\columnwidth]{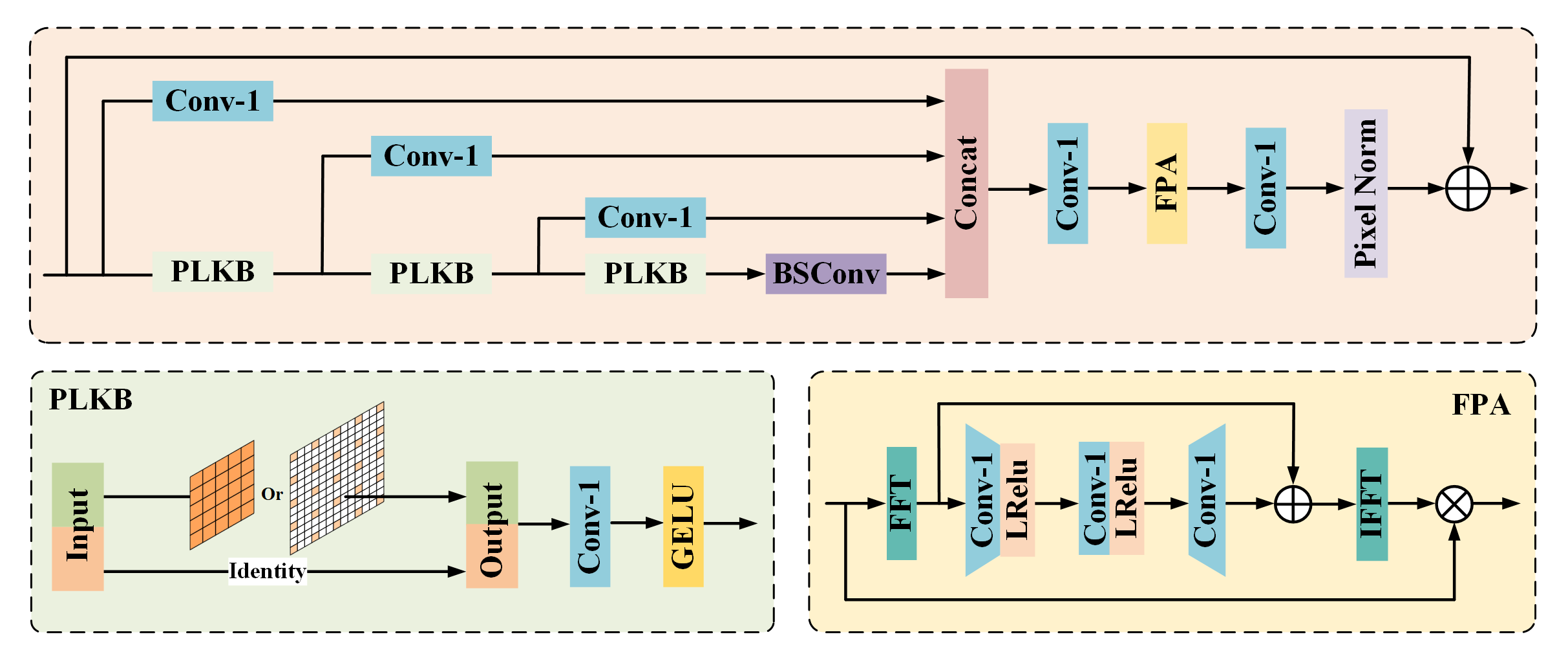}
\caption{\textit{Team ZHEstar:} Large Kernel Frequency-enhanced Block (LKFB)}
\label{fig:LKFB}
\end{figure}
\noindent{\textbf{Method.}}
The ZHEstar team proposed Large Kernel Frequency-enhanced Network (LKFN)~\cite{chen2024lkfn}. The architecture is shown in Fig.\ref{fig:LKFN}. It is based on BSRN~\cite{li2022blueprint}. They replaced the RBSB in BSRN with their Partial Large Kernl Block (PLKB) shown in Fig.\ref{fig:LKFB}. Inspired by PConv ~\cite{chen2023run}, PLKB first divides the input feature map into two halves in the channel dimension. One half undergoes a $5 \times 5$ depth-wise convolution (The third PLKB with a dilation rate of 3.), and the result is then concatenated with the unprocessed other half. A $1 \times 1$ convolution is subsequently used to perform data exchange between these two parts.
For their Frequency-enhanced Pixel Attention (FPA) module, it first transforms the spatial domain feature map to the frequency domain through Fourier transform, then pass the frequency domain map through a three-layer $1 \times 1$ convolution, followed by two LeakyReLUs. The result is added to the initial frequency domain map via a residual connection to obtain the enhanced frequency domain attention map. Then it was transformed back to the spatial domain and multiplied by the input spatial feature map.

\noindent{\textbf{Training Details.}}
The proposed LKFN consists of 8 LKFBs and the feature channel is set to 28. The training data includes 800 images from DIV2K~\cite{agustsson2017ntire} and the first 10K images from LSDIR ~\cite{lilsdir}. They use the default parameter settings of the Adan optimizer~\cite{xie2022adan} in the whole process. The training process is as follows:
\begin{enumerate}
    \item Training with an input patch size of $64 \times 64$ and a mini-batch size of 64 from scratch by minimizing the $\mathcal{L}_{1}$ loss. The initial learning rate is set to $5 \times 10^{-3}$. The learning rate decay is following cosine annealing with $T_{max}$ = total iterations, 
 $\eta_{min}$ = $1 \times 10^{-7}$. The total number of iterations is 1000K.
    \item Finetuning with an input patch size of $120 \times 120$ and a mini-batch size of 64  by minimizing the MSE loss. The learning rate is set to $2 \times 10^{-5}$ during this stage.  The total number of iterations is 150K.
\end{enumerate}

\subsection{VPEG\_O}

\begin{figure}
	\centering
	\includegraphics[width=0.48\textwidth]{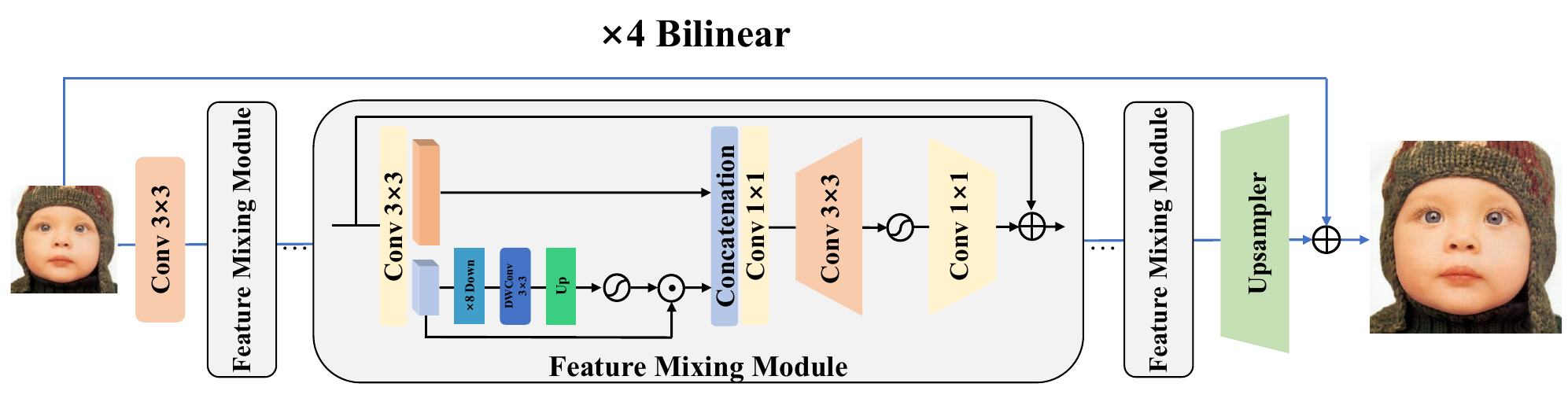}
	\vspace{-3mm}
	\caption{\textit{Team VPEG\_O}: The overall network architecture of their proposed SAFMN++.}
	\label{fig:framework_team23}
\end{figure}

\textbf{Method.}
The VPEG\_O team introduces SAFMN++, an improved version of SAFMN~\cite{sun2023safmn} for solving efficient SR.
This solution mainly concentrates on improving the effectiveness of the spatially adaptive feature modulation (SAFM)~\cite{sun2023safmn} layer.
Different from the original SAFM, as shown in Fig~\ref{fig:framework_team23}, the improved SAFM (SAFM+) can extract both local and non-local features.
In SAFM+, a 3$\times$3 convolution is first utilized to extract local features and a single scale feature modulation is then applied to a portion of the extracted features for non-local feature interaction.
After this process, these two sets of features are aggregated by channel concatenation and fed into a 1$\times$1 convolution for feature fusion.

\noindent\textbf{Training details.}
The proposed SAFMN++ consists of 6 feature mixing modules, and the number of channels is set to 36.
They train the proposed SAFMN++ on the LSIDR~\cite{lilsdir} dataset. 
The cropped LR image size is $120\times120$ and the mini-batch size is set to 64. 
The SAFMN++ is trained by minimizing L1 loss and the frequency loss\cite{cho2021rethinking} with Adam optimizer for total 800, 000 iterations. 
They set the initial learning rate to $3\times10^{-3}$ and the minimum one to $1\times10^{-6}$, which is updated by the Cosine Annealing scheme~\cite{SGDR}. 
 
\subsection{CMVG}

\begin{figure*}[htbp]
\begin{center} 
\includegraphics [width=0.8\textwidth]{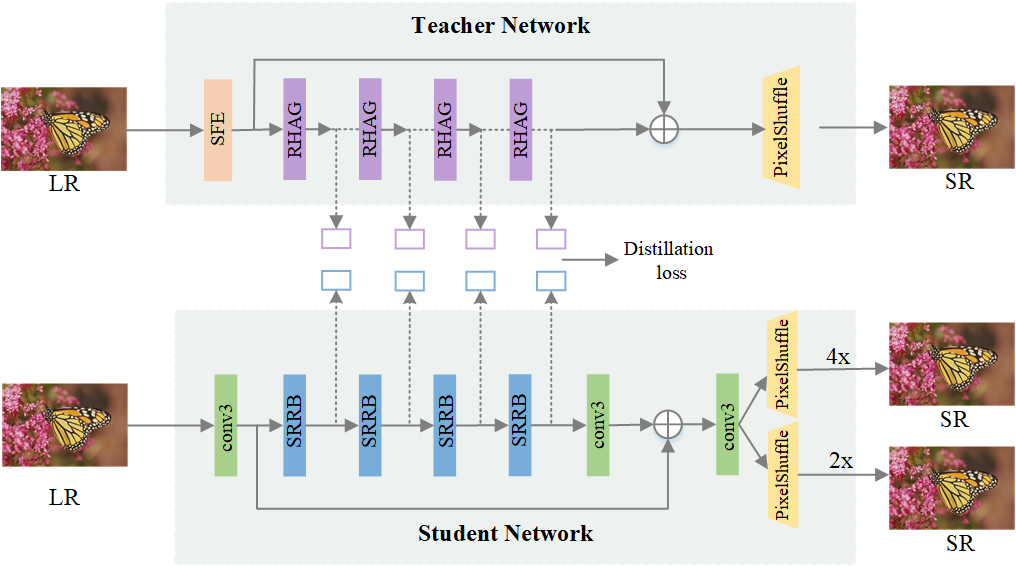} 
\caption{\textit{Team CMVG: }The framework of RDEN} 
\label{fig1}
\end{center} 
\end{figure*}

\begin{figure*}[htbp]
\begin{center} 
\includegraphics [width=0.5\textwidth]{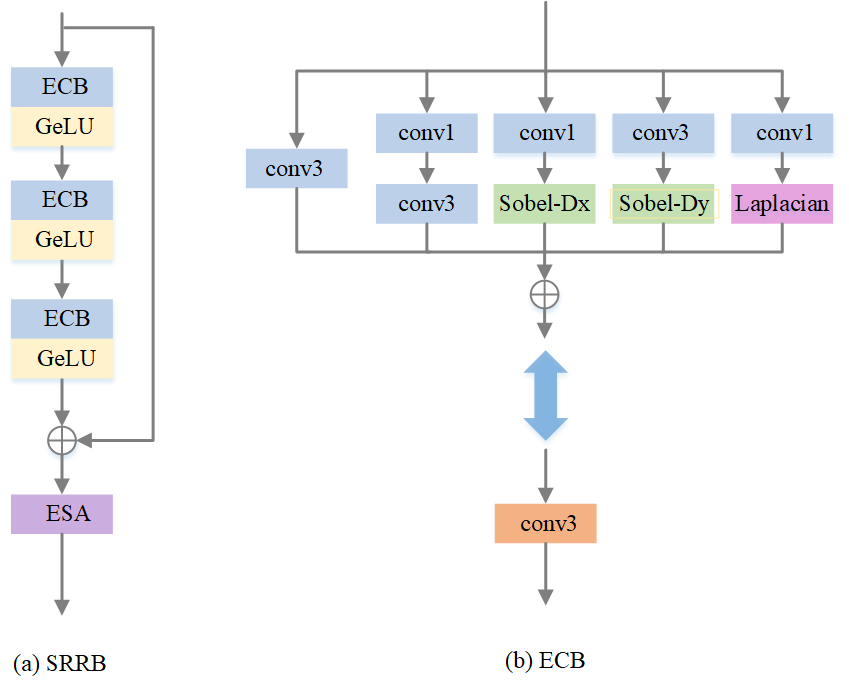} 
\caption{\textit{Team CMVG: }The framework of SRRB} 
\label{fig2}
\end{center} 
\end{figure*}

\textbf{Method}. They propose a residual knowledge distillation super-resolution network named RDEN for efficient super-resolution(SR) as shown in Figure.\ref{fig1}. 
The RDEN model is composed of a large teacher network and a lightweight student network, they apply the superior SR model HAT\cite{chen2023activating} as a teacher network. The distillation training provides additional effective supervision information for student training, and enhance the performance and generalization ability of the student network. The student network includes the shallow feature extraction, the depth feature extraction, and the reconstruction modules. There are four simplified reparameterization residual blocks (SRRB) in depth feature extraction modules, as shown in Figure.\ref{fig2}. The reconstruction module includes a 4$\times$ and 2$\times$ upsampling head. A SRRB block is composed of three residual connected ECB\cite{zhang2021edge} blocks and ESA\cite{kong2022residual}. They utilize ECB blocks during the training phase, while they can be merged into a single 3x3 convolution layer during inference through mathematical transformation. The number of feature maps of Conv3 and SRRB is set to 38 respectively, after the pruning, the feature maps are reduced to 37 finally. Although the model is small, the knowledge distillation and reparameterization provide the information compensation, which enables their model to achieve good reconstruction performance. They train the teacher network with L1 loss, the teacher loss is denoted as follows:

\begin{equation}
\label{eq:team26_eq1}
\begin{split}
    L_T = \|Y^{SR}_{T} - Y^{HR}_{T}\|
\end{split}
\end{equation}

Where the  {$Y_{T}^{SR}$} is SR image from teacher and  {$Y_{T}^{HR}$} is HR image. The student network is trained by the distillation loss, L1 loss, and joint supervision loss. They extract features from the 0th, 2nd, 4th, and 5th blocks of the teacher model, and features from each SRRB block, then they use a 1*1 convolution to expand the student feature dimension to match corresponding teacher features. The distillation loss is denoted as follows:
\begin{equation}
\label{eq:team26_eq2}
\begin{split}
    L_{distillation} = \lambda_i\sum_{i=0}^4\|F_{T}^{i}-F_{S}^{i}\|+\mu\|Y^{SR}_{T}-X^{SRx4}_S\|
\end{split}
\end{equation}

where {$F_{S}^{i}$} represents the feature map of the output of the i-th block of the student network, and  {$F_{T}^{i}$} is the corresponding features of teacher model.  {$X_{S}^{SRx4}$} represents the output of 4x upsample heads in student network. The joint supervision loss is composed of GM loss, FFT loss, and 2x supervision loss, denoted as follows:
\begin{equation}
\label{eq:team26_eq3}
\begin{split}
    L_{GM} = \|GM(X_S^{SRx4})-GM(X_S^{HRx4})\|
\end{split}
\end{equation}

\begin{equation}
\label{eq:team26_eq4}
\begin{split}
    L_{FFT} = \|FFT(X_S^{SRx4})-FFT(X_S^{HRx4})\|
\end{split}
\end{equation}

\begin{equation}
\label{eq:team26_eq5}
\begin{split}
    L_{X2} = \|X_{S}^{SRx2}-X_{S}^{HRx2}\|
\end{split}
\end{equation}

where {$GM(.)$} and {$FFT(.)$} respectively represent the gradient map and focal frequency extraction operators, the {$X_{S}^{SRx2}$} denotes the output from 2$\times$ upsampling head, the {$X_{S}^{HRx4}$} and {$X_{S}^{HRx2}$} are corresponding 4$\times$and 2$\times$ HR image. Finally, the student loss for training student network is :

\begin{equation}
\label{eq:team26_eq6}
\begin{split}
    &L_S = \alpha\|X_{S}^{SRx4}-X_{S}^{HRx4}\|+\beta L_{distillation}+\gamma L_{GM} \\
    &+\delta L_{FFT}+\epsilon L_{X2}
\end{split}
\end{equation}

\noindent\textbf{Training Details.} They train their model on DIV2K, Flickr2K, and LSDIR datasets, and the multi-stage progressive training strategy is used to optimize and finetune. The progressive training strategy gradually increases patch size, changes loss function, and loads weights from the previous step to improve performance. The training details are described as follows:\\
\textbf{Stage1} \textit{Training teacher network:} The teacher network HAT is trained from scratch with teacher loss.\\
\textbf{Stage2} \textit{Training student network:} The teacher network is fixed, and they pretrain a 2$\times$ network to initialize the student network. Then, the student network is optimized through distillation training by student loss. The initial learning rate is set to 5e-4 and halved at every 60 epochs and the total number of epochs is 500. The batch size and patch size are set to 64 and 256 separately. Data augmentation is also adopted. \\
\textbf{Stage3} The finetune steps of the student network are described as follows: \\
(1) The student model is initialized from Stage 2 and trained with the same settings as in the previous step. \\
(2) The student model is initialized from Stage 3.1 and trained with the same settings as Stage 3.1, especially since the loss function is only MSE loss. \\
(3) The student model is initialized from the previous step and finetuned by MSE loss further, it is worth noting that the patch size is set to 512. Other parameter settings are not changed and finally, the student model is finetuned with 640 HR patches and MSE loss. 
\subsection{LeESR}

\begin{figure}[t]
\centering
\includegraphics[width=1\columnwidth]{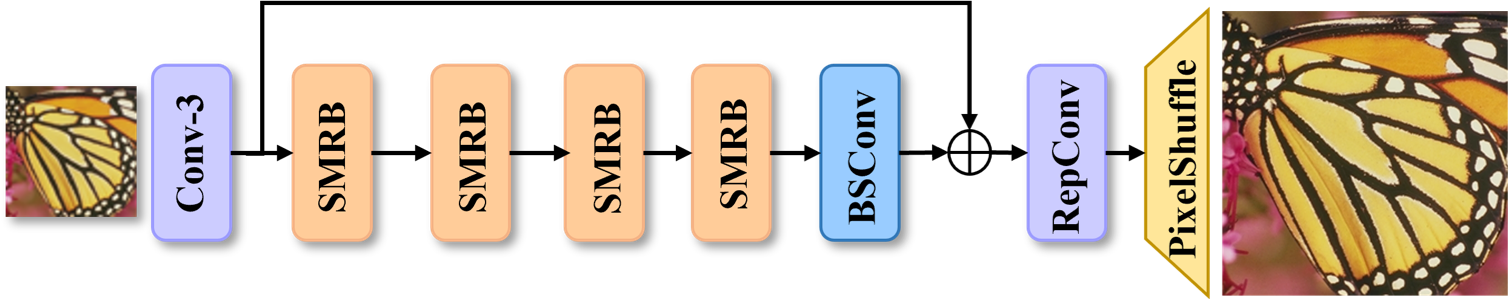}
\caption{\textit{Team LeESR: }The framework of the proposed Separable-Mixable Residual Network (SMRN).}
\label{fig:pipline_team10}
\vspace{-0.2cm}
\end{figure}

\begin{figure*}[th]
\centering
\includegraphics[width=1.0\textwidth]{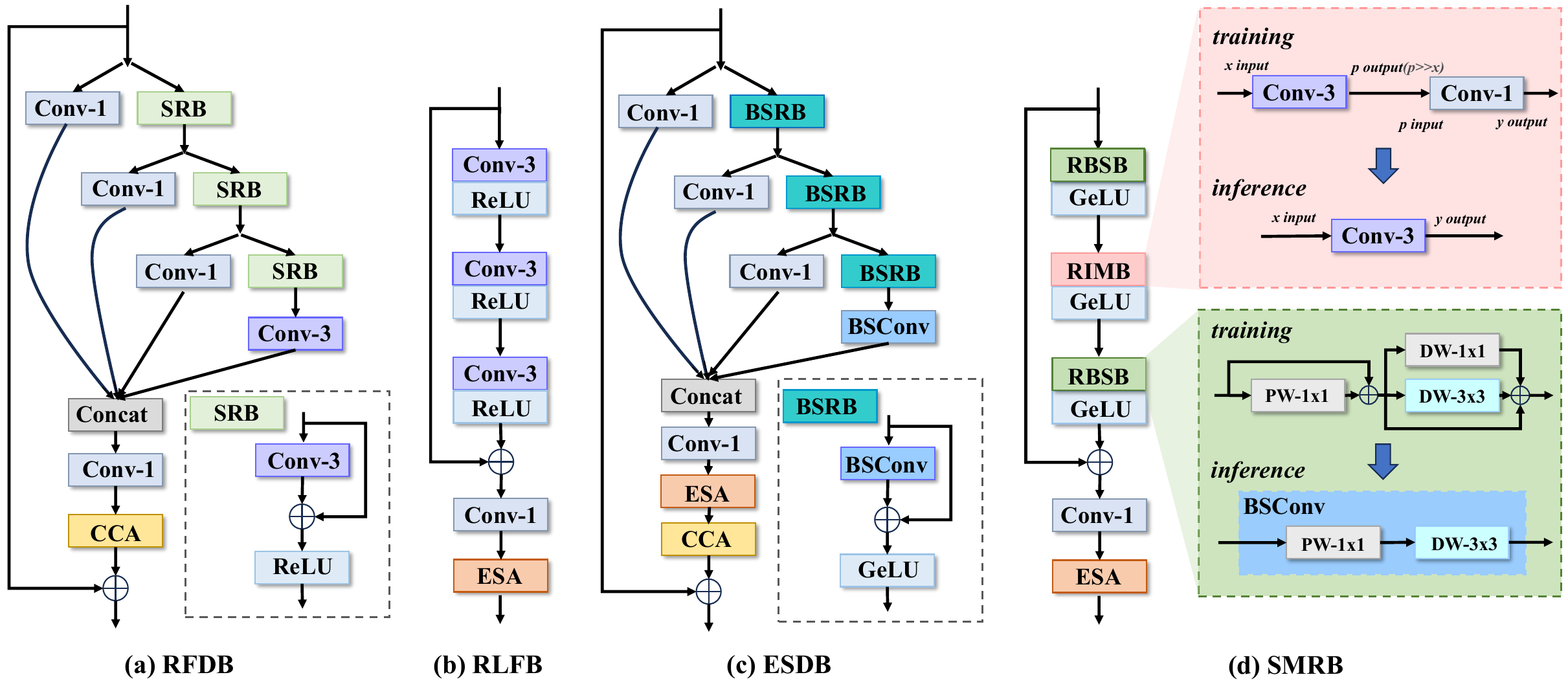}
\caption{\textit{Team LeESR: }(a) Residual feature distillation block (RFDB). (b) Residual local feature block (RLFB). (c) Efficient
separable residual block (ESDB). (d) The proposed Separable-Mixable Residual Block (SMRB). }
\label{fig:block}
\vspace{-0.2cm}
\end{figure*}

\noindent\textbf{Method.}
Inspired by the RLFN \cite{kong2022residual} and BSRN \cite{li2022blueprint}, this team proposed a Separable-Mixable Residual Network (SMRN) as illustrated in Fig.\ref{fig:pipline_team10} for Efficient image Super-Resolution, which can maintain lower parameters and computation while performing faster. 
%
Unlike the popular  RFDB (see Fig.\ref{fig:block}(a)) in RFDN \cite{liu2020residual}, the RLFB (see Fig.\ref{fig:block}(b)) in RLFN, and the ESDB (see Fig.\ref{fig:block}(c)), 
the BSConv (see Fig.\ref{fig:block}(d)) in the proposed SMRB consists of a 1$\times$1 pixel-wise convolution and a depth-wise convolution and considers the intra kernel correlation. Among it, a kernel on a single channel (as a blueprint) is multiplied with different weights (e.g., 1$\times$1 pixel-wise convolution) to obtain the convolution kernels on other different channels. Obviously, the strategy can greatly simplify traditional convolution operations. However, only the kernel of different channels separately is limited, the features of different channels still need to mix for performance improvement. Therefore, as shown in Fig.\ref{fig:block}(d), a Separable-Mixable Residual Block (SMRB) is designed, which consists of both the blueprint separable operation and the information mixable operation. Specifically, they introduce traditional convolutional kernels to be used in conjunction with BSconv for feature separation and mixable. 

Reparameterization has improved the performance of ESR without introducing any inference cost.
In Separable-Mixable Residual Block (SMRB), they introduce the Re-parameterized Blueprint Separable Block (RBSB) to improve the BSconv representation, while a Re-parameterized Information Mixable Block (RIMB) is designed to replace the regular convolution. 
Specifically, inspired by SESR \cite{sesr}, during training, the conv in RIMB is a collapsible block which consists of a 3 $\times$ 3 convolution and a 1$\times$1 convolution in the expanded space. During inference, they collapse them to a single 3$\times$3 convolution as depicted in Fig.\ref{fig:block}(d). In image reconstruction, a re-parameterized 3 $\times$ 3 convolution (RepConv) as shown in Fig.\ref{fig:pipline_team10} is utilized to replace the single 3 $\times$ 3 convolution.

\noindent\textbf{Training Details.}
The proposed SMRN contains four SMRBs, in which they set the number of feature maps to 48. Also, the channel number of the ESA is set to 16 similar to \cite{kong2022residual}. Besides, the collapsed channel number in the collapsible block is 256.
Throughout the entire training process, they use the Adam optimizer \cite{kingma2014adam}, where $\beta{1}$ = 0.9 and $\beta{2}$ = 0.999. The model is trained for 1000k iterations in each stage.
Input patches are randomly cropped and augmented. Data augmentation strategies included horizontal and vertical flips, and random rotations of 90, 180, and 270 degrees.
Model training was performed using Pytorch 1.11.0 \cite{2019PyTorch} on one NVIDIA A100 40G GPUs.
Specifically, the training strategy consists of several steps as follows.

1. In the starting stage, they train the model from scratch on the 800 images of DIV2K \cite{DIV2K} and the first 10k
images of LSDIR \cite{lilsdir} datasets.
The model is trained for total $10^6$ iterations by minimizing Charbonnier loss and FFT loss \cite{cho2021rethinking}. The HR patch size is set to 256$\times$256, while the mini-batch size is set to 64. They set the initial learning rate to 1 $\times$ $10^{-3}$ and the minimum one to 1 $\times$ $10^{-6}$, which is updated by the Cosine Annealing scheme.

2. In the second stage, they increase the HR patch size to 384. The model is fine-tuned by minimizing the Charbonnier
loss and FFT loss. They utilize the MultiStepLR scheduler with a warm-up strategy (2,000 iterations for warm-up), where the initial learning rate is set to 5 $\times$ $10^{-4}$ and halved at {200k, 400k, 800k}-iterations. 

3. In the third stage, the model is initialized with the pre-trained weights of Stage 2, and fine-tuned with larger HR patches of size 480x480. Other settings are the same as in the second stage.

4. In the fourth stage, the model is further fine-tuned with 480×480 HR patches, however, the loss function is changed to minimize the combination of L2 loss and FFT loss. Other settings are the same as Stage 3.

\subsection{AdvancedSR}
\begin{figure*}
    \centering
    \includegraphics[width=0.9\linewidth]{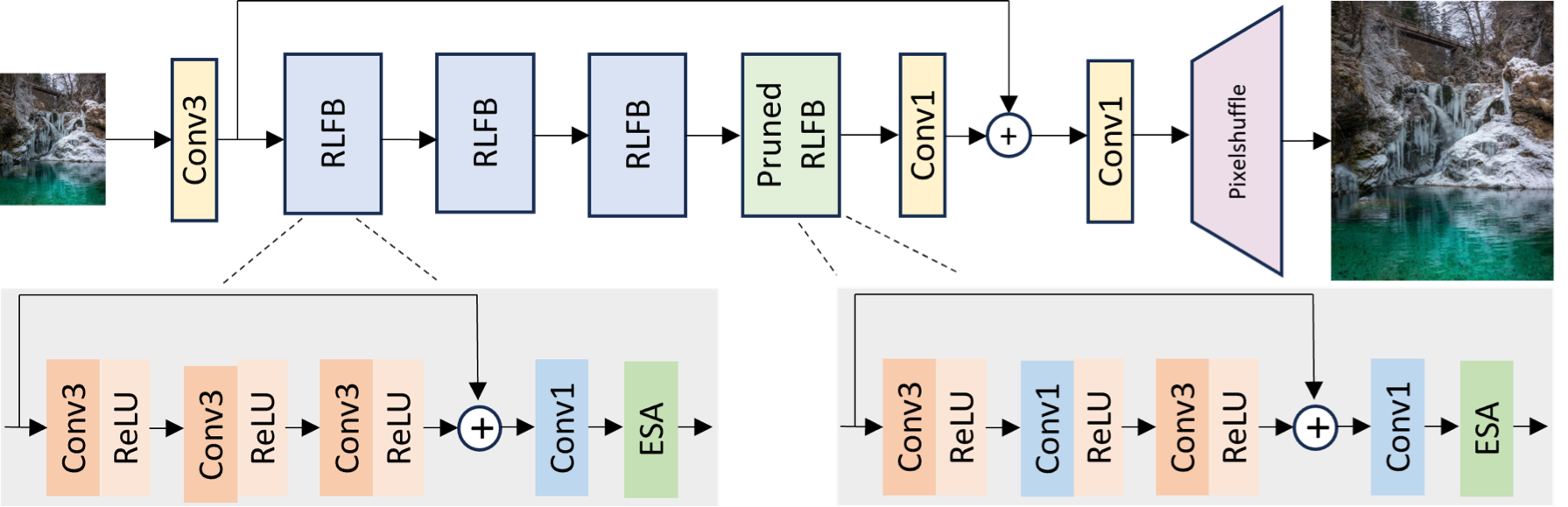}
   \caption{\textit{Team AdvancedSR}:  The overall architecture of AdvancedSR.}
    \label{fig:AdvancedSR}
\end{figure*}
\textbf{Method.}
The AdvancedSR team proposes a kernel-pruning-based Residual Local Feature Network~\cite{kong2022residual} for efficient SR. The overall network architecture is illustrated in Figure.\ref{fig:AdvancedSR}. It is a lightweight network consisting of a series of RLFB blocks, similar to RLFN. However, they prune the second convolution in the pruned-RLFB block based on sensitivity analysis. Additionally, the pixel shuffle block is used for image restoration.

Additionally, this team perform the pruning steps based on NTIRE2024 official baseline model~\cite{kong2022residual}. The pruning process consists of two stages. In the first stage, they apply progressive kernel-pruning on the re-parameterized model inspired by UPDP\cite{liu2024updp}. And in the second stage, they apply bias prune in their model to accelerate the runtime. 

Stage 1. Kernel Pruning. They conducted a sensitivity analysis on all the conv3$\times$3 of the RLFN Network.  Based on the proportion of responses of the center weights in Conv3, they replaced the conv3 with conv1 progressively. 
The experiments show that replacing Conv3 with Conv1s scarcely degrades the performance but can enhance runtime efficiency. 
Despite channel pruning boasting fewer FLOPs, its runtime performance falls short of kernel pruning. 

Stage 2. Bias Pruning. After kernel pruning, they obtain a subnet with a modified RLFB network. They preserve the biases of the first conv and ESA module, then retrain the subnet by removing the remaining biases thus further enhancing runtime efficiency.






\noindent\textbf{Training Details.}
The model is trained on LSDIR dataset\cite{lilsdir} and they use RLFN (the official baseline model of NTIRE2024) as their basemodel. The training HR patch size is set as 256$\times$256 with data augmentation such as rotation and horizontal flip in order to enhance the comprehensive ability of the model.  They set the batch size as 64 in the training process with total of 500 epochs. The model is trained by minimizing L2 loss with Adam optimizer. The initial learning rate is set to 2e-5 and the learning rate is decayed by half at 100 epochs. 


\subsection{ECNU\_MViC}
\begin{figure*}[ht]
    \centering
    \includegraphics[width=0.9\textwidth]{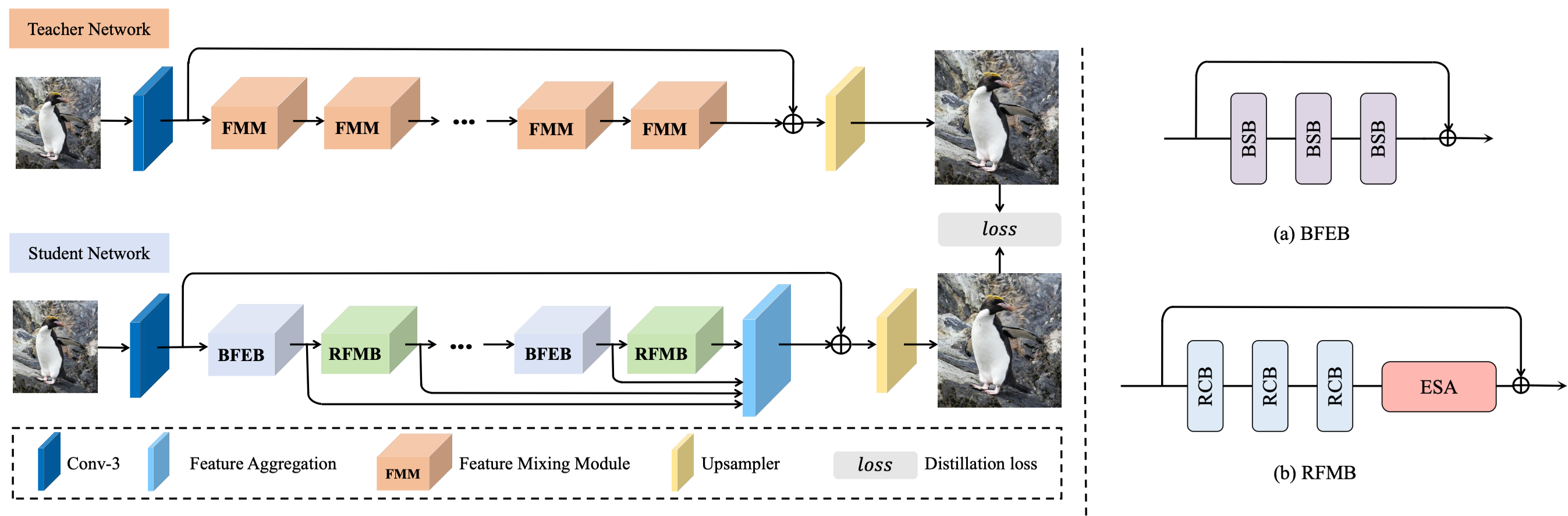}
    \caption{\textit{Team ECNU\_MViC:} The pipeline of IFADNet. (a) Structure of RCB. (b) Structure of RFMB.}
    \label{fig:image1_team33}
\end{figure*}

\noindent{\textbf{Method.}} As shown in Fig.~\ref{fig:image1_team33}, they propose an intermittent feature aggregation network named IFADNet. The architecture comprises of three parts: the shallow feature extraction, the deep feature extraction based on alternating BFEB blocks and RFMB blocks, and the reconstruction stage. \\
   They employ a single 3$\times$ 3 convolution to extract the shallow feature $F_{s} \in \mathbb{R}^{C \times H \times W }$ in the first stage $H_{F}$:
    \begin{equation}
    F_{s}=H_{F}(I_{i}),
    \end{equation}
where $I_{i}$, $C$, $H$, $W$ are the input image, the embedding channel dimension, height and width of the input, respectively.
    
\begin{figure*}[ht]
    \centering
    \includegraphics[width=\textwidth]{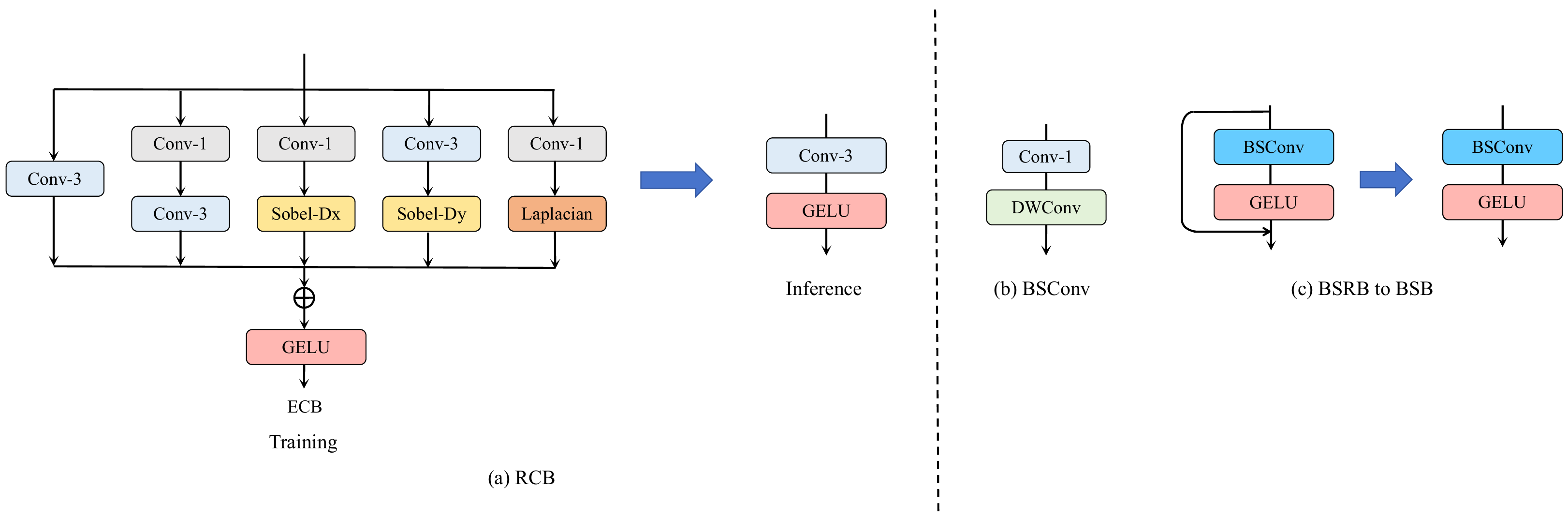}
    \caption{\textit{Team ECNU\_MViC:} (a) Structure of BFEB. (b) Structure of BSConv. (c) From BSRB to BSB.}
    \label{fig:image2_team33}
\end{figure*}

During the second stage, six intermittent blocks are used to extract the deep feature $F_{d} \in \mathbb{R}^{C \times H \times W }$:
    \begin{equation}
    F_{d}=H_{D}(F_{s}).
    \end{equation}
Specifically, $H_{D}$ consists of an alternating blueprint feature extraction block~(BFEB) and a parameterized feature modulation block~(RFMB). The details of BFEB and RFMB will be introduced in the next paragraph. By using $F_{s}$ and $F_{d}$ as inputs, the high-quality image $I_r$ is generated in the reconstruction stage denoted by $H_{R}$ as:
    \begin{equation}
    I_{r}=H_{R}(F_{s}+F_{d}),
    \end{equation}
where $H_{R}$ involves a single 3$\times$3 convolution followed by a pixel shuffle operation~\cite{shi2016real}.
    
Inspired by the blueprint shallow residual block~\cite{li2022blueprint}, they designed a blueprint feature extraction block to reduce the computation time, which did not significantly reduce model performance. As shown in Fig.~\ref{fig:image1_team33} (a), the BFEB contains three blueprint shallow blocks ~(BSB) which contain a 1$\times$1 point-wise convolution with a 3 $\times$ 3 depth-wise convolution followed by GELU activation. The reparamaterized feature modulation block~(RFMB) consists of three reparameterized convolution blocks~(CB) and an enhanced spatial attention block~\cite{kong2022residual} is employed to extract and modulate the deep feature fully. The detailed structure is illustrated in Fig.~\ref{fig:image2_team33}. They further observe that the intermittent setting of blocks significantly reduces model complexity while not largely impairing model effectiveness. At the end of the second stage, the extracted features of each block are concatenated and aggregated using two convolutions. 

 Reparameterization has shown a strong ability to improve the feature presentation. Different from the reparameterization module design of high-level tasks, they design an isotropic edge-oriented convolutional block in their model. As shown in Fig.~\ref{fig:image2_team33}(a), the Sobel-Dx and Sobel-Dy employ isotropic Sobel functions to enhance the network’s representation capabilities. For inference, the output is computed in a simplified $3\times3$ convolution, which significantly reduces computation cost.

\noindent{\textbf{Training Details.}} They use DIV2K~\cite{agustsson2017ntire} and the first 10K images of LSDIR~\cite{lilsdir} to train their model. The training dataset is augmented with horizontal flips and 90-degree rotations. Knowledge distillation is applied to improve the model performance. They use the large version of pre-trained SAFMN\cite{sun2023safmn} as their teacher model.The student model IFADNet has 6 blocks(3 BFEB and 3 RFMB). The channel of their network is 36. The training details are as follows:
\begin{itemize}
    \item
    Training from scratch. The HR patch size is 256. The mini-batch size is set to 64. The model is trained by minimizing L1 loss and distillation loss (also L1 loss) with Adam optimizer \cite{kingma2014adam}. The initial learning rate is set to 2$\times 10^{-3}$ and halved at \{$100k$, $500k$, $800k$, $900k$, $950k$\}-iteration. The total number of iterations is 1000k.
    \end{itemize}

    \begin{itemize}
    \item
    Finetuning with larger patches. The HR patch size is set to 640. The model is finetuned with MSE loss. Other settings are the same as in the previous step.
\end{itemize}
\subsection{HiSR}

\begin{figure*}[t]
	\centering
	\includegraphics[width=0.85\textwidth]{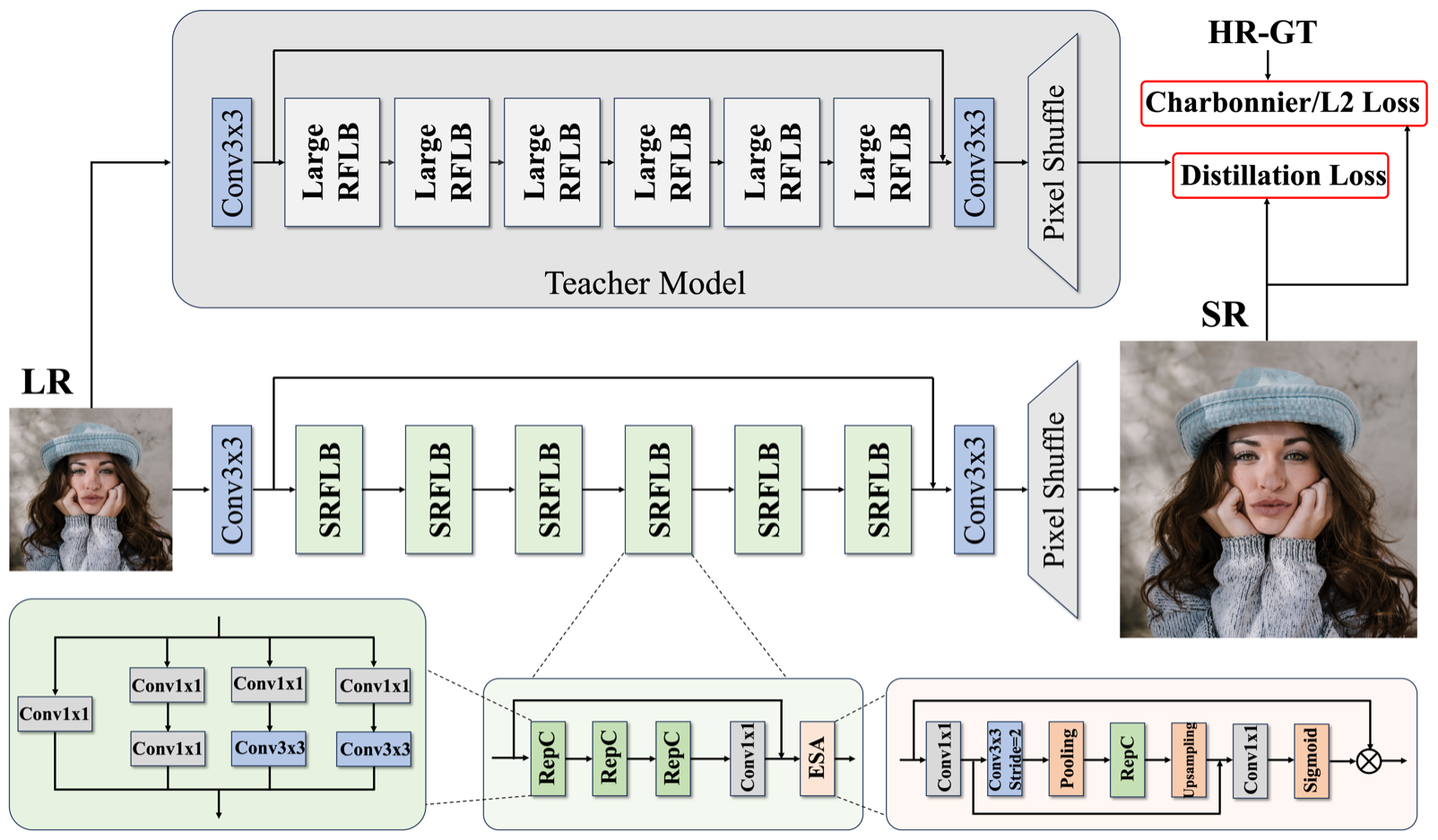}
	\caption{\textit{Team HiSR}: Network architecture of SlimRLFN.}
	\label{fig:pipeline_team21}
\end{figure*}

\textbf{Method.}
They propose the SlimRLFN for the efficient super-resolution task.
The network architecture is inspired by the design of RLFN\cite{kong2022residual}, while fully exploring the capacity of reparameterizable convolution, light distillation, and iterative model pruning. 
The whole architecture is shown in Fig.\ref{fig:pipeline_team21}, which mainly consists of six SRLFB modules and a pixel shuffle module.
Reparameterizable convolutions are utilized in the SRLFB module, aiming to improve the super-resolution capability without introducing any additional parameter overhead during the inference stage.
Meanwhile, the network is optimized by the pixel-wise loss such as charbonnier loss or L2 loss, along with the distillation loss provided by a light but efficient teacher model.
Last but not least, they use iterative pruning to shrink the model size while maintaining the promising performance at the last training stage.

\noindent\textbf{Training Details.}
They choose DIV2K, Flickr2K, and LSDIR datasets as their training datasets, and they augment them with horizontal/vertical flips and rotations during the training stage.
They set the batch size as 64 for all training stages and other hyperparameters such as patch size or learning rate are determined by the specific training stage.
The whole training process is summarised as follows.
\begin{itemize}
\item[1)] Training the teacher model.
They choose the large RLFN as their teacher model. They set the patch size as 256$\times$256, and use charbonnier loss and Adam optimizer for optimization.
They train the teacher model for 100 epochs and set the initial learning rate as 1e-3.
The learning rate decay follows cosine annealing with $T_{max}$ as 100 and $eta_{min}$ as 1e-5.
\item[2)] Training the SlimRLFN with light distillation.
They set the patch size as 256$\times$256, and use two loss functions for training.
The first one is the regular charbonnier loss with ground truth HR image, and the second one is the charbonnier loss between SlimRLFN's output and the teacher's output.
They also choose the Adam optimizer for optimization. They train the SlimRLFN model for 100 epochs, and set the initial learning rate as 1e-3.
The learning rate decay follows cosine annealing with $T_{max}$ as 100 and $eta_{min}$ as 1e-5.
Then they repeat this stage two more times without distillation loss, and the pretrained model is adopted from the last stage.

\item[3)] Training the SlimRLFN with a larger patch size progressively.
They set the patch size as \{384$\times$384,  512$\times$512\}, and set the initial learning rate as \{5e-4, 2.5e-4\} respectively.
Each stage's pretrained model is adopted from the last stage, and they train the model under the same patch size three times in total. 
The other details are the same as before.

\item[4)] Training the SlimRLFN with the patch size of 640$\times$640. 
They use the L2 loss in this stage and set the initial learning rate as 1e-4.
They also adopt cosine annealing with $T_{max}$ as 100 and $eta_{min}$ as 1e-6 for learning rate decay. 

\item[5)] [Optional] SlimRLFN pruning stage.
After training from the above several epochs, they adopt the iterative pruning for the SlimRLFN which has obtained promising performance.

\end{itemize}

\subsection{MViC\_SR}
\begin{figure*}[ht]
    \centering
    \includegraphics[width=0.8\textwidth]{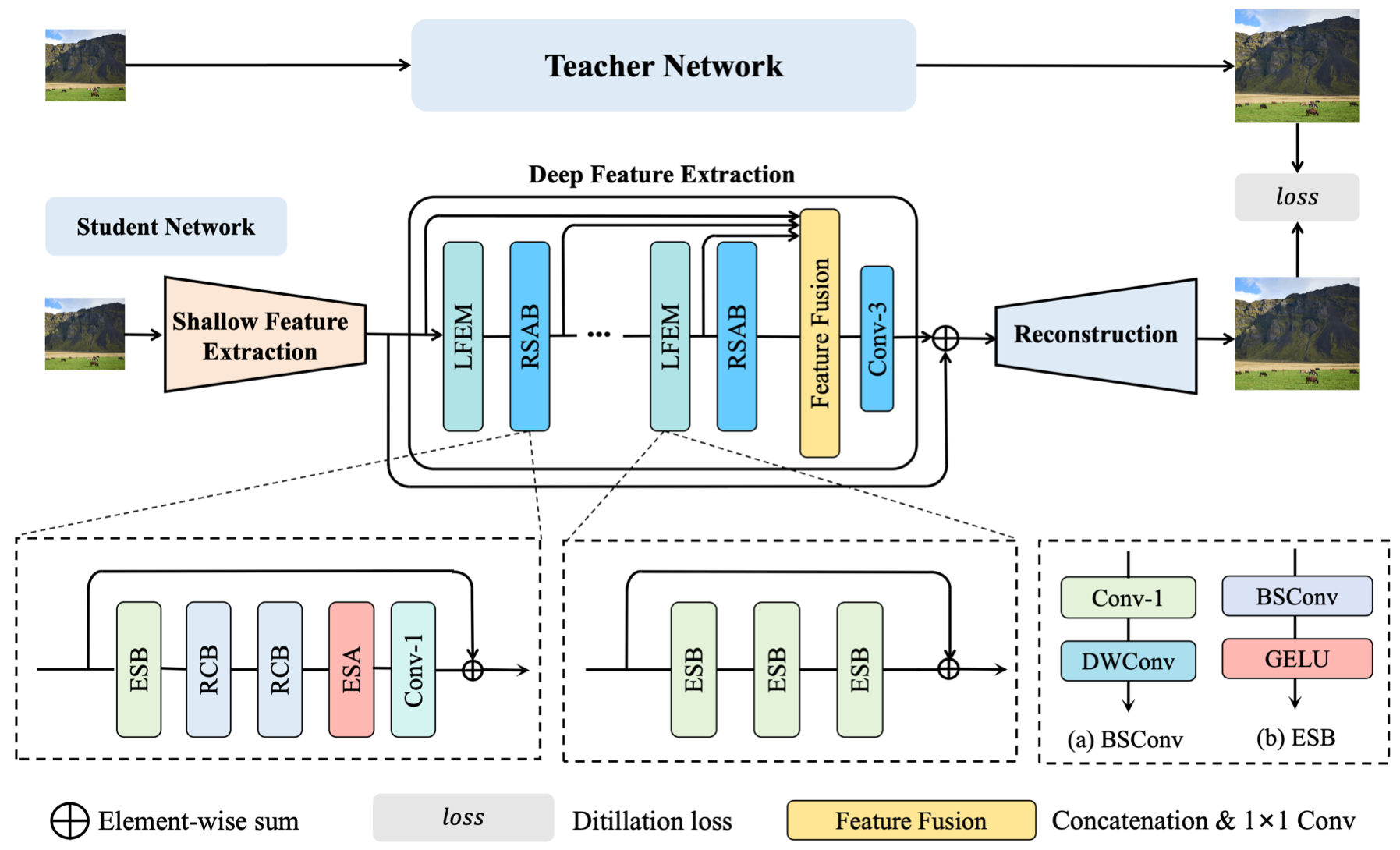}    
    \caption{\textit{Team MViC\_SR:} The pipeline of LSANet. (a) Structure of BSConv. (b) Structure of ESB.}
    \label{fig:image1}
\end{figure*}

\textbf{Method.} As shown in Figure~\ref{fig:image1}, this team propose a network with lightweight spaced attention~(LSANet). The architecture of LSANet consists of the following parts: the shallow feature extraction, the deep feature extraction based on spaced local feature extraction module~(LFEM) and reparamaterized spaced attention Block~(RSAB), and the reconstruction. 
    
Given the LR input $I_{LR}$, a single 3$\times$3 convolution is applied to extract the shallow feature $F_{0} \in \mathbb{R}^{C \times H \times W }$ in the first part:
\begin{equation}
    F_{0}=\operatorname{Conv}\left(I_{LR}\right),
\end{equation}
where $I_{LR}$, $C$, $H$, $W$ are the input LR image, channel dimension, height and width of the input, respectively.
    
In the second part, they use six alternate blocks to extract the deep feature $F_{d} \in \mathbb{R}^{C \times H \times W }$:
\begin{equation}
    F_{d}=H_{D}(F_{0}).
\end{equation}
Specifically, $H_{D}$ is comprised of local feature extraction module~(LFEM) and reparamaterized spaced attention Block~(RSAB). By taking $F_{s}$ and $F_{d}$ as inputs, the HR image $I_{HR}$ is reconstructed with an upsampler as:
\begin{equation}
    I_{HR}=H_{RC}(F_{0}+F_{d}),
\end{equation}
where $H_{RC}$ is the reconstruction module involves a single 3$\times$3 convolution followed by a pixel shuffle operation~\cite{shi2016real}.

Previous methods mostly use several plain convolutions to extract features, which is inefficient with soaring computational complexity. Inspired by the blueprint shallow residual block~\cite{li2022blueprint}, they design an efficient yet effective local feature extraction module~(LFEM) to alleviate computing burden while largely maintaining model performance. As illustrated in Figure~\ref{fig:image1}(a), the LFEM contains three efficient shallow blocks~(ESB) which include a 1$\times$1 point-wise convolution with a 3$\times$3 depth-wise convolution followed by GELU activation~\cite{hendrycks2016gaussian}. 
    
\begin{figure}[t]
    \centering
    \includegraphics[width=0.48\textwidth]{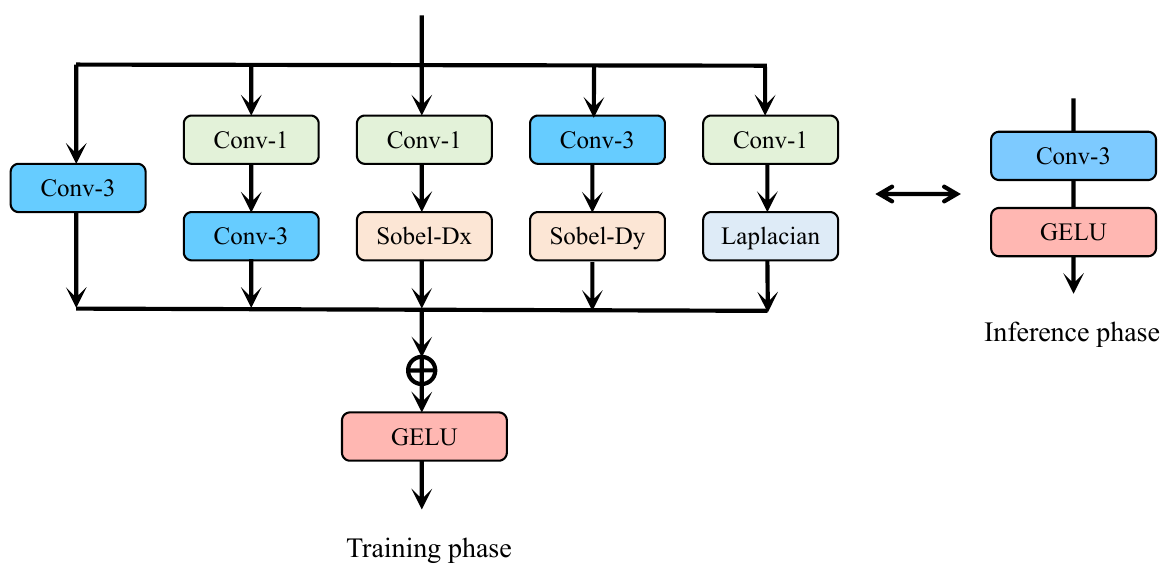}
    \caption{\textit{Team MViC\_SR:} The structure of RCB.}
    \label{fig:image2}
    \vspace{-0.5cm}
\end{figure}

    
Consequently, they propose a reparamaterized spaced attention Block~(RSAB) to preserve the representation ability to the adjacent blocks, which is composed of an efficient shallow block and two reparameterized convolution blocks~(RCB) followed by an enhanced spatial attention block~(ESA)\cite{kong2022residual} and a convolution. The spaced ESA blocks only used in RSAB are employed for comprehensive extraction and modulation of deep features. They use a 1$\times$1 convolution after the ESA block to further refine the weighted feature and capture local patterns. The detailed structure is shown in Figure~\ref{fig:image2}. Intermittently configurating attention blocks substantially lower model complexity without a noticeable performance drop. At the end of this part, features extracted from all blocks are concatenated and aggregated using two convolutions. 

Reparameterization \cite{zhang2021edge, du2022fast} has proven effective in enhancing feature representation without introducing additional computational overhead. Different from the reparameterization module design of high-level tasks, they design an isotropic edge-oriented convolutional block in their model. As shown in Figure~\ref{fig:image2}(a), the Sobel-Dx and Sobel-Dy employ the isotropic Sobel function to improve the representation capabilities of their model. During the inference phase, all branches are combined to a simplified 3$\times$3 convolution, which significantly reduces computation cost.

\noindent\textbf{Training Details.} They apply two stages to train their network on DIV2K and the first 10K data of LSDIR. They randomly augment the input with a flip and 90-degree rotation to enhance the robustness of the network. They design a teacher-student distillation strategy for training, which makes the large version of SAFMN~\cite{sun2023safmn} as a teacher and the proposed network as a student. Three LFEM and RSAB blocks are stacked alternately in their student network, and the feature channel is 36. The mini-batch size is fixed to 64 in all stages. The details of the two training stages are as follows:
\begin{itemize}
    \item \textbf{Stage 1.} They minimize the L1 loss~(student prediction and ground-truth) and distillation loss ~(student prediction and teacher preditction) to optimize the student network by Adam optimizer~\cite{kingma2014adam} for $1000k$ iterations. The initial learning rate is set to 2$\times10^{-3}$, which will be halved at $100k$, $500k$, $800k$, $900k$, $950k$. The HR patch size is 256 during this stage.
    \item \textbf{Stage 2.} They initialize the weight of the student network with the pre-trained student in Stage 1. They enlarge the HR patch size to 640 and minimize MSE loss instead of L1 loss.
\end{itemize}
\subsection{LVTeam}

\begin{figure}[t]
    \centering
    \includegraphics[width=0.28\textwidth]{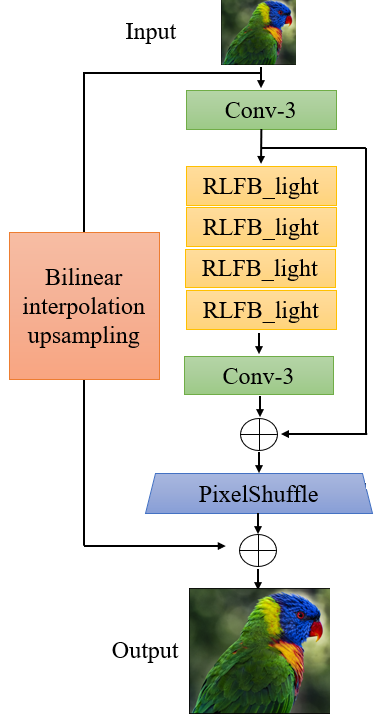}
    \caption{\textit{LvTeam}: The architecture of LightRLFN.}
    \label{fig:LVTeam_arch}
\end{figure}

\textbf{Method.}
Accurately inferring and reconstructing the fine details missing in the LR images based on the learned models of image textures and structures, while maintaining the efficiency of the inference process, poses a challenging task. Since the provided baseline RLFN \cite{kong2022residual} is already lightweight, they wanted to further explore the limits of the RLFN model while keeping the performance (26.99 dB). For this purpose, they made two improvements based on the RLFN and formed a novel architecture LightRLFN. Firstly, a shortcut connection between the inputs and the model outputs. To achieve this, they perform 4$\times$ bilinear interpolation on the input image to match the resolution of the output. Secondly, they reduced the width of the model (46 channels to 36 channels). As a result, the designed LightRLFN model is lightweight and efficient enough, as shown in fig.\ref{fig:LVTeam_arch}.

\noindent\textbf{Training Details.}
The proposed architecture is based on PyTorch 2.2.1 and an NVIDIA 2080Ti with 11G memory. They set
400 epochs for training with batch size 32, using AdamW with $\beta_1=0.9$ and $\beta_2=0.999$ for optimization.
The initial learning rate was set to 0.001, and cosine annealing was used for learning rate adjustment. Regarding the use of the DIV2K \cite{agustsson2017ntire} and LSDIR \cite{lilsdir} datasets, they first copied the DIV2K dataset to 85 times its original number (i.e., 800$\times$85), and then merged it with the LSDIR dataset to form the final training set, which contains a total of 152,991 images. For data augment, they first randomly crop the LR image to 96$\times$96 (the corresponding HR resolution is 384$\times$384) and then perform horizontal flip with probability 0.5. Regarding the loss function, they use the L1 loss and the frequency domain reconstruction loss with 0.1 weights.

\subsection{Fresh}
\begin{figure}[!ht]
    \centering
    \vspace{-6.5mm}
    \includegraphics[width=0.46\textwidth]{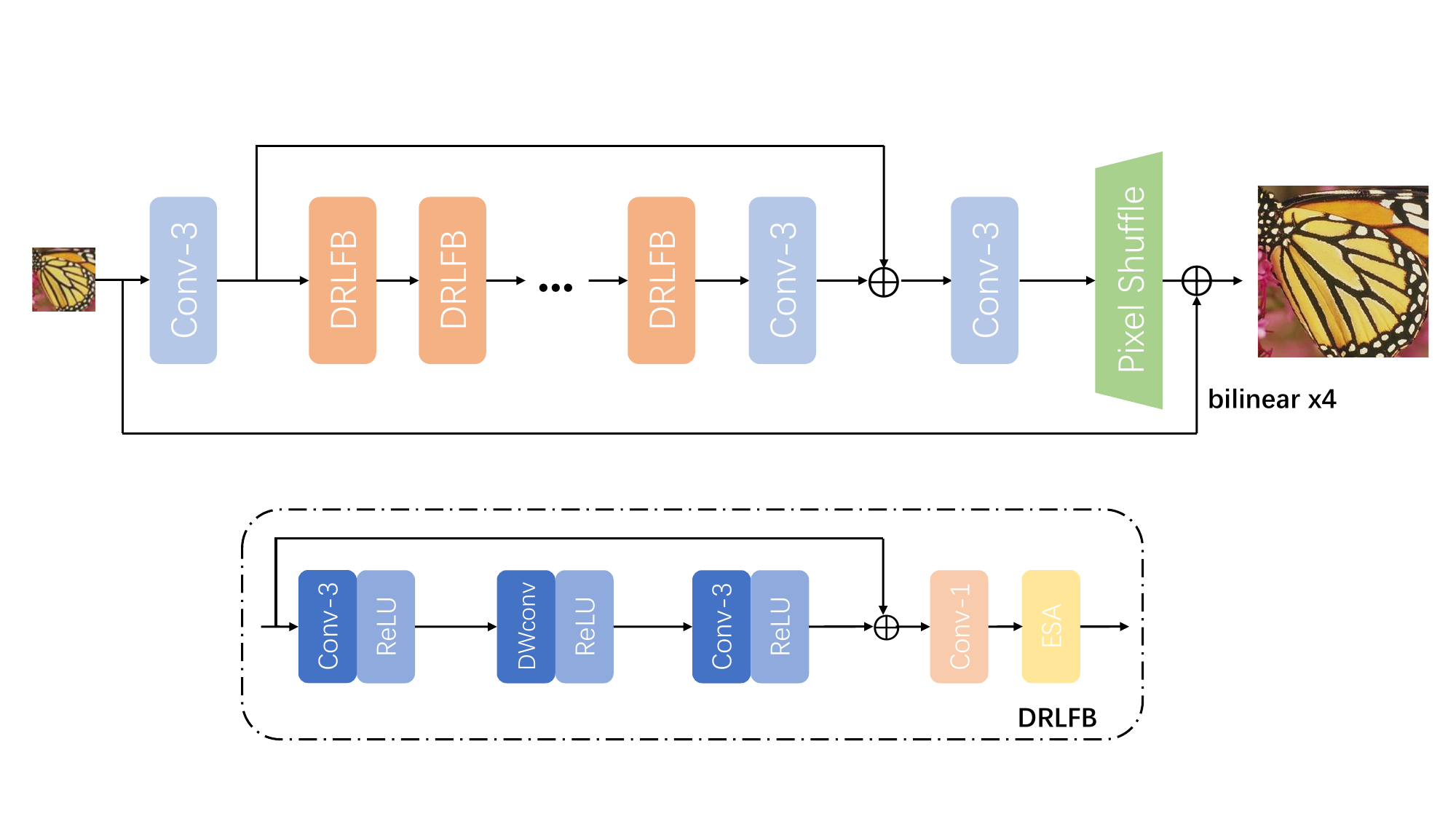}
    \vspace{-2.5mm}
    \caption{\textit{Team Fresh}: The architecture of DepthRLFN.}
    \label{fig:Fresh_arch}
\end{figure}

\noindent\textbf{Method.}
To further reduce the network's parameter count and enhance its efficiency, they propose Depth residual local feature network for super-resolution (DepthRLFN), which is modified from the RLFN \cite{kong2022residual} As illustrated in Figure~\ref{fig:Fresh_arch}, they employ depth-wise separable convolutions to replace the conventional convolutions in the RLFB structure of the baseline, and they add the low-resolution (LR) image subjected to bilinear interpolation before the network's output. This approach not only better preserves the details of the original image but also reduces the model's parameter count and enhances the final image quality. Moreover, to mitigate the discrepancies observed between global operation behaviors during training and testing phases, which adversely affect super-resolution performance, their approach integrates the Test-time Local Converter (TCL) architecture as introduced in ~\cite{kong2022residual}.\\

\noindent\textbf{Training Details.}
They trained DepthRLFN using a total of 85,791 image pairs from the DIV2K and LSDIR \cite{lilsdir} datasets on PyTorch 2.2.1 and an NVIDIA A40 with 40G memory. The training process was divided into two stages, with DepthRLFN comprising 4 DRLFBs and having a channel count of 64. For data augment, they first randomly crop the image to 192$\times$192 and then perform a horizontal flip with probability 0.5. They set 500,000 iterations for training, using AdamW with $\beta_1=0.9$ and $\beta_2=0.999$ for optimization. They set the initial learning rate to \(1\times10^{-4}\) and decay the learning rate by 0.5 every 150, 000 iterations. Additionally, they use L1 loss and Frequency loss for the training phase. The cropped HR image size is 256$\times$256 and the mini-batch size is set to 64 for the finetuning stage. The DepthRLFN is trained by L2 loss for a total of 800,000 iterations. They set the initial learning rate to \(1\times10^{-5}\)  and decay the learning rate by 0.5 every 250,000 iterations.
\subsection{Lanzhi}
\begin{figure}[ht]
  \centering
  \includegraphics[width=\columnwidth]{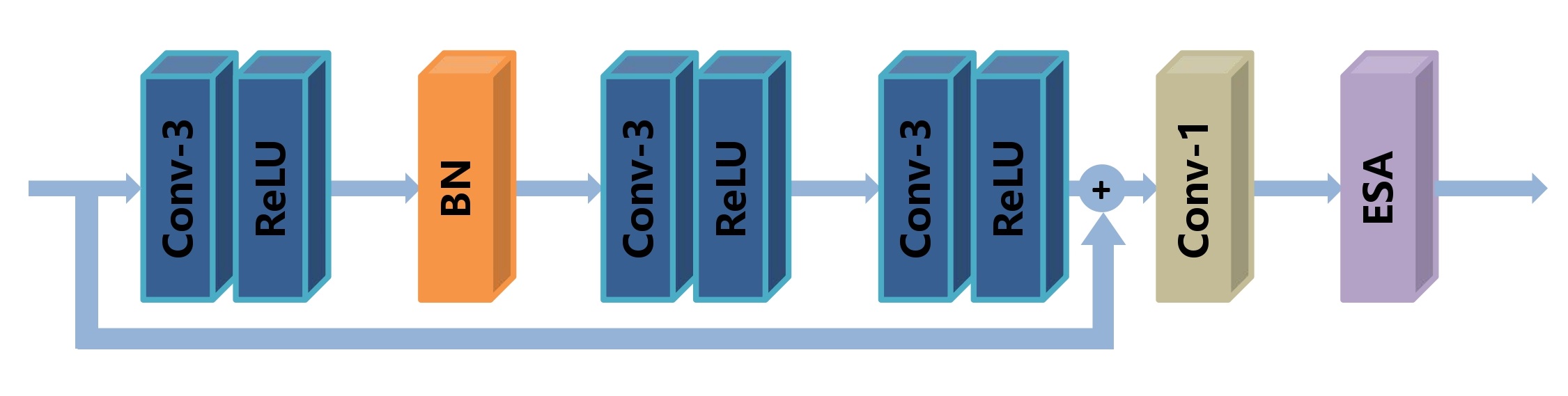}
  \caption{\textit{Team Lanzhi:} RLFB\_BN: Residual local feature block with embedded batch normalization layer.}
  \label{fig:lanzhi_model}
\end{figure}

\noindent\textbf{Method.}
In order to facilitate more stable model training, they employ an expedited convergence rate and simultaneously enhance the generalization capability of the model. As shown in Figure \ref{fig:lanzhi_model}, they introduce a batch normalization layer after the first CONV3+RELU layer in the residual local feature blocks (RLFBs) within the Residual Local Feature Network (RLFN) architecture~\cite{kong2022residual}. 
By incorporating BN before each convolution, every convolutional layer can benefit from the normalized input features. Additionally, the final convolutional layer can make certain adjustments to pixel biases before incorporating them into the next branch, thereby enhancing performance to a certain extent~\cite{lanzhi3}.
Ensemble learning is a powerful technique to improve model performance and robustness. In the inference stage, they utilize model fusion as a means of its implementation. Specifically, they employ a weighted averaging strategy to fuse the prediction results of the models. By adjusting the weight values, they can flexibly control the influence of each model. Through experiments, they find that this methodology can significantly improve the effect of image super-resolution. They conducted a series of experiments and comparisons between RLFN\_BN and the original RLFN model. Through modifications to the RLFN architecture, they find that the runtime, FLOPs, and number of parameters remained comparable. On top of that, with the application of ensemble learning strategy, their approach achieved a certain improvement in the peak signal-to-noise ratio (PSNR) while maintaining comparable structural similarity (SSIM) scores, further demonstrating the feasibility and effectiveness of their approach for efficient image super-resolution.\\

\noindent\textbf{Training Details.}
During the training phase, they utilize the training data from the DIV2K dataset, comprising 800 pairs of low-resolution and high-resolution images, as well as a subset of the LSDIR dataset~\cite{lilsdir}, consisting of randomly selected 8,500 pairs of low-resolution and high-resolution images. Before training, they pre-process the images by decoding all PNG files and saving them as binary files. The training process employs the Adam optimizer. On the other hand, they employ a learning rate decay strategy to optimize training stability and adapt to changes in the data distribution. The initial learning rate is set to $2 \times 10^{-4}$. The total number of epochs is 1000. The other training configurations, such as the batch size is set as 16 and the RGB range is set as 255. Additionally, they utilize the L1 loss function as it tends to generate sharper images compared to the L2 loss. The implementation of their approach is carried out using the PyTorch framework.
\subsection{Supersr}
\begin{figure}[t]
\centering
\includegraphics[width=1\linewidth]{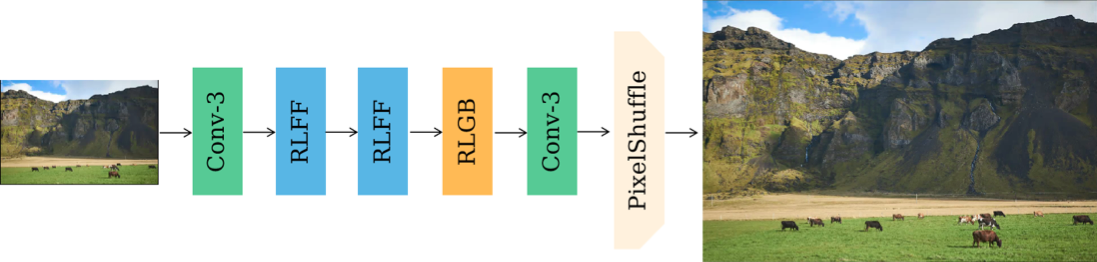}
\caption{\label{framework_team05}\textit{Team Supersr:}The overall network architecture of our RLFF.}
\end{figure}

\begin{figure}[t]
\centering
\includegraphics[width=1\linewidth]{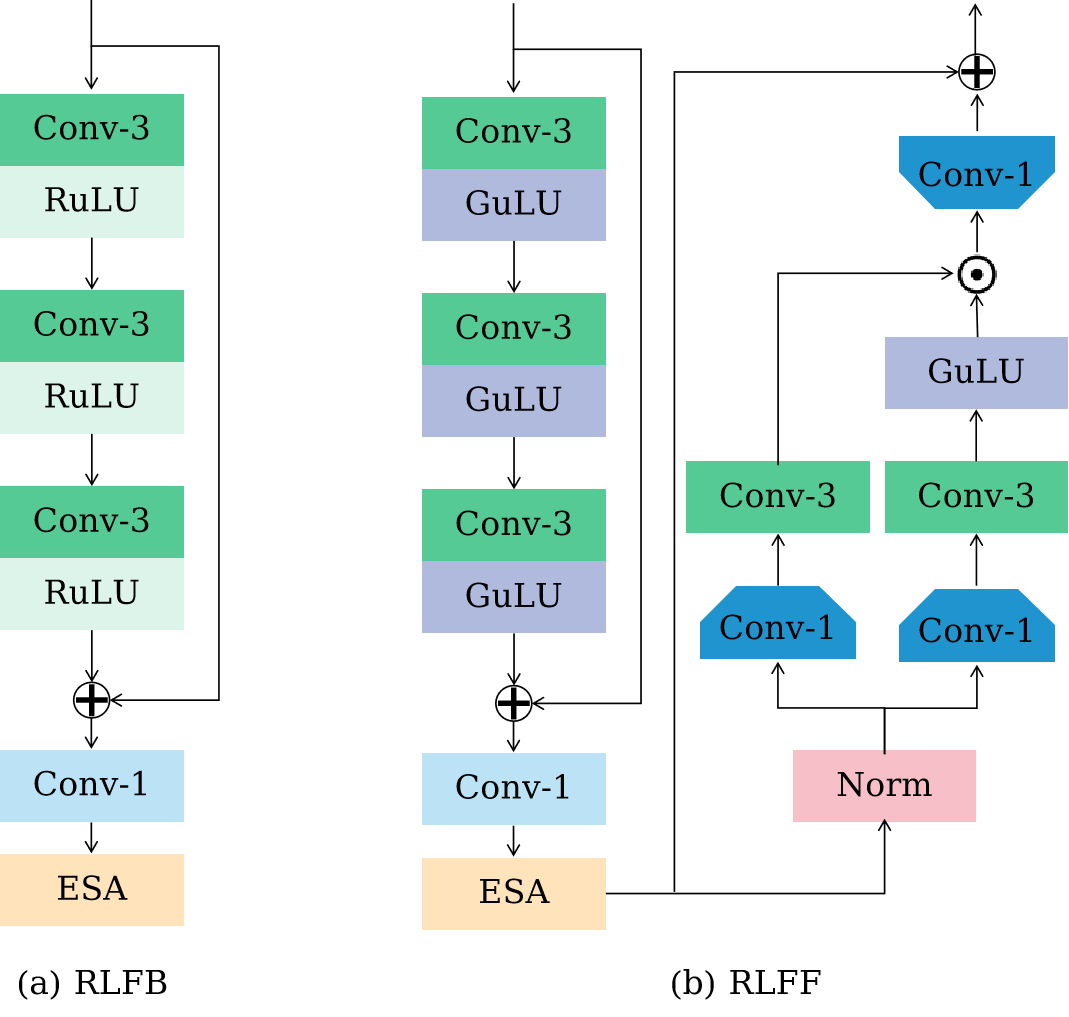}
\caption{\textit{Team Supersr:} (a) The original RLFB module. (b) Our RLFF module.}
\end{figure}

\noindent\textbf{Method.}
It is designed based on the baseline approach RLFN \cite{kong2022residual}. As depicted in Fig. \ref{framework_team05}, they replace the original RLFB block with their Residual Local Feed Forward (RLFF) block, which is efficient for super-resolution tasks. The detailed information will be introduced in the next section. The residual Local Feed Forward (RLFF) block is the core block in our framework. Firstly, they use the GeLU as the activation function. As presented in Restormer \cite{zamir2021restormer}, the Gated Dconv Feed-Forward Network, by managing the flow of information across the hierarchical levels within the pipeline, empowers each level to concentrate on intricate details that complement those addressed by other levels. Inspired by this, they put it into the basic module as a subsequent processing of features to improve the feature learning capability of the network further.

\noindent\textbf{Training Details.}
Two datasets provided are used for training: DIV2K~\cite{agustsson2017ntire} and LSDIR~\cite{lilsdir}. They use the augmentation strategy of vertical/horizontal flips and 90-degree rotation. They adopt the Adam optimizer algorithm, $\beta_1 = 0.9$, and $\beta_2 = 0.99$. The batch size and patch size are set to 64 and 256. All experiments are conducted on a single NVIDIA RTX 3090 GPU. Only the L1 is used during the training process. It consists of three phases: 1. patch size/learning rate: 256/2e-3 for 100, 000 iterations; 2. patch size/learning rate: 384/1e-3 for 50, 000 iterations; 3. patch size/learning rate: 512/5e-4 for 50, 000 iterations.
\subsection{MeowMeowMeow}
\begin{figure}[t]
    \centering 
    \includegraphics[width=1\linewidth]{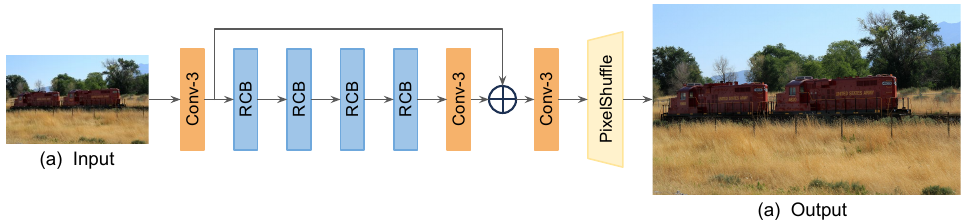} 
    \caption{\textit{Team MeowMeowMeow:} Reparameterized Convolutional Network (RCN).}
    \label{fig:main}
\end{figure}

\noindent\textbf{Method.}
Their model is designed based on the baseline approach RLFN~\cite{kong2022residual}, as depicted in Figure~\ref{fig:main}. They replace the RLFB block with a modified version, which they call Reparameterized Convolutional Block (RCB), incorporate two extra Layer Norm layers, and replace ReLU with GELU. Inspired by ECBSR~\cite{zhang2021edge}, they employ Edge-oriented Convolutional Block (ECB) as their reparameterization module to replace the original 3$\times$3 convolution layer. In the training phase, the ECB module adds five extra branches to the original 3$\times$3 convolution layer as shown in Figure~\ref{fig:main}. In the inference phase, the ECB module can be converted into a 3$\times$3 convolution layer without any computational overhead.  

\noindent{\textbf{Training Details.}} They use DIV2K~\cite{agustsson2017ntire} and LSDIR~\cite{lilsdir} datasets for training, as well as vertical/horizontal flips and 90-degree rotation as data augmentation. AdamW is leveraged as the optimizer with a weight decay of 0.01, $\beta_1 = 0.9$, and $\beta_2 = 0.99$. EMA decay is set to 0.999, and batch size is set to 128. They use a warmup iteration of 2000. All training was conducted on RTX A6000 GPU. They used three functions, including standard L1 and L2 loss, and our proposed PSNR loss. The whole training procedure contains four stages. Inspired by PEFT techniques~\cite{han2024parameter,han2023zero}, they employ LoRA with rank $r=4$ to the last stage. Tunable hyperparameters include patch size (PS), learning rate (LR), scheduler (Sche), training iterations (Iter), and loss, across the four stages. They summarize these hyperparameters in Table~\ref{tab:hyper} and Table~\ref{tab:scheduler_hyper}.
\begin{figure}[t]
    \centering 
    \includegraphics[width=1\linewidth]{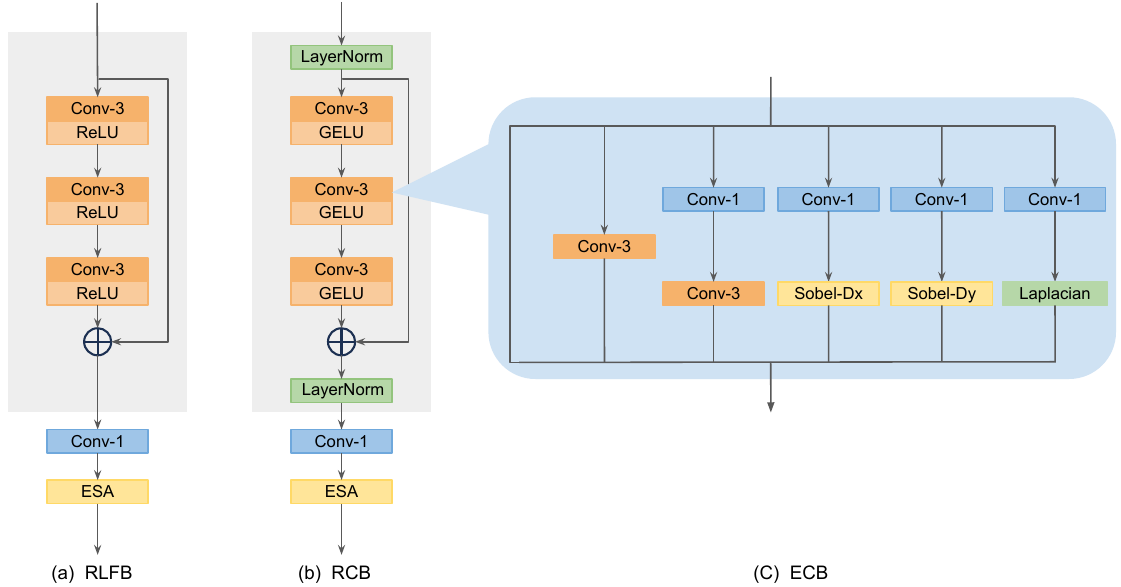} 
    \caption{\textit{Team MeowMeowMeow:} Reparameterized Convolutional Block (RCB) and Edge-oriented Convolutional Block (ECB).}
    \label{fig:arch}
\end{figure}

\begin{table}[t]
\centering
\caption{\textit{Team MeowMeowMeow:} Hyperparameters for different training stages.}
\label{tab:hyper}
\resizebox{\columnwidth}{!}{%
\begin{tabular}{lcccc}
\hline
\multicolumn{1}{c}{} & Stage 1      & Stage 2                  & Stage 3                  & Stage 4                  \\ \hline
PS                   & 196          & 256                      & 384                      & 512                      \\
LR                   & $2e-3$         & $2e-3$                     & $1e-3$                     & $2e-4$                     \\
Sche                 & None (fixed LR) & CosineAnnealingRestartLR & CosineAnnealingRestartLR & CosineAnnealingRestartLR \\
Iter                 & 200,000      & 50,000                   & 50,000                   & 20,000                   \\
Loss                 & 1*L1         & 1*L1 + 0.001*PSNR        & 1*L2 + 0.001*PSNR        & 1*L2 + 0.001*PSNR        \\ \hline
\end{tabular}
}
\end{table}

\begin{table}[!ht]
\centering
\caption{\textit{Team MeowMeowMeow:} Hyperparameters of CosineAnnealingRestartLR scheduler for stage 2, stage 3 and stage 4.}
\label{tab:scheduler_hyper}
\resizebox{\columnwidth}{!}{%
\begin{tabular}{lccc}
\hline
Parameter & Stage 2 & Stage 3 & Stage 4 \\
\hline
Periods & [12500, 12500, 12500, 12500] & [12500, 12500, 12500, 12500] & [20000] \\
Restart\_weights & [1, 0.9, 0.8, 0.7] & [1, 0.9, 0.8, 0.7] & [1] \\
Eta\_min & $1e-7$ & $1e-7$ & $1e-7$ \\
\hline
\end{tabular}
}
\end{table}
\subsection{Just Try}
\begin{figure}
    \centering
    \includegraphics[width=1\linewidth]{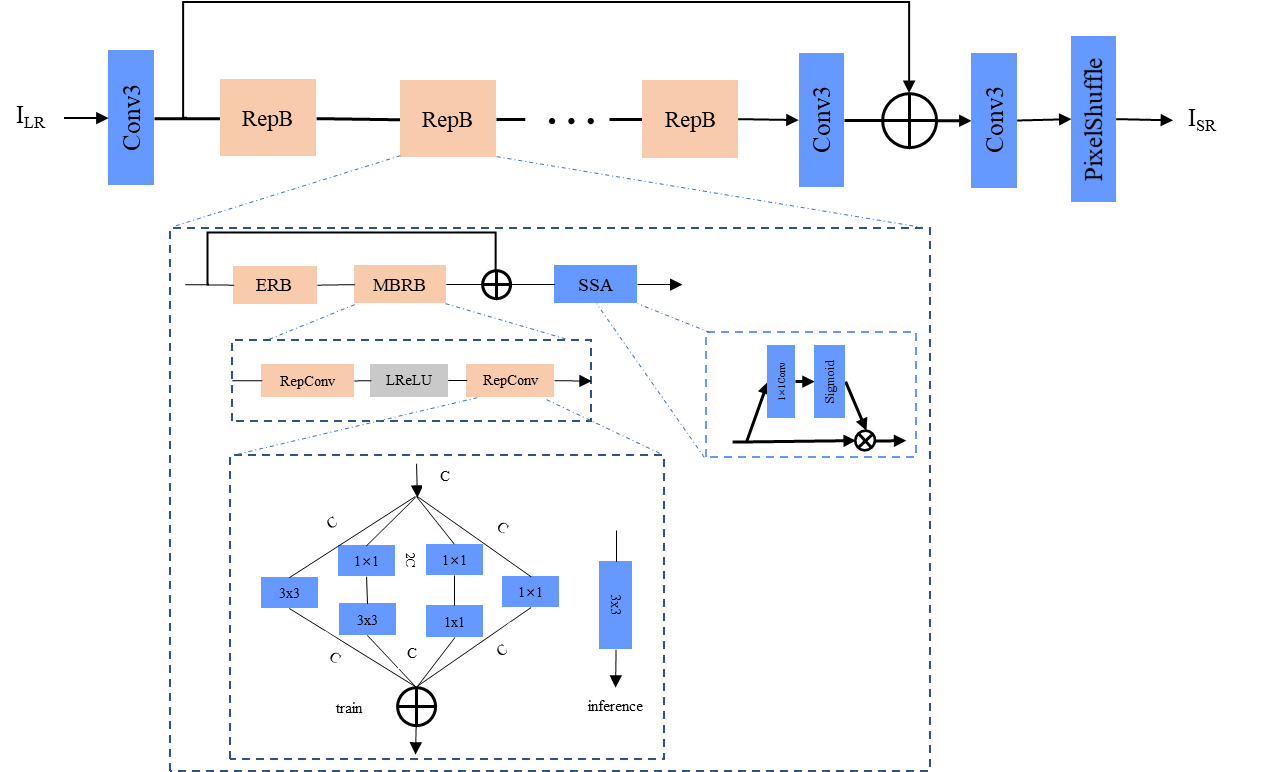}
    \caption{\textit{Team Just Try: }Overall structure of ERRN}
    \label{fig:enter-label}
\end{figure}

\textbf{Method}. Inspired by RLFN~\cite{kong2022residual}, FMEN~\cite{du2022fast}, and DBB~\cite{DBB}, they designed a Enhanced Reparameterize Residual Network(ERRN) as shown in the Fig.~\ref{fig:enter-label}. The reparameterize block (RepB) consists of ERB~\cite{du2022fast}, multi-branch reparameterize block (MBRB), and simple spatial attention(SSA). The ERB and MBRB are both reparameterization blocks, which use a complex RepConv structure during the training phase and convert to a 3$\times$3 convolutional during inference. Every RepB uses SSA to enhance the output feature. In ERRN, they use six RepBs, and the number of feature channel C is set to 40.

\noindent\textbf{Training Details}. They use DIV2K, LSDIR, and Flickr2K datasets as training datasets. For each mini-batch, they randomly crop 16 patches from the LR images with the size of 64$\times$64. They use a cosine annealing learning scheme, the learning rate is initialized as 2×$10^{-4}$ and the minimum learning rate is 1$\times$$10^{-7}$, a total of 1000k iterations, the period of cosine is 250k iterations. They use Adam optimizer with$\beta_1$= 0.9,$\beta_2$=0.99. The loss function is L1 loss. Then fine-tuning on the same datasets, the LR size is set to 128$\times$128, the initialized learning rate is 1$\times$$10^{-4}$, a total of 400k iterations, and the period of cosine is 100k iterations. Other settings are the same as above. Final L2 loss is used for fine-tuning on the same datasets. Other settings are the same as above.  
\subsection{VPEG\_E}

\begin{figure}
	\centering
	\includegraphics[width=0.48\textwidth]{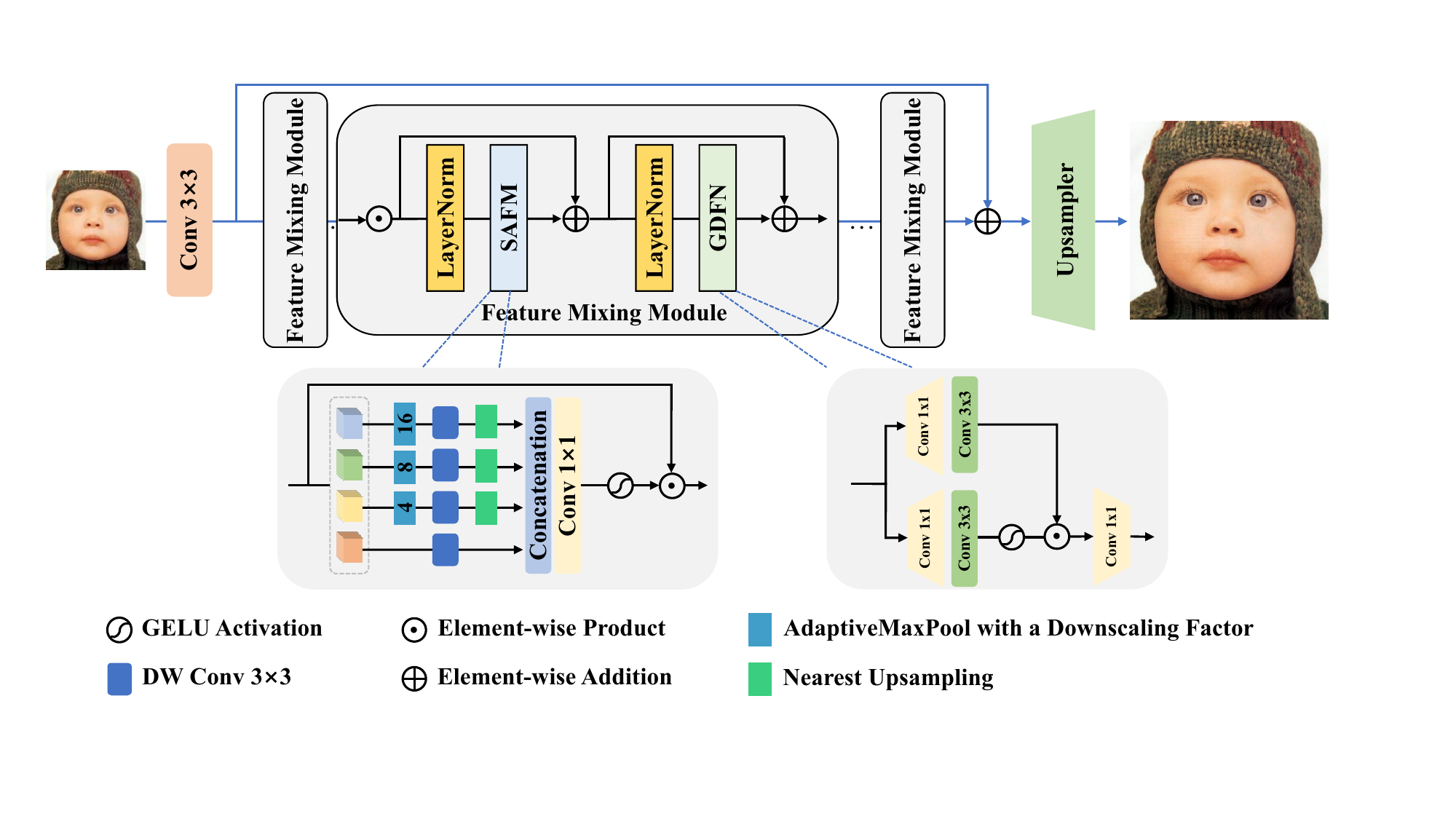}
	\vspace{-3mm}
	\caption{\textit{Team VPEG\_E}: The overall network architecture of our proposed EGFMN.}
	\label{fig:framework_team43}
\end{figure}

\textbf{Method.}
The VPEG\_E team introduces an enhanced gated feature modulation network (EGFMN) for efficient SR, which is modified from the SAFMN~\cite{sun2023safmn}.
To make the EGFMN more lightweight, the VPEG\_E team replaced the used convolutional channel mixer (CCM) with the gated-dconv feed-forward network~\cite{zamir2021restormer} (GDFN).
Figure~\ref{fig:framework_team43} shows that EGFMN first uses a convolution layer maps the input image to feature space
and employs 7 feature mixing modules (FMMs) for learning discriminative feature representation, where each FMM block has a spatially-adaptive feature modulation (SAFM) layer and a GDFN module.
To recover the HR target image, the VPEG\_E team introduced a global residual connection to learn high-frequency details and employ a lightweight upsampling layer for fast reconstruction, which only contains a 3×3 convolution and a pixel-shuffle~\cite{shi2016real} layer.

\noindent\textbf{Training Details.}
The VPEG\_E team trains the proposed EGFMN on the LSDIR dataset. 
The cropped LR image size is $640\times640$ and the mini-batch size is set to 64. 
The EGFMN is trained by minimizing L1 loss and the frequency loss~\cite{cho2021rethinking} with Adam optimizer for total of 600, of 000 iterations. 
The VPEG\_E team set the initial learning rate to $2\times10^{-3}$ and the minimum one to $1\times10^{-6}$, which is updated by the Cosine Annealing scheme~\cite{SGDR}.
\subsection{BU-ESR}
\begin{figure}[htbp]
\centering
\includegraphics[width=\columnwidth]{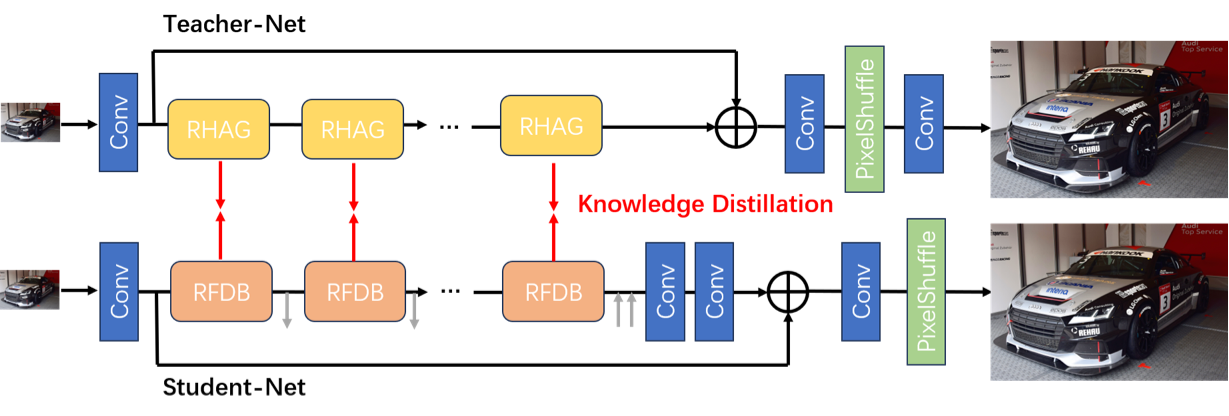}
\caption{\textit{Team BU-ESR: }Their teacher network is designed based on a Hybrid Attention Transformer backbone, while their student network is implemented using an RFDN network. The dilation loss provides latent space feature supervision to enhance the lightweight RFDN's image super-resolution results.}
\label{team30_buesr_pipline}
\end{figure}

\textbf{Method.}
Inspired by~\cite{teacher_student_ct}, this team employs a knowledge distillation method, as shown in Fig.\ref{team30_buesr_pipline}. Initially, the team utilizes a substantial teacher network equipped with a Hybrid Attention Transformer backbone~\cite{chen2023activating}. This teacher network is specifically designed to learn and distill high-quality features. Its augmented parameter capacity is crucial for capturing nuanced feature information, thereby providing a robust foundation for supervising the student network.
In the second stage, this team further trains the student network under the supervision of the pretrained features from the teacher network to enhance its performance. Considering that architectures based on self-attention significantly increase the model's parameter count~\cite{self_att_1,self_att_2}, their work is meticulously designed around the Residual Feature Distillation Network (RFDN) model, which serves as the backbone for the student network. The RFDN framework is notable for its lightweight structure, consisting of four pivotal components: an initial feature extraction convolution, multiple Residual Feature Distillation Blocks (RFDBs) stacked sequentially, a feature fusion layer, and a final reconstruction block. The process begins with a $3 \times 3$ convolutional layer that extracts coarse features from the input low-resolution (LR) image. Then, the core of the RFDN, comprising four RFDBs, is sequentially employed by the team to progressively refine these features. The refined features from each RFDB are amalgamated using a $1 \times 1$ convolution layer, followed by an additional $3 \times 3$ convolutional layer to enhance the smoothness of the aggregated features. Finally, the high-resolution image is produced through a pixel shuffle operation by the team.


\noindent\textbf{Training Details.} This team utilized two datasets, DIV2K and LSDIR, for their experiments. They augmented the training dataset with geometric transformations, including vertical and horizontal flips and 90-degree rotations, to enhance the model's comprehensive abilities. During the Teacher Model Training Phases, the model was initially trained from scratch. High-resolution (HR) patches of size $192 \times 192$ were randomly cropped from HR images, with a mini-batch size of 16. The training employed the smooth L1 loss function and the Adam optimizer. Considering the impact of the choice of learning rate on the results \cite{qlabgrad}, the team made multiple adjustments to the initial learning rate and finally set it at an initial learning rate of $2 \times 10^{-4}$. The training spanned 40,000 epochs. Subsequently, the student model was fine-tuned using pre-trained weights, with the initial learning rate reduced to $1 \times 10^{-4}$, and the training ran for an additional 100 epochs. This stage retained the same settings as the initial phase but incorporated a loss function that includes smooth L1, MS-SSIM loss, perceptual loss, and teacher-student dilation loss.
\subsection{Lasagna}

\begin{figure*}[t]
	\centering
	\includegraphics[width=0.8\textwidth]{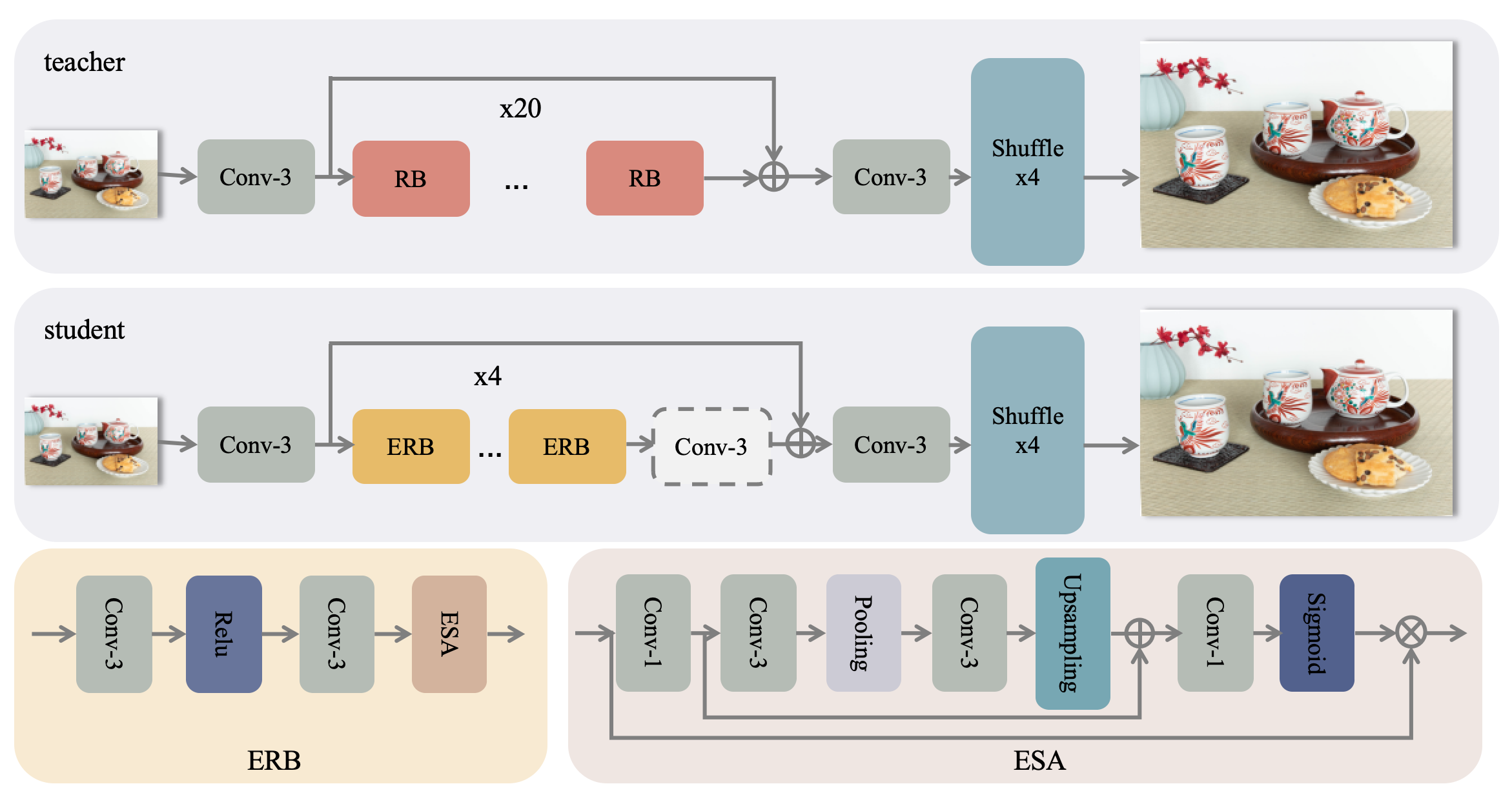}
	\caption{\textit{Team Lasagna: } The structure of the proposed EERN.}
	\label{Fig.structure}
\end{figure*}

\noindent\textbf{Method.}
This team propose an efficient enhanced residual network (EERN) for efficient image super-resolution, the primary architecture of which is illustrated in the Fig.\ref{Fig.structure}. Although the EDSR\cite{lim2017enhanced} network achieves high performance in the field of single-image super-resolution, its network structure is relatively bulky. Therefore, they have adopted a strategy of reducing the number of blocks to lighten the model. However, merely decreasing the number of blocks are not able to meet the performance standards set by the competition. Therefore, they have made certain adjustments to its structure by incorporating an ESA\cite{liu2020residual}  after each block, and to reduce the parameters, they have eliminated several convolutional layers. Additionally, they appended a convolutional layer at the end of the ERB block for the purpose of fine-tuning. This convolutional layer was not included at the onset of the training process but was incorporated during the final fine-tuning phase to participate in the computation. Simultaneously, to increase inference speed and reduce GPU memory usage,  this team  have removed the residual connections in the RB modules of EDSR. Experiments show that removing residual connections has minimal impact on the module's performance. Moreover, considering that excessive changes in the number of feature channels are detrimental to performance in lightweight networks, they opted to replace two successive shuffle2 operations with a single shuffle4 operation to avoid the final convolutional reduction of channels from 256 to 3. Lastly, they trained the network using a knowledge distillation approach. Due to the mismatch in the number of feature channels between the teacher model and the student model, they  did not utilize the loss of feature maps. Instead, the loss was calculated based on the outputs of both models.

\noindent\textbf{Training Details.}
The number of ERB modules and the number of its feature channels were set to 4 and 84, respectively. They trained a total of 1700 epochs to bring the model to convergence. The process was divided into three stages. In the first stage, they initiated the learning rate at 1e-4, employing a cosine annealing strategy to decrease the learning rate to 5e-7 by the 400th epoch. During this cycle, they utilized L1 loss, with the dataset limited to DIV2K. In the second stage, the starting learning rate was set to 1e-5, again using a cosine annealing method to reduce the learning rate to 5e-7 by the 300th epoch. Moreover, in this cycle, PSNR loss was employed for fine-tuning, and the dataset was expanded to include both DIV2K and folders named 0001000 to 0010000 from the LSDIR dataset. In the final stage, the convolutional layer was introduced and initialized to zero. This stage comprised a total of 1000 epochs, with all other settings being identical to those of the second stage.
\subsection{BlingBling}
\begin{figure*}[!htbp]
\centering
\includegraphics[width=6.5in]{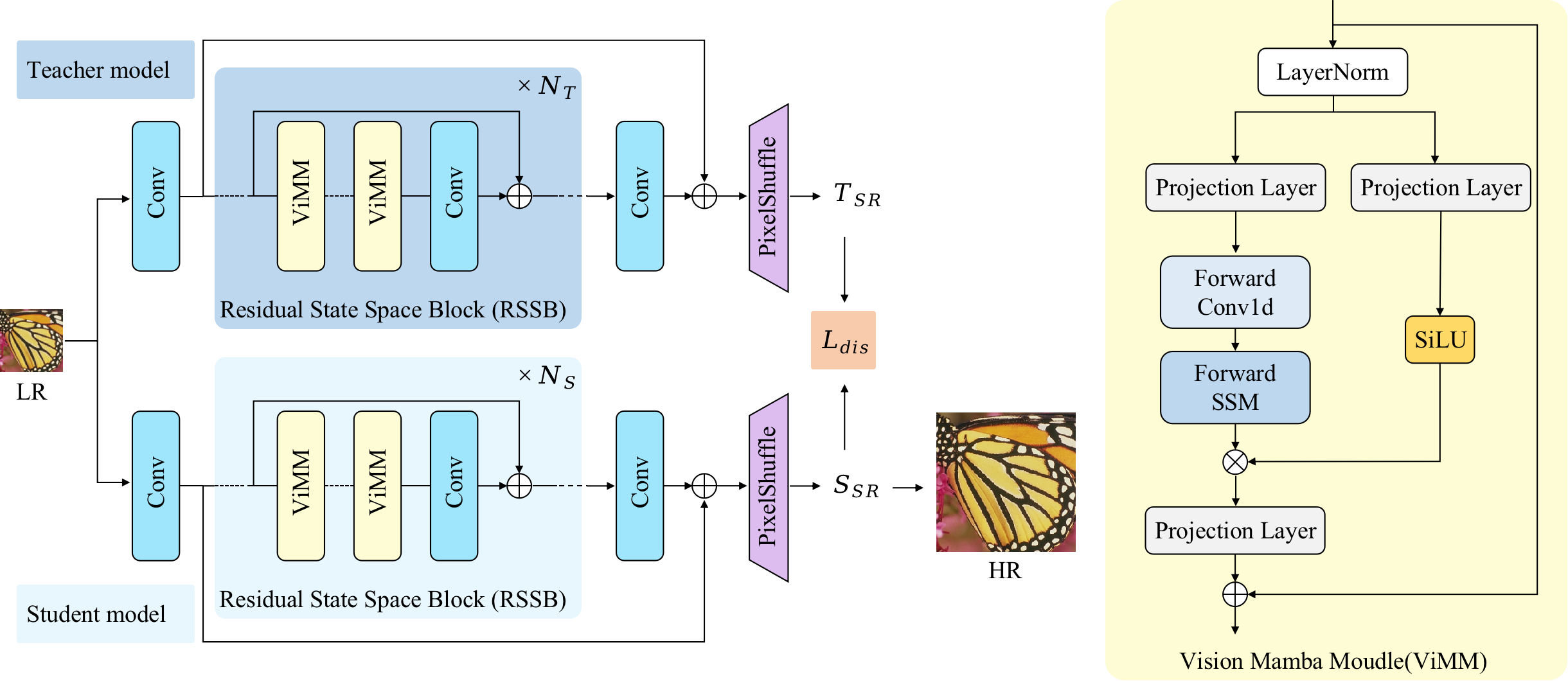}
\caption{\textit{Team BlingBling:}The overall network architecture of our DVMSR, as well as  Vision Mamba Module (ViMM).}
\label{fig:dvmsr}
\end{figure*}

\noindent\textbf{Method.}
In this work, the BlingBling team propose Distillated Vision Mamba SR (DVMSR), a novel lightweight Image SR network that incorporates Vision Mamba and a distillation strategy. The framework of their DVMSR is illustrated in Fig.\ref{fig:dvmsr}. It consists of three main modules: feature extraction convolution, multiple stacked Residual State Space Blocks (RSSBs), and a reconstruction module. The feature extraction convolution employs a $3\times 3$ convolutional layer to extract shallow features. Subsequently, deep features are extracted by several stacked Vision Mamba Modules (ViMMs)~\cite{mamba,vim}, which leverage the long-range modeling capability of Mamba to effectively reduce computational complexity. A $3\times 3$ convolutional layer refines the features extracted from the ViMMs at the end of each RSSB. The final stage involves the aggregation of shallow and deep features using a long skip connection, followed by the use of a sub-pixel convolution layer to upsample the feature and reconstruct the high-quality image. The teacher and student models have similar network structures, with variations in the number of RSSBs, ViMMs, and channels.

\noindent\textbf{Training Details.}
They employ DIV2K~\cite{DIV2K} and LSDIR~\cite{lilsdir} to construct the training dataset. 
The High-Resolution (HR) images are cropped to $256\times 256$ patches for the training procedure. 
They use $\mathcal{L}_{1}$ loss with the Adam optimizer for the network optimization.
The initial learning rate is set to $2\times 10^{-4}$. 
The total number of iterations is 500k. They adopt a multi-step learning rate strategy, where the learning rate will be halved when the iteration reaches 250,000, 400,000, 450,000, and 475000, respectively. In the teacher learning phase, they utilize the DIV2K dataset with 2K resolution to train the teacher network, which comprises 6 RSSB and 2 ViMM blocks with 180 channels. The teacher network can learn the rich representation knowledge for the distillation stage. During the distillation training phase, they merge the DIV2K and LSDIR datasets for the student network, which contains 4 RSSB and 2 ViMM blocks with 60 channels. The teacher network remains fixed during this process. They employ $\mathcal{L}_{1}$ loss to align the student network feature with the teacher network feature. This process can transfer the knowledge of the teacher network to the student network. Formally,

\begin{equation} \label{eq1}
  \begin{split}
  & \mathcal{L}_{out}= \lambda _{dis} \mathcal{L}_{dis}+\lambda _{1}\mathcal{L}_1, \\
  & \mathcal{L}_{dis} = \left \|  \mathcal{T}(I_{LR})-\mathcal{S}(I_{LR})\right \|_{1},\\
  & \mathcal{L}_1= \left \| I_{HR}-\mathcal{S}(I_{LR})\right \|_{1},\\
  \end{split}
\end{equation}
where $\lambda _{dis}$ and $\lambda _{1}$ represents the coefficient of the $\mathcal{L}_{dis}$ loss function and the coefficient of the $\mathcal{L}_{1}$ loss function, respectively. They are set 1. $\mathcal{T}$ represents the function of our teacher network and $\mathcal{S}$ denotes the function of our proposed network. $I_{LR}$ and $I_{HR}$ are the input LR images and the corresponding ground-truth HR images, respectively. More information of $\mathcal{L}_{dis}$ can be seen from Fig.\ref{fig:dvmsr}. 

\subsection{Minimalist}

\begin{figure*}[!ht]
  \includegraphics[width=1.\textwidth]{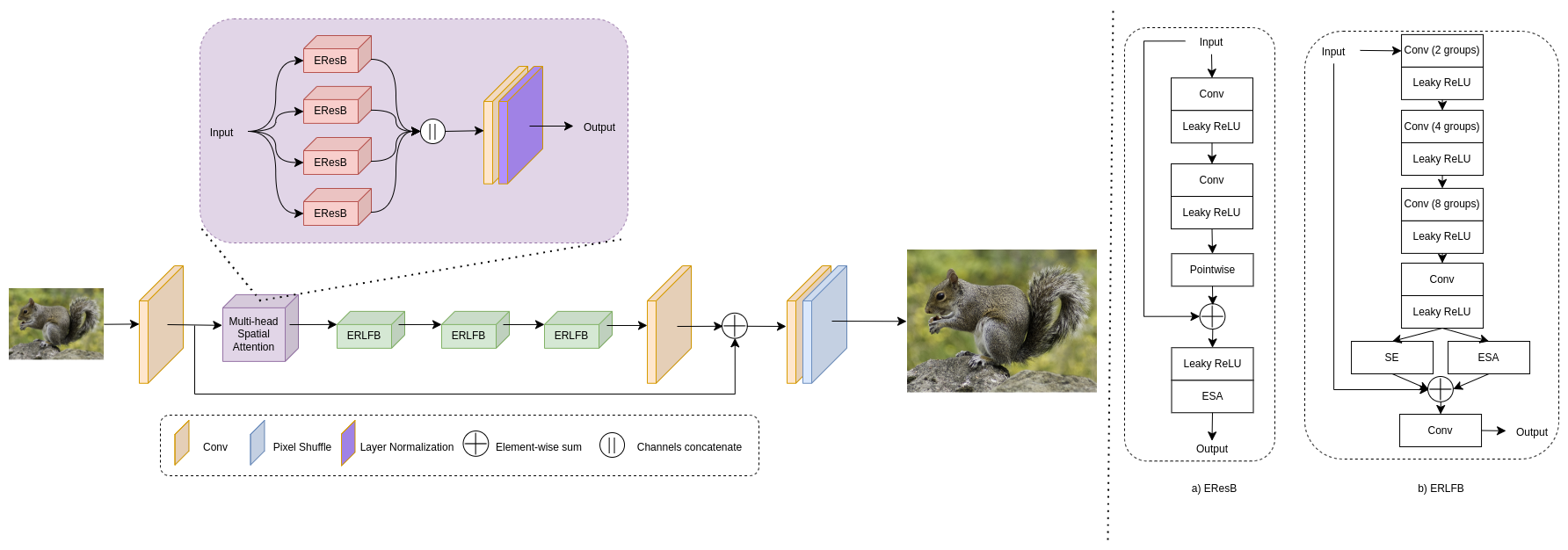}
  \caption{\textit{Team Minimalist: }The overall EMaxGMan architecture is shown on the left side, with a) EResB and b) ERLFB on the right side.}
  \label{fig:team15}
\end{figure*}

\textbf{Method}. Given the inherently lightweight nature of the baseline Residual Local Feature Network (RLFN), the Minimalist aims to enhance its capabilities by introducing strategic modifications to its layers and incorporating additional methods aimed at enhancing its capacity to capture wider spatial and more semantically rich information. A pivotal contribution to their architectural advancements lies in the integration of a novel module termed Efficient Multi-head Spatial Attention (MHSA), alongside a refined iteration of the Residual Local Feature Block (RLFB), drawing inspiration from the foundational work presented in RLFN \cite{kong2022residual}. The comprehensive layout of their team proposed methodology is depicted in Figure \ref{fig:team15}. The proposed Efficient Multi-head Spatial Attention (MHSA)  mechanism introduces a novel approach by partitioning feature maps into multiple heads. Each head is dedicated to capturing spatial dependencies within distinct groups, thereby facilitating a more thorough and efficient representation of features. This concept of spatial attention draws inspiration from prior work such as \cite{kong2022residual}, while the integration of multi-head convolution mixture-normalization is similarly influenced by the findings in \cite{vaswani2017attention}. The incorporation of spatial attention across various groupings within different heads enables the model to discern diverse features at different group levels. Subsequently, they leverage the distilled knowledge from multi-head learning and employ layer normalization to facilitate effective learning with enhanced gradient smoothness. Additionally, their Efficient Residual Local Feature Block (ERLFB) iteratively enhances feature representations through the utilization of local spatial attention and channel-wise attention mechanisms, dynamically assigning priority to specific spatial and channel dimensions. To manage the potential increase in parameters resulting from the introduction of new layers, they integrated group convolution, effectively balancing the trade-off between parameter efficiency and performance, thereby maintaining inference speed comparable to that of the RLFB block.

\noindent\textbf{Training Details.} The Low-Resolution (LR) images are cropped to $48 \times 48$ patches for the training procedure. They use MSE loss with the Adam optimizer for the network optimization. The initial learning rate is set to $0.001$. The total number of epochs is $500$ and the batch size is $64$. To prevent the explosion of gradients they apply the threshold $10$ for the clip gradient norm. They adopt a multi-step learning rate strategy, where the learning rate will decrease to $60$\% of the current learning rate after each $50$ epoch. Data augmentation strategies included horizontal flips, and random rotations of $90$, $180$, and $270$ degrees.

\subsection{MagicSR}

\begin{figure}[t]
  \centering
  \includegraphics[width=0.47\textwidth]{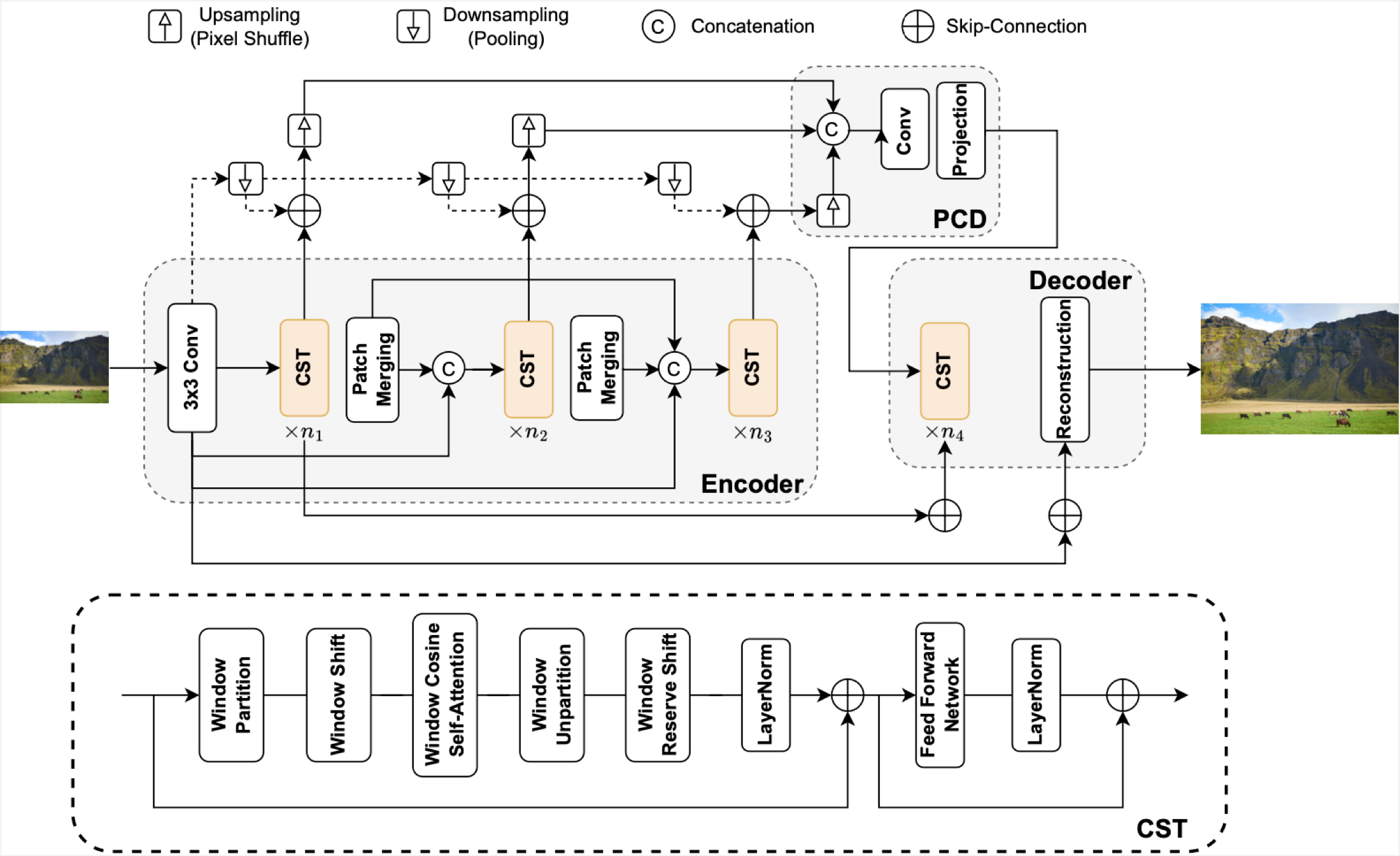}
  \caption{\textit{Team MagicSR: }Overall architecture of MagicSR-Light and CST. In the CST module, they adopt an asymmetric U-Net decoder architecture consisting of pixel-shuffle, concatenation, depth-wise convolution, and point-wise (linear) projection. This variant efficiently integrates multi-scale outputs from the encoder part, encompassing the shallow module.
}
\label{fig_magicsrframework}
\end{figure}

\textbf{Method.}
MagicSR team adopts the structure of the Swin Transformer with window self-attention (WSA) for single image super-resolution (SR). However, the plain WSA ignores the broad regions when reconstructing high-resolution images due to a limited receptive field, and requires intensive computations due to the nature of its structure. To overcome these problems, they propose context-aware neighbor local windows, inspired by N-Gram \cite{choi2023n}, to produce neighboring embeddings and interact with each other by sliding WSA to produce the context-aware features before window partitioning. As illustrated in Fig.~\ref{fig_magicsrframework}, the proposed MagicSR-Light consists of five components: a shallow module (a 3 × 3 convolution), three hierarchical encoder stages (with patch-merging) that contain CST blocks (context-aware Swin transformer modules), PCD bottleneck (pixel-shuffle, concatenation, depth-wise convolution, point-wise projection), a small decoder stage with CSTs, and an image reconstruction module. They employ CSTs by using the context-aware neighbor local windows and the scaled-cosine attention proposed by Swin V2 \cite{liu2022swin}. PCD bottleneck, which takes multi-scale outputs of the encoder, is a variant of bottleneck
from U-Net. They adopt a decoder module composed of a CST block and a reconstruction layer to produce the final RGB output. They train their model in an adversarial way \cite{luo2024and} to improve the robustness. Given a low-resolution (LR) image $I$, a shallow module (comprising a 3 × 3 convolution) is employed to extract pertinent features. These features subsequently traverse three encoder stages, each comprising $n_i$ CST Blocks, and undergo a 2 × 2 patch-merging process, with the exception of the final stage. Notably, the patch-merging mechanism mirrors that of the Swin Transformer \cite{liu2022swin}, albeit with a dimensional reduction from 4D to D instead of 2D. This reduction in network dimensionality via patch merging results in a notable decrease in attention computation requirements for the CST. In their model, they set $n_1,n_2,n_3 = 6,4,4$. 

As illustrated in the bottom of Fig.~\ref{fig_magicsrframework}, their Context-aware Swin Transformer (CST) adopts scaled-cosine attention and post-normalization proposed in SwinV2 \cite{liu2022swin}. The window size of the WSA module is set to 8 by default. In window partitioning, they implement the context-aware neighbor local windows by following \cite{choi2023n}. This algorithm is also identically applied to other Swin Transformer models (SwinIR-light \cite{liang2021swinir}), focusing only on better performances. 
The window shifts are operated in the even-numbered blocks, the same as in the Swin Transformer. The decoder module comprises $n_4$ CST blocks along with a reconstruction layer, where they set $n_4$ to 6. The reconstruction module is structured with a convolution layer, followed by a pixel-shuffler, and another convolution layer for output conversion. The input to this module incorporates a global skip connection originating from the shallow module of the encoder.

\subsection{DIRN}

\begin{figure}
\centering
\includegraphics[width=1\columnwidth]{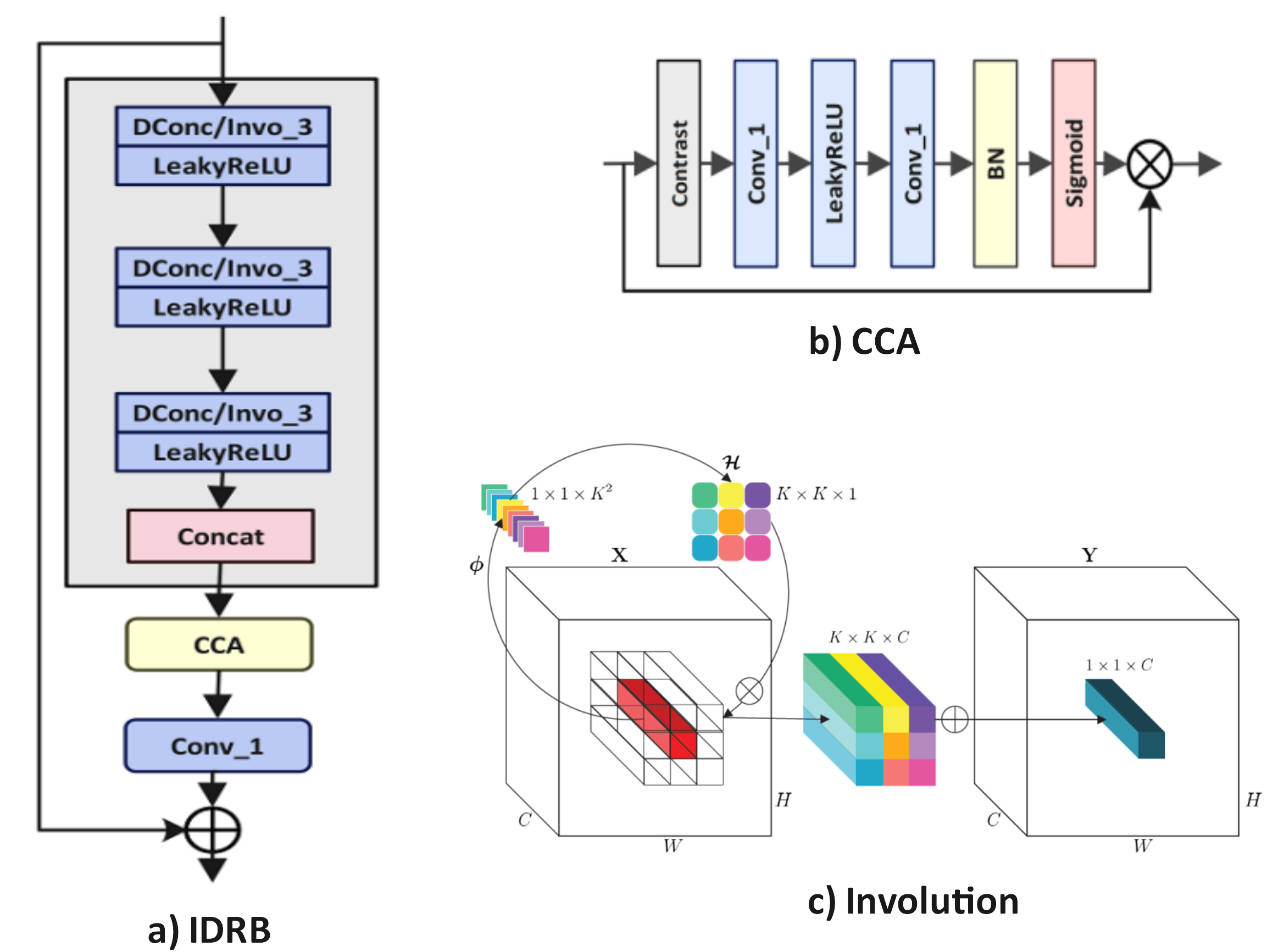}
\caption{\textit{Team DIRN: }(a) Depthwise convolution and Involution Residual Block(DIRB). (b) Contrast-Aware Channel Attention Block (CCA). (c) Involution\cite{li2021involution}.}
\label{fig:19_DIRN}
\end{figure}

\textbf {Method.}
Despite the rapid development of neural network architecture, convolution remains the mainstay of deep neural networks. In recent years, depth-wise separable convolution \cite{howard2017mobilenets} has been proposed to speed up deep models.  Duo Li.\cite{li2021involution} introduced the involution operation to match visual patterns concerning the location and reduce inter-channel redundancy. The reverse function of involution enhances efficiency and reduces parameters. A visual and efficient involution kernel belonging to a specific location can be generated by only considering the feature vector of the corresponding location, as shown in Figure. \ref{fig:19_DIRN}(c). By sharing the involution kernel along the channel dimension, the redundancy of the kernels is reduced. Considering the focus of involution on visual performance and self-attention, in the remaining blocks, they use involution along with depth convolution to extract deep features. They propose that the structure of network blocks varies according to the depth. Therefore depth convolution is used more in the initial blocks.  As shown in Figure.\ref{fig:19_DIRN}(a), the Depthwise convolution and Involution Residual Block (DIRB) block uses three layers of Invo/DConv 3$\times$3 + LeakyReLU to extract deep features. Feature maps are then connected along the channel dimension. Inspired by IMDN, they modified the Contrast-Aware Channel Attention Block (CCA) \cite{IMDN} and used it to evaluate the contrast degree of feature maps and improve performance.  They also used a 1$\times$1 convolution to reduce the number of channels and connections left at the end of the block.

\noindent\textbf{Training Details}.
They are using DIV2K\cite{DIV2K} and LSDIR\cite{lilsdir} images as the dataset. HR image patches of size \(96 \times 96\) are randomly cropped from HR images, and the small batch size is set to 16. They adopt the Adam optimizer and for the loss function, they use the L1 loss to measure the difference between the SR images and the ground truth. In DIRN, they set the number of DIRBs to 6. 
To investigate the Combination of Involution and Depthwise convolution at different depths, four different models were considered. In each of the models, as the network deepens, involution is used more than Depthwise convolution in the DIRB block, and in none of the models, the structure of the blocks is the same. Experiments with 100 epochs and evaluation results with validation dataset DIV2K\_LSDIR\_valid are presented. 
According to the results, it is appropriate to use Depthwise convolution in the initial blocks where feature maps have fewer changes and are similar to pixel values. As the network gets deeper, feature maps have undergone many changes along the network, and by ensuring correlation using Depthwise convolution, using involution as a visual operator can yield better results. According to the results of the fourth model, it performed better than other models and they chose it as the main model. The DIRN model is lightweight with 97K parameters and 6.48G for flops.

\subsection{ACVLAB}

\textbf{Method.}  The method of ACVLAB is based on \cite{li2023craft}, which maintains performance with a smaller number of parameters by incorporating high-frequency prior information. It includes three key components: high-frequency enhanced residual block (HFERB), shifted rectangular window attention block (SRWAB) for capturing high-frequency information to capture global information, and hybrid fusion Blocks (HFB) are designed to enhance the global representation. They reduce the network depth and adapt the small convolutional kernels in the deep feature extraction stage to further reduce the number of parameters.

\begin{figure*}
    \begin{center}
    \includegraphics[width=1\textwidth]{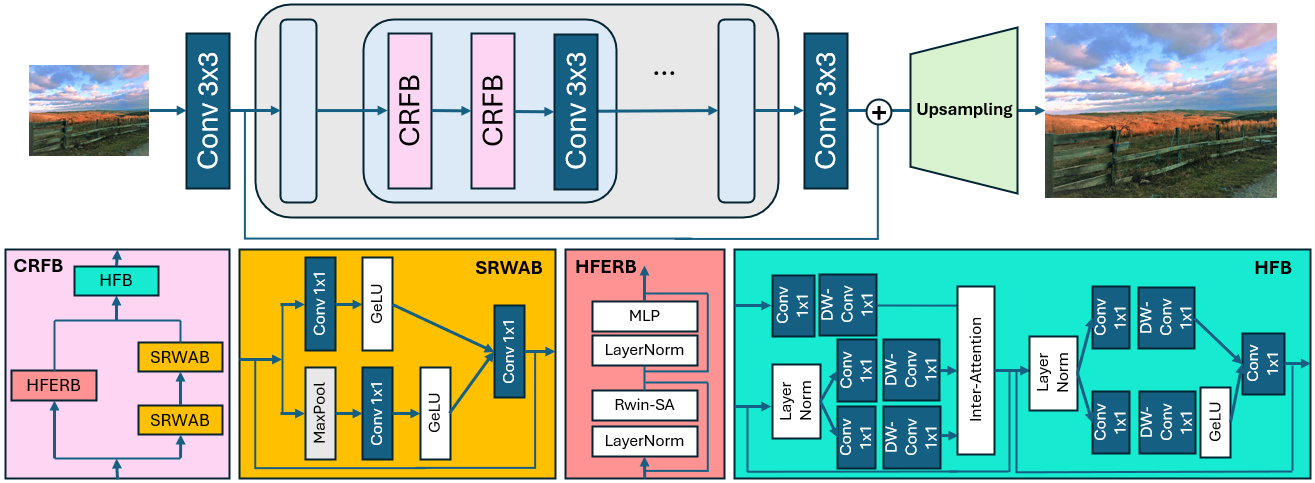}
    \end{center}
    \caption{\emph{Team ACVLAB:} The overall architecture.}
     \label{craft.png}
\end{figure*}

\noindent\textbf{Training Details.} They use LSDIR \cite{lilsdir} and DIV2K \cite{DIV2K} dataset for training. Their training stage can be divided into two stages. Throughout the entire training process, they adapt the Adam optimizer with $\beta_{1}$ = $0.9$, and $\beta_{2}$ = $0.999$ and train for $500000$ iterations in each stage. The learning rate is set to 2e-4, the multi-step learning scheduler is also used. The learning rate is halved at the [250000, 400000, 450000, 475000] iterations respectively. Weight decay is not applied. For the training stage, HR patches of size $256$ $\times$ $256$ are randomly cropped from HR images. The random horizontal flips and the random rotation are used for augmentation.  In the first stage, they use the L1-loss for the model optimization with a batch size of 16. In the second stage, they use the MSE-loss for enhancement. The model was implemented using Pytorch 1.13.1 and trained on a single NVIDIA-GeForce-RTX-3090.

\subsection{KLETech-CEVI\_Lowlight\_Hypnotise}
\begin{figure}[!htp]
    \centering
    \includegraphics[width=1\linewidth]{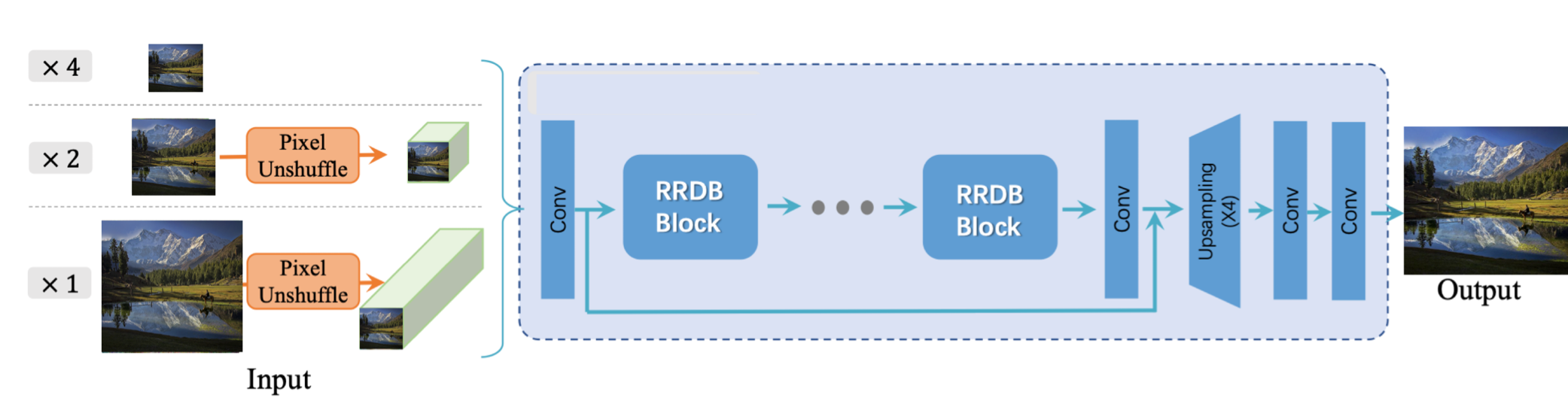}
    \caption{\textit{Team KLETech-CEVI\_Lowlight\_Hypnotise: }Overview architecture diagram of Efficient SRGAN, illustrating the network structure and flow of information during the super-resolution process. (Figure reproduced from \cite{wang2021real})}
    \label{fig:srgan}
\end{figure}

\textbf{Method.}  In this work, the team proposes an architecture named Efficient SRGAN for image super-resolution, aimed at achieving improved performance in terms of both reconstruction accuracy and computational efficiency.
Image super-resolution techniques like \cite{5557890}, \cite{10.1007/978-981-10-8633-5_41}, \cite{chen2024mffn}, \cite{al2024single} aim at improving the resolution of images.
Real-ESRGAN, an extension of the Enhanced Super-Resolution Generative Adversarial Network (ESRGAN), presents a powerful architecture for image super-resolution tasks. The architecture incorporates a high-order degradation model, to simulate complex real-world degradations accurately. Additionally, Real-ESRGAN addresses artifacts such as ringing and overshooting during synthesis, leading to improved visual quality in the output images. To further enhance performance, Real-ESRGAN utilizes a U-Net discriminator with spectral normalization, increasing discriminator capability and stabilizing training dynamics. Real-ESRGAN is trained on synthetic data, for improved generalization on real-world image restoration challenges. They extend their work by training Real-ESRGAN \cite{wang2021real} on custom weighted combinational loss function as shown in Equation \ref{eqn:sr}. However, Real-ESRGAN suffers from substantial texture information loss in the reconstructed image. To overcome this, they propose to use VGG-19 Perceptual loss inspired from \cite{Desai_2023_ICCV}, combined with $L1$ loss.
\begin{equation}
    \mathcal{L}_{SR} = \alpha * \mathcal{L}_{VGG} + \beta * \mathcal{L}_{1}
    \label{eqn:sr}
\end{equation}
where, $\mathcal{L}_{VGG}$ is VGG-19 perceptual loss \cite{vggloss}, and is given as, 
\begin{equation}
    \mathcal{L}_{VGG} = \frac{1}{WH} \sum{i=1}^{W}\sum{j=1}^{H}(\phi(\hat{y}){i,j}-\phi(x){i,j})
    \label{eqn:sr2}
 \end{equation}
 where, $\phi(.)$ is the activation of $j^{th}$ layer of network $\phi$ when processing on image $x$. $W$ and $H$ are width and height of image.
 $\alpha$ and $\beta$ are weights to the losses, and are set to $0.7$ and $0.5$ heuristically.

\noindent\textbf{Training Details}: They train the proposed methodology using the dataset provided by NTIRE 2024 Efficient Super-Resolution Challenge. They train the model using Python and PyTorch frameworks, on a patch resolution of 339*510, with a batch size of 8. They use Adam optimizer with $\beta1$ set to $0.9$ and $\beta2$ set to $0.999$. They train the model for 1000 epochs at a learning rate of $0.0002$. During testing, they use full-resolution images (339*510), on a single RTX 3090 GPU. The average testing time for a single image on the full resolution is 0.9s on RTX 3090 GPU.

\section*{Acknowledgments}
This work was partially supported by the Humboldt Foundation. We thank the NTIRE 2024 sponsors: Meta Reality Labs, OPPO, KuaiShou, Huawei and University of W\"urzburg (Computer Vision Lab).

\appendix

\section{Teams and affiliations}
\label{sec:teams}
\subsection*{NTIRE 2024 team}
\noindent\textit{\textbf{Title: }} NTIRE 2024 Efficient Super-Resolution Challenge\\
\noindent\textit{\textbf{Members: }} \\
Bin Ren$^{1,2}$ (\href{mailto:bin.ren@unitn.it}{bin.ren@unitn.it}),\\
Yawei Li$^3$ (\href{mailto:yawei.li@vision.ee.ethz.ch}{yawei.li@vision.ee.ethz.ch}),\\
Nancy Mehta$^4$ (\href{mailto:nancy.mehta@uni-wuerzburg.de}{nancy.mehta@uni-wuerzburg.de}),\\
Radu Timofte$^{3,4}$ (\href{mailto:Radu.Timofte@uni-wuerzburg.de}{Radu.Timofte@uni-wuerzburg.de})\\
\noindent\textit{\textbf{Affiliations: }}\\
$^1$ University of Pisa, Italy\\
$^2$ University of Trento, Italy\\
$^3$ Computer Vision Lab, ETH Z\"urich, Switzerland\\
$^4$ University of W\"urzburg, Germany\\

\subsection*{XiaomiMM}
\noindent\textit{\textbf{Title: }} Swift Parameter-free Attention Network for Efficient Image Super-Resolution\\
\noindent\textit{\textbf{Members: }} \\
Hongyuan Yu$^1$ (\href{mailto:yuhyuan1995@gmail.com}{yuhyuan1995@gmail.com}),\\
Cheng Wan$^2$, 
Yuxin Hong$^3$, 
Bingnan Han$^1$, 
Zhuoyuan Wu$^1$, 
Yajun Zou$^1$, 
Yuqing Liu$^1$, 
Jizhe Li$^1$, 
Keji He$^4$, 
Chao Fan$^5$, 
Heng Zhang$^1$, 
Xiaolin Zhang$^1$, 
Xuanwu Yin$^1$, 
Kunlong Zuo$^1$ \\
\noindent\textit{\textbf{Affiliations: }} \\ 
$^1$ Multimedia Department, Xiaomi Inc. \\
$^2$ Georgia Institute of Technology \\
$^3$ Lanzhou University \\
$^4$ Institute of Automation, Chinese Academy of Sciences \\
$^5$ Beijing University of Technology \\

\subsection*{Cao Group}
\noindent\textit{\textbf{Title: }} Double Reparameterization Network for Efficient Image Super-Resolution \\
\noindent\textit{\textbf{Members: }} \\
Bohao Liao$^1$ (\href{mailto:liaobh@mail.ustc.edu.cn}{liaobh@mail.ustc.edu.cn}),\\
Peizhe Xia$^1$, 
Long Peng$^1$, 
Zhibo Du$^1$, 
Xin Di$^1$, 
Wangkai Li$^1$, 
Yang Wang$^1$, 
Wei Zhai$^1$, 
Renjing Pei$^2$, 
Jiaming Guo$^2$, 
Songcen Xu$^2$, 
Yang Cao$^1$, 
Zhengjun Zha$^1$ \\
\noindent\textit{\textbf{Affiliations: }} \\ 
$^1$ University of Science and Technology of China \\
$^2$ Huawei Noah's Ark Lab \\
\subsection*{BSR}
\noindent\textit{\textbf{Title: }} Lightening Partial Feature Distillation Network for Efficient Super-Resolution\\
\noindent\textit{\textbf{Members: }} \\
Yan Wang$^1$ (\href{mailto:wyrmy@foxmail.com}{wyrmy@foxmail.com}),\\
Yi Liu$^2$, 
Qing Wang$^2$, 
Gang Zhang$^2$, 
Liou Zhang$^2$, 
Shijie Zhao$^2$ \\
\noindent\textit{\textbf{Affiliations: }} \\ 
$^1$ Nankai University \\
$^2$ ByteDance Inc. \\

\subsection*{PiXupt}
\noindent\textit{\textbf{Title: }}Hierarchical Attention Residual Network for Efficient Image Super-Resolution\\
\noindent\textit{\textbf{Members: }} \\
Yanyu Mao$^{1}$ (\href{mailto:bolt35982@gmail.com}{bolt35982@gmail.com}),\\
Ruilong Guo$^{1}$,
Nihao Zhang$^{1}$, 
Qian Wang$^{1,2}$\\
\noindent\textit{\textbf{Affiliations: }} \\ 
$^1$ Xian University of Posts and Telecommunications, Xi’an, China \\
$^2$ National Engineering Laboratory for Cyber Event Warning and Control Technologies\\

\subsection*{XJU\_100th Ann}
\noindent\textit{\textbf{Title: }} Attention Guidance Distillation Network for Efficient Image Super-Resolution\\
\noindent\textit{\textbf{Members: }} \\
Shuli Cheng$^1$ (\href{mailto:cslxju@xju.edu.cn}{cslxju@xju.edu.cn}),\\
Hongyuan Wang$^1$, 
Ziyan Wei$^1$, 
Qingting Tang$^1$, 
Liejun Wang$^1$, 
Yongming Li$^1$ \\
\noindent\textit{\textbf{Affiliations: }} \\ 
$^1$ School of Computer Science and Technology, Xinjiang University, Ürümqi, China \\

\subsection*{VPEG\_C}
\noindent\textit{\textbf{Title: }} SMFAN: A Lightweight Self-Modulation Feature Aggregation Network for Efficient Image Super-Resolution \\
\noindent\textit{\textbf{Members: }} \\
Mingjun Zheng$^1$ (\href{mailto:jun_zmj@qq.com}{jun\_zmj@qq.com}),\\
Long Sun$^1$, 
Jinshan Pan$^1$,
Jiangxin Dong$^1$, 
Jinhui Tang$^1$ \\
\noindent\textit{\textbf{Affiliations: }} \\ 
$^1$ Nanjing University of Science and Technology \\

\subsection*{ZHEstar}
\noindent\textit{\textbf{Title: }} Large Kernel Frequency-enhanced Network for Efficient Single Image Super-Resolution\\
\noindent\textit{\textbf{Members: }} \\
Jiadi Chen (\href{mailto:zjnucjd@163.com}{zjnucjd@163.com}), \\
Huanhuan Long, 
Chunjiang Duanmu \\
\noindent\textit{\textbf{Affiliations: }} \\ 
Zhejiang Normal University, Jinhua, China  \\

\subsection*{VPEG\_O}
\noindent\textit{\textbf{Title: }} SAFMN++: Improved Feature Modulation Network for Efficient Image Super-Resolution\\
\noindent\textit{\textbf{Members: }} \\
Long Sun$^1$ (\href{mailto:yourEmail@xxx.xxx}{cs.longsun@gmail.com}), \\
Jinshan Pan$^1$,
Jiangxin Dong$^1$, 
Jinhui Tang$^1$ \\
\noindent\textit{\textbf{Affiliations: }} \\ 
$^1$ Nanjing University of Science and Technology \\
 
\subsection*{CMVG}
\noindent\textit{\textbf{Title: }} Multi-supervisory Residual Knowledge Distillation Network for Efficient Super-resolution \\
\noindent\textit{\textbf{Members: }} \\
Xin Liu$^1$ (\href{mailto:yourEmail@xxx.xxx}{liuxin9976@163.com}),\\
Min Yan$^1$, 
Qian Wang$^1$ \\
\noindent\textit{\textbf{Affiliations: }} \\ 
$^1$ China Mobile Research Institute \\

\subsection*{LeESR}
\noindent\textit{\textbf{Title: }} Separable-Mixable Residual Network for Efficient image Super-Resolution\\
\noindent\textit{\textbf{Members: }} \\
Menghan Zhou$^1$ (\href{mailto:zhoumh3@lenovo.com}{zhoumh3@lenovo.com}),\\
Yiqiang Yan$^1$ \\
\noindent\textit{\textbf{Affiliations: }} \\ 
$^1$ Lenovo Research \\
\subsection*{AdvancedSR}
\noindent\textit{\textbf{Title: }} Progressive Kernel Pruning for Efficient Super-Resolution\\
\noindent\textit{\textbf{Members: }} \\
Yixuan Liu$^1$ (\href{mailto:yixuanl@amd.com}{yixuanl@amd.com}),\\
Wensong Chan$^1$, 
Dehua Tang$^1$, 
Dong Zhou$^1$, 
Li Wang$^1$, 
Lu Tian$^1$, 
Barsoum Emad$^1$ \\
\noindent\textit{\textbf{Affiliations: }} \\ 
$^1$ Advanced Micro Devices, Inc., Beijing, China  \\

\subsection*{ECNU\_MViC}
\noindent\textit{\textbf{Title: }} Intermittent Feature Aggregation with Distillation for Efficient Super-Resolution\\
\noindent\textit{\textbf{Members: }} \\
Bohan Jia$^1$ (\href{mailto:10205501423@stu.ecnu.edu.cn}{10205501423@stu.ecnu.edu.cn}),\\
Jincheng Liao$^1$, 
Junbo Qiao$^{1,2}$, 
Yunshuai Zhou$^1$, 
Yun Zhang{$^{2,3}$}, 
Wei Li{$^{2}$}, 
Shaohui Lin{$^{1}$} \\
\noindent\textit{\textbf{Affiliations: }} \\ 
$^1$ East China Normal University \\
$^2$ Huawei Noah's Ark Lab \\
$^3$ The Hong Kong University of Science and Technology

\subsection*{HiSR}
\noindent\textit{\textbf{Title: }} SlimRLFN: Making RLFN Smaller and Faster Again\\
\noindent\textit{\textbf{Members: }} \\
Shenglong Zhou$^1$ (\href{mailto:yourEmail@xxx.xxx}{slzhou96@mail.ustc.edu.cn}),\\
Binbin Chen$^2$ \\
\noindent\textit{\textbf{Affiliations: }} \\ 
$^1$ University of Science and Technology of China \\
$^2$ Huazhong University of Science and Technology \\

\subsection*{MViC\_SR}
\noindent\textit{\textbf{Title: }} LSANet: Efficient Image Super-Resolution with Lightweight Spaced Attention Mechanism\\
\noindent\textit{\textbf{Members: }} \\
Jincheng Liao$^1$ (\href{mailto:71265901062@stu.ecnu.edu.cn}{71265901062@stu.ecnu.edu.cn}),\\
Bohan Jia$^1$, 
Junbo Qiao$^{1,2}$, 
Yunshuai Zhou$^1$, 
Yun Zhang{$^{2,3}$}, 
Wei Li{$^{2}$}, 
Shaohui Lin{$^{1}$} \\
\noindent\textit{\textbf{Affiliations: }} \\ 
$^1$ East China Normal University \\
$^2$ Huawei Noah's Ark Lab \\
$^3$ The Hong Kong University of Science and Technology

\subsection*{LVTeam}
\noindent\textit{\textbf{Title: }} Explore RLFN's Extreme for Efficient Super Resolution\\
\noindent\textit{\textbf{Members: }} \\
Suiyi Zhao$^1$ (\href{mailto:meranderzhao@gmail.com}{meranderzhao@gmail.com}),\\
Zhao Zhang$^1$, 
Bo Wang$^1$, 
Yan Luo$^1$, 
Yanyan Wei$^1$ \\
\noindent\textit{\textbf{Affiliations: }} \\ 
$^1$ Hefei University of Technology \\

\subsection*{Fresh}
\noindent\textit{\textbf{Title: }} Depth Residual Local Feature Network for Efficient Super-Resolution\\
\noindent\textit{\textbf{Members: }} \\
Feng Li$^1$ (\href{mailto:fengli@hfut.edu.cn}{fengli@hfut.edu.cn}),\\
Mingshen Wang$^1$,
Yawei Li$^1$, 
Jinhan Guan$^1$, 
Dehua Hu$^1$ \\
\noindent\textit{\textbf{Affiliations: }} \\ 
$^1$ Hefei University of Technology \\

\subsection*{Lanzhi}
\noindent\textit{\textbf{Title: }} Residual Local Feature Block with Batch Normalization\\
\noindent\textit{\textbf{Members: }} \\
Jiawei Yu$^1$ (\href{mailto:yourEmail@xxx.xxx}{yujiawei@nudt.edu.cn}),\\
Qisheng Xu$^1$, 
Tao Sun$^1$, 
Long Lan$^1$, 
Kele Xu$^1$ \\
\noindent\textit{\textbf{Affiliations: }} \\ 
$^1$ National University of Defense Technology \\

\subsection*{Supersr}
\noindent\textit{\textbf{Title: }} Residual Local Feed Forward for Effcient Image Super-resolution\\
\noindent\textit{\textbf{Members: }} \\
Xin Lin$^1$ (\href{mailto:yourEmail@xxx.xxx}{linxin@stu.scu.edu.cn}),\\
Jingtong Yue$^1$,
Lehan Yang$^2$,
Shiyi Du$^3$,
Lu Qi$^4$,
Chao Ren$^1$ \\
\noindent\textit{\textbf{Affiliations: }} \\ 
$^1$ Sichuan University \\
$^2$ The University of Sydney \\
$^3$ Carnegie Mellon University \\
$^4$ The University of California, Merced \\

\subsection*{MeowMeowMeow}
\noindent\textit{\textbf{Title: }} Reparameterized Convolutional Network for Efficient Image Super-Resolution\\
\noindent\textit{\textbf{Members: }} \\
Zeyu Han$^1$ (\href{mailto:hanzeyu2001@outlook.com}{hanzeyu2001@outlook.com}),\\
Yuhan Wang$^1$,
Chaolin Chen$^1$ \\
\noindent\textit{\textbf{Affiliations: }} \\ 
$^1$ Sichuan University \\

\subsection*{Just Try}
\noindent\textit{\textbf{Title: }} Enhanced Reparameterize Residual Network for Efficient Image Super-Resolution\\
\noindent\textit{\textbf{Members: }} \\
Haobo Li$^1$ (\href{mailto:1114940412@qq.com}{1114940412@qq.com}),\\
\noindent\textit{\textbf{Affiliations: }} \\ 
$^1$Independent Researcher
\subsection*{VPEG\_E}
\noindent\textit{\textbf{Title: }} Enhanced Gated Feature Modulation for Efficient Super-Resolution\\
\noindent\textit{\textbf{Members: }} \\
Zhongbao Yang$^1$ (\href{mailto:yourEmail@xxx.xxx}{yangzhongbao40@gmail.com}),\\
Long Sun$^1$,
Lianhong Song$^1$,
Jinshan Pan$^1$,
Jiangxin Dong$^1$, 
Jinhui Tang$^1$ \\
\noindent\textit{\textbf{Affiliations: }} \\ 
$^1$ Nanjing University of Science and Technology \\

\subsection*{BU-ESR}
\noindent\textit{\textbf{Title: }} Boosting Residual Feature Distillation Networks through Knowledge Dilation\\
\noindent\textit{\textbf{Members: }} \\
Xingzhuo Yan$^1$ (\href{mailto:ayx1sgh@bosch.com}{ayx1sgh@bosch.com}),\\
Minghan Fu$^2$ \\
\noindent\textit{\textbf{Affiliations: }} \\ 
$^1$ Bosch Investment Ltd. \\
$^2$ University of Saskatchewan \\

\subsection*{Lasagna}
\noindent\textit{\textbf{Title: }} Efficient Enhanced Residual Network for Efficient Image Super-Resolution\\
\noindent\textit{\textbf{Members: }} \\
Jingyi Zhang$^1$ (\href{mailto:jingyizhang0806@163.com}{jingyizhang0806@163.com}),\\
Baiang Li$^1$, 
Qi Zhu$^2$, 
Xiaogang Xu$^{3,4}$, 
Dan Guo$^1$, 
Chunle Guo*$^5$ \\
\noindent\textit{\textbf{Affiliations: }} \\ 
$^1$ Hefei University of Technology \\
$^2$ University of Science and Technology of China \\
$^3$ The Chinese University of Hong Kong \\
$^4$ Zhejiang University\\
$^5$ Nankai University \\

\subsection*{BlingBling}
\noindent\textit{\textbf{Title: }} DVMSR:  Distillated Vision Mamba Super-Resolution\\
\noindent\textit{\textbf{Members: }} \\
Xiaoyan Lei$^1$ (\href{mailto:xyan\_lei@163.com}{xyan\_lei@163.com}),\\
Jie Liu$^1$, 
Weilin Jia$^1$, 
Weifeng Cao$^1$, 
Wenlong Zhang$^2$ \\
\noindent\textit{\textbf{Affiliations: }} \\ 
$^1$ Zhengzhou University of Light Industry \\
$^2$ The Hong Kong Polytechnic University \\

\subsection*{Minimalist}
\noindent\textit{\textbf{Title: }} Efficient Image Super-Resolution with Lightweight Multi-head Spatial Attention \\
\noindent\textit{\textbf{Members: }} \\
Manoj Pandey$^1$ (\href{mailto:pandeymanoj@deltax.ai}{pandeymanoj@deltax.ai}),\\
Maksym Chernozhukov$^1$, 
Giang Le$^1$ \\
\noindent\textit{\textbf{Affiliations: }} \\ 
$^1$ DeltaX \\
\subsection*{MagicSR}
\noindent\textit{\textbf{Title}} Context-aware Transformer for Efficient Image Super-resolution \\
\noindent\textit{\textbf{Members: }} \\
Yanhui Guo$^1$ (\href{mailto:guoy143@mcmaster.ca}{guoy143@mcmaster.ca}),\\
Hao Xu$^1$ \\
\noindent\textit{\textbf{Affiliations: }} \\ 
$^1$ McMaster University \\

\subsection*{DIRN}
\noindent\textit{\textbf{Title: }} Deep combination of Depthwise and Involution for the residual network of super-resolution image\\
\noindent\textit{\textbf{Members: }} \\
Akram Khatami-Rizi $^1$ (\href{mailto:akramkhatami67@gmail.com}{akramkhatami67@gmail.com}),\\
Ahmad Mahmoudi-Aznaveh $^2$ \\
\noindent\textit{\textbf{Affiliations: }} \\ 
$^1$ Cyberspace Research Institute of Shahid Beheshti University of Iran\\

\subsection*{ACVLAB}
\noindent\textit{\textbf{Title: }} Solution for NTIRE 2024 Efficient SR Challenge\\
\noindent\textit{\textbf{Members: }} \\
Chih-Chung Hsu$^1$ (\href{mailto:cchsu@gs.ncku.edu.tw}{cchsu@gs.ncku.edu.tw}),\\
Chia-Ming Lee$^1$ ,
Yi-Shiuan Chou$^1$ \\
\noindent\textit{\textbf{Affiliations: }} \\
$^1$ Institute of Data Science, National Cheng Kung University \\ 

\subsection*{KLETech-CEVI\_Lowlight\_Hypnotise}
\noindent\textit{\textbf{Title: }{Efficient SRGAN Towards Super-Resolution of Images}}\\
\noindent\textit{\textbf{Members: }} \\
Amogh Joshi$^1$ (\href{mailto:joshiamoghmukund@gmail.com}{joshiamoghmukund@gmail.com}),\\
Nikhil Akalwadi$^{1,3}$, 
Sampada Malagi$^{1,3}$, 
Palani Yashaswini$^{1,2}$, 
Chaitra Desai$^{1,3}$, 
Ramesh Ashok Tabib$^{1,2}$, 
Ujwala Patil$^{1,2}$, 
Uma Mudenagudi$^{1,2}$ \\
\noindent\textit{\textbf{Affiliations: }} \\ 
$^1$ Center of Excellence in Visual Intelligence (CEVI), KLE Technological University, Hubballi, Karnataka, INDIA \\
$^2$ School of Electronics and Communication Engineering, KLE Technological University, Hubballi, Karnataka, INDIA \\
$^3$ School of Computer Science and Engineering, KLE Technological University, Hubballi, Karnataka, INDIA \\

{
    \small
    \bibliographystyle{ieeenat_fullname}
    \bibliography{main}
}

\end{document}